\pdfoutput=1
\documentclass[final,5p,10pt,twocolumn]{elsarticle}

\usepackage{amssymb}
\usepackage{amsmath}
\usepackage{hyperref}

\usepackage{lineno}
\usepackage{hyperref}
\usepackage{enumitem}
\modulolinenumbers[5]

\usepackage{subcaption}
\usepackage{mathtools}
\usepackage{amsmath}
\usepackage{amssymb}
\usepackage{stmaryrd}
\usepackage{amsthm}
\usepackage{amsfonts}
\usepackage{dsfont}
\usepackage{graphicx}
\usepackage{upgreek}
\usepackage{bm}
\usepackage{color}
\usepackage{float}
\usepackage{tikz-cd}
\usepackage{scrextend}
\usepackage[cal=boondoxo]{mathalfa}
\usepackage{nicefrac}
\usepackage{algorithm,algpseudocode} 
\usepackage{todonotes}
\usepackage{cleveref}

\biboptions{numbers,sort&compress}

\DeclareFontFamily{OT1}{pzc}{}
\DeclareFontShape{OT1}{pzc}{m}{it}{<-> s * [1.200] pzcmi7t}{}
\DeclareMathAlphabet{\mathpzc}{OT1}{pzc}{m}{it}

\newcommand{\bx}{{\mathbf{x}}}

\newcommand{\bu}{{\mathbf{u}}}

\newcommand{\bff}{{\mathbf{f}}}

\newcommand{\bK}{{\mathbf{K}}}
\newcommand{\indic}{{\mathbf{1}}}

\newcommand{\dimm}{\mathrm{d}}
\newcommand{\vol}{{\mathsf{vol}}}

\newcommand{\R}{{\mathds{R}}}

\newcommand{\SE}[1]{{\mathrm{SE}(#1)}}

\newcommand{\shape}{S}

\newcommand{\motion}{M}

\newcommand{\field}{\mathcal{f}}
\newcommand{\gield}{\mathcal{g}}

\newcommand{\cell}{C}

\newcommand{\weight}{w}

\newcommand{\Gield}{\mathcal{G}}

\newcommand{\TS}{\mathcal{T}}

\renewcommand{\th}{$^\text{th}$ }

\theoremstyle{definition}

\newcommand{\eq}[1]{(\ref{#1})} %Use \eq{...} for referincing equations as (...)
\newcommand{\com}[1]{} %Use \com{...} to commenting out multiple paragraphs

\journal{Solid Physical Modeling Symposium 2023}

\begin{document}

\begin{frontmatter}

%% Title, authors and addresses
% \title{Topology Optimization for Co-design of Moving Parts for Compliance and Collision Avoidance}
\title{Co-design Optimization of Moving Parts for Compliance and Collision Avoidance}
\author[PARC]{Amir M. Mirzendehdel}
\author[PARC]{Morad Behandish}
\affiliation[PARC]{organization={Palo Alto Research Center},%Department and Organization
            addressline={3333 Coyote Hill Road}, 
            city={Palo Alto},
            postcode={94304}, 
            state={CA},
            country={USA}}

% %%Graphical abstract
% \begin{graphicalabstract}
% \includegraphics{grabs}
% \end{graphicalabstract}

% %%Research highlights
% \begin{highlights}
% \item Research highlight 1
% \item Research highlight 2
% \end{highlights}

\begin{abstract}
%% Text of abstract
Design requirements for moving parts in mechanical assemblies are typically specified in terms of interactions with other parts. Some are purely kinematic (e.g., pairwise collision avoidance) while others depend on physics and material properties (e.g., deformation under loads). Kinematic design methods and physics-based shape/topology optimization (SO/TO) deal separately with these requirements. They rarely talk to each other as the former uses set algebra and group theory while the latter requires discretizing and solving differential equations. Hence, optimizing a moving part based on physics typically relies on either neglecting or pruning kinematic constraints in advance, e.g., by restricting the design domain to a collision-free space using an unsweep operation. In this paper, we show that TO can be used to co-design two or more parts in relative motion to simultaneously satisfy physics-based criteria and collision avoidance. We restrict our attention to maximizing linear-elastic stiffness while penalizing collision measures aggregated in time. We couple the TO loops for two parts in relative motion so that the evolution of each part’s shape is accounted for when penalizing collision for the other part. The collision measures are computed by a correlation functional that can be discretized by left- and right-multiplying the shape design variables by a pre-computed matrix that depends solely on the motion. This decoupling is key to making the computations scalable for TO iterations. We demonstrate the effectiveness of the approach with 2D and 3D examples.
\end{abstract}


\end{frontmatter}

 % \linenumbers

\section{Introduction} \label{sec_intro}
Design for Assembly (DFA) describes a set of principles and guidelines widely used in modern product design that enable manufacturers to improve product quality and performance while reducing assembly time and cost.  
The DFA process typically begins with a thorough analysis of the product design to identify any potential assembly problems or areas for improvement. Subsequently, through part consolidation and reducing the need for specialized tools and equipment during assembly, the entire design-to-fabrication-to-assembly workflow becomes more sustainable and profitable. 
Current processes mainly rely on expensive, time-consuming, and labor-intensive iterations. Recent advances in shape and topology optimization (TO) have enabled engineers to explore the feasible design spaces more rapidly and avoid tedious trial and error at the early stages of design. 

In recent years, incorporating various physics objectives \cite{li2022three} and manufacturing constraints \cite{mirzendehdel2020topology,liu2018current} in TO have been widely researched and significant effort has been spent on developing and refining algorithms for finding high-performance lightweight structures in aerospace \cite{guanghui2020aerospace}, automotive \cite{jankovics2019customization}, and medical \cite{wu2021advances} applications. However, despite its importance in many engineering design problems, less attention has been paid to incorporating kinematic constraints into the optimization process. One such constraint is collision avoidance, whose incorporation into TO requires simultaneous reasoning about interactions and mechanics of multiple moving parts.

In this study, we develop a TO method to co-design multiple parts in relative motion by coupling stiffness with collision avoidance. TO is a computational design approach to distribute material in 2D or 3D space. Incorporating the collision avoidance constraint into gradient-descent TO requires formulating a collision measure, whose differentiation leads to a locally evaluable sensitivity field with respect to material placement at different locations in the design domain. A challenge is that alterations made to one component could introduce or eliminate collisions with other components. As a result, a co-design process is required in which the components are modified simultaneously, where the evolution of one component directly impacts the design of the rest of the assembly. Further, evaluating pairwise collisions during the entire motion at every optimization loop becomes time intensive.   

To the best of our knowledge, there is no co-design method that enables the creation of intricate shapes while accounting for collision under arbitrary movements. This study presents a framework to design sets of components by simultaneously considering collision avoidance along with other constraints such as performance and manufacturing. 
We focus on mechanics under linear-elastic small deformations and use stiffness as the objective function for physics-based requirements. To ensure collision avoidance, we penalize the objective sensitivity field with the gradient of collision measures aggregated over time. Specifically, the paper outlines a Pareto-tracing TO method for combining the topological sensitivity field (TSF) \cite{mirzendehdel2015pareto} with collision gradient \cite{stockli2020topology,morris2022co}. Moreover, to make the collision evaluation scalable for TO iterations, the collision measures are computed using a correlation functional that can be discretized by left- and right- multiplication of the topology-dependent design variables with a pre-computed matrix dependent only on the motion. 
While other constraints are not the focus of this paper, a general approach for incorporating them with collision constraints using the principles outlined in \cite{mirzendehdel2019exploring} is demonstrated. 

% so that each part's evolving shape is taken into account when computing collision penalties for the other part. The collision measures are computed using a correlation functional that can be discretized by left- and right-multiplying the shape-dependent design variables with a pre-computed matrix that only depends on the motion. This decoupling is crucial for making the computation scalable for TO iterations.

% In this paper, we present a 
%  Recent advances in computation design optimization such as topology optimization (TO) have automated 

% In the design of moving parts in mechanical assemblies, the requirements are typically specified based on their interactions with other components. These requirements can be categorized into two types: kinematic and physics-based. Kinematic requirements only depend on geometry and motion, such as avoiding collision and maintaining contact, while physics-based requirements take into consideration material properties and the effects of loads applied at the interfaces.

% Kinematic design methods and physics-based shape/topology optimization (SO/TO) are two separate approaches that address these requirements. However, they are not typically integrated as they use different techniques, such as set algebra and group theory for kinematic design and discretizing and solving differential equations for SO/TO.

% 

\subsection{Related work} \label{sec_litReview}

Apart from different performance and manufacturing requirements imposed on individual parts, there are other crucial design factors whose consideration requires spatial reasoning about the relative movement (both translation and rotation) of parts. Collision avoidance, in particular, is critical in assembly, packaging, navigation, and accessibility requirements. These factors cannot be simply resolved through techniques commonly utilized in TO, such as design rules or sensitivity filtering. Instead, they are often expressed through kinematic constraints, expressed in a set- and group-theoretic language (such as affine transformations, Boolean operations, and containment relations), in contrast to the language of real-valued functions utilized for (in)equality constraints in TO.  

Ilie\c{s} and Shapiro \cite{ilies2000shaping,iliecs2002class} introduced the unsweep as a fundamental operation in solid modeling, which is used to generate the maximal allowable shape of a rigid part that can move against another part with a given shape while satisfying collision avoidance or containment constraints. %The unsweep operation involves computing the sweep of one solid along a prescribed path and then taking the union of all possible sweeps. The result is a solid that represents the maximum volume or shape that can be obtained while ensuring that there are no collisions or interference between the parts.
The idea of shaping through motion was further generalized to configuration space operations \cite{nelaturi2011configuration} using group morphology \cite{lysenko2010group}, which were subsequently used in solving problems in manufacturing analysis and process planning for additive and subtractive processes \cite{behandish2018automated,behandish2019classification,nelaturi2019automatic}.

Maximal sets (including unsweeps) can be represented implicitly (via point membership classification) in terms of pointwise constraints. Noting that every subset of the maximal collision-free set satisfies collision avoidance, unsweep can be used prior to TO to prune the feasible design domain \cite{mirzendehdel2019exploring}. For applications in which maximal sets cannot be defined, such as accessibility for multi-axis machining, the kinematics constraints can be directly incorporated in the optimization loop through a spatially varying field (e.g., inaccessibility measure field) that augments the primary sensitivity field \cite{mirzendehdel2020topology,mirzendehdel2021optimizing,mirzendehdel2022topology}.

St{\"o}ckli and Shea \cite{stockli2020topology} proposed a rule-based TO approach for generating collision-free rigid bodies with given inertia properties. They introduced the concept of collision matrix, which needs to be computed only once. The subsequent collision evaluation at each optimization iteration can be achieved through efficient matrix-vector multiplications. This idea was employed in \cite{morris2022co} to extend the notion of maximality to pairs of objects and to incrementally co-generate ``maximal pairs'' of collision-free parts only based on kinematic constraints.

Current methods for creating shapes that comply with constraints that have to do with collisions and contacts under arbitrary motion are overly restrictive, limiting the potential for more efficient and cost-effective assembly designs. Our method builds on the previous works in \cite{stockli2020topology,morris2022co} to provide a general gradient-based formulation for automated  co-design of high-performance collision-free solids. 

\subsection{Contributions \& Outline} \label{sec_contributions}

This article presents a TO-based computational framework for co-generating collision-free shapes in arbitrary relative motion. We show that:

\begin{enumerate}
    \item The scope of TO can be broadened to co-optimize moving components of an assembly with respect to physics-based performance and collision avoidance.
    
    \item The collision of \textit{multiple} solids in relative motion can be measured locally and globally to use as a topologically differentiable collision measure for gradient-based optimization.
    \item The collision measure can be augmented with other sensitivity fields to attractively and incrementally co-generate sets of collision-free solids.
    \item The optimization process can co-generate nontrivial shapes in 2D and 3D. 
\end{enumerate}

One possible use-case for this approach is the design of spatial linkages in which the motion is pre-determined from the kinematic analysis. Current approaches typically use simple geometry (e.g., use straight rods for links) to make collision avoidance tractable. Our approach enables designing links of arbitrarily complex shapes to achieve the best mechanical performance while avoiding collisions.

\section{Collision Measures} \label{sec_collisionMeasure} \label{sec_col}

%For given initial designs of a pair of solid parts and a given relative motion between them, we begin by defining a field over each solid which quantifies the severity of collision between the two solids. The goal is to quantify how much each point on each solid contributes to this collision, thereby guiding addition and removal of points in iterative co-design.
	
	%
	%
    %Throughout this paper, a `design' $\Omega$ refers to a computational model of a {\it single} designed artifact. We begin the process by specifying a bounded {\it design domain} ${\mathbf \Omega} \subset \R^{\dimm}$ ($\dimm = 2$ or  $3$) whose corresponding {\it design space} $\dspace$ is the collection of all `solids',i.e., closed-regular semianalytic pointsets in $\dimm-$space \cite{requicha1980representations}.
    
    %
    %
    %% Clint Comment: In theory we can do all of these definitions for n domains. Is it worth formulating? 
    %Multiple design domains can simultaneously occupy $\R^{\dimm}$ allowing us to define two distinct design domains, ${\mathbf {\Omega_A}}$ and ${\mathbf {\Omega_B}}$ with associated designs, ${\Omega_A}$ and ${\Omega_B}$. From these designs, we can define indicator functions, $\indic_A(\mathbf {x})$ and $\indic_B(\mathbf x)$  which determine whether a query point, $\mathbf {x} \in \R^{\dimm}$, belongs to ${\Omega_A}$ and ${\Omega_B}$, respectively. It should be noted, that query points can simultaneously belong to multiple designs if the designs overlap, i.e. $\Omega_A \cap \Omega_B \neq \emptyset$.
 
In this section, we briefly review the set-theoretic and discretized formulations for collision measures under motion and provide generalized expressions for multiple parts. 

\subsection{Set-Theoretic Formulation} \label{sec_set}

\begin{figure} [t!]
    \centering
    \begin{subfigure}[t]{\linewidth}
    \centering
        \includegraphics[width=0.5\linewidth]{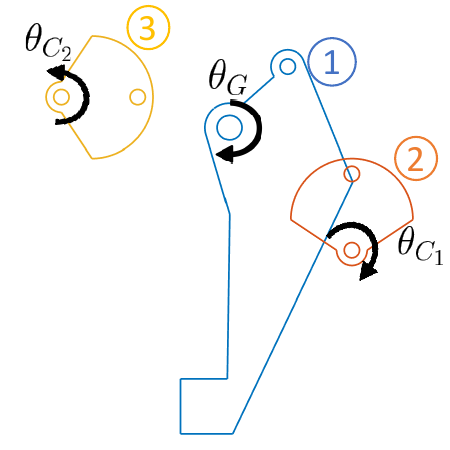}
        \caption{Motion of the gripper and two cams.}
        \label{fig_GripperCamsMotion}
    \end{subfigure}

    \centering
    \begin{subfigure}[t]{\linewidth}
        \includegraphics[width=\linewidth]{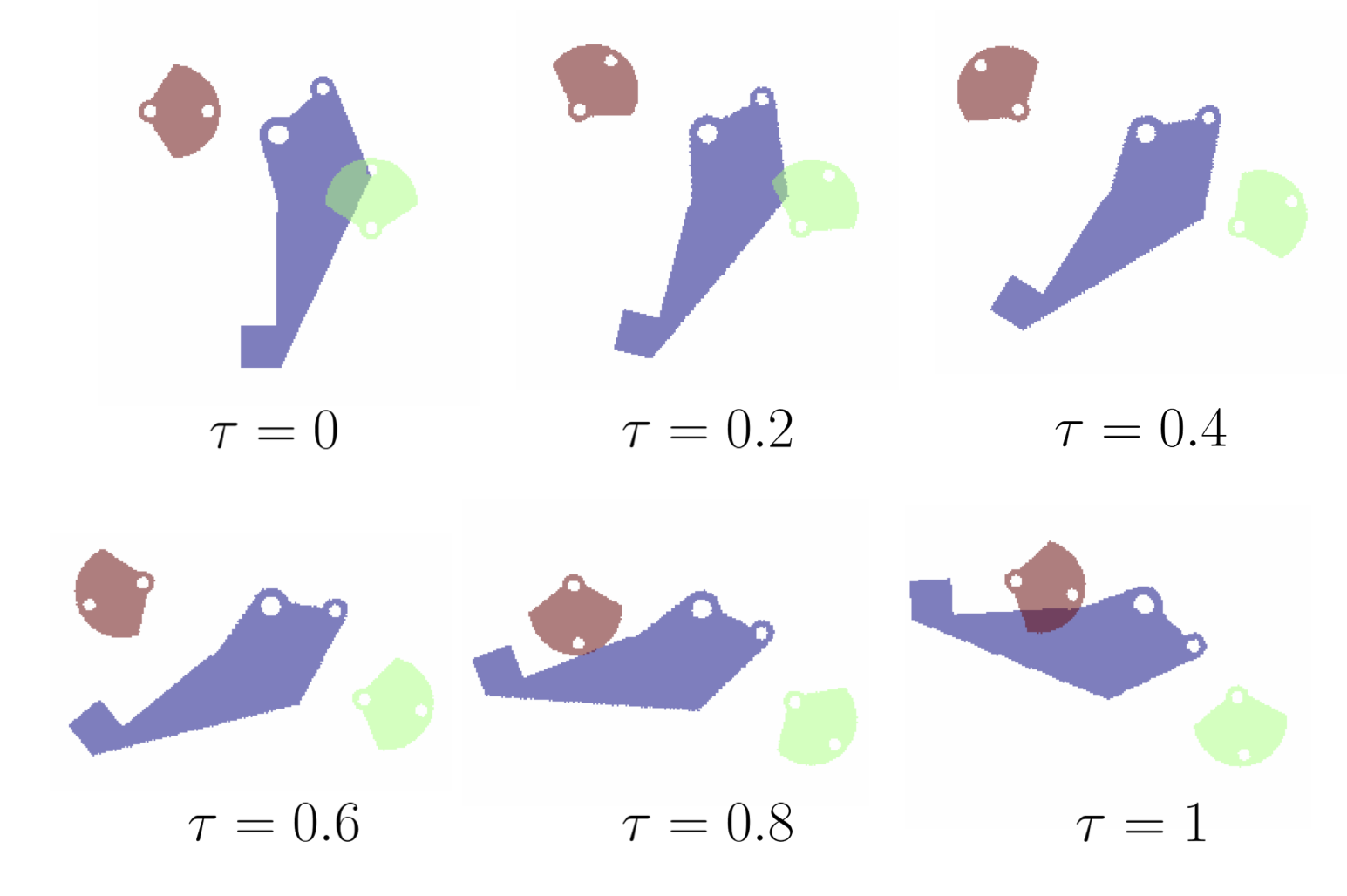}
        \caption{Sample snapshots.}
        \label{fig_GripperSnapshots}
    \end{subfigure}
    \centering
    \caption{Motion of the Gripper and two cams.}
    \label{fig_GripperCamsBC}
\end{figure}

Let us consider multiple design domains under motion $\Omega_i \subseteq \R^\dimm,~i=1,\cdots,N$ for the $N$ solids we aim to co-design ($\dimm = 2, 3$). Let us begin from the given initial designs $\shape_i \subseteq \Omega_i$ representing the shape of each solid at before applying any motion. Let $\motion_i \in \SE{\dimm}$ be the motion of the $i^{th}$ solid with respect to a common frame of reference. For instance for one-parametric motions:
\begin{align}
    \motion_i \triangleq \big\{ \tau_i(t) ~|~ 0 \leq \tau \leq 1 \big\},~i=1,\cdots,N
\end{align}
where $\tau_i : [0, 1] \to \SE{\dimm}$ are continuously time-variant configurations, and can be represented by homogeneous matrices. The displaced solids at any given time $t \in [0, 1]$ are:
\begin{align}
    \shape_i(t) \triangleq \tau_i(t)\shape_i = \big\{ \tau_i(t) \bx ~|~ \bx \in \shape_i \big\}, ~i=1,\cdots,N
\end{align}
Without loss of generality, we assume $\tau_i(0)$ to be identity, so $\shape_i(0) = \shape_i, \forall i$, as intended.

To formulate collision measures, it is more convenient to represent the pointsets implicitly via binary membership tests, also called indicator or characteristic functions $\indic_{\shape_i} : \R^\dimm \to \{0, 1\}$, defined generally by:
\begin{equation}
    \indic_{\shape}(\bx) \triangleq \left\{
    \begin{array}{ll}
         1 & \text{if}~ \bx \in \shape,  \\
         0 & \text{otherwise}.
    \end{array}
    \right.
\end{equation}
Note that indicator functions are contra-variant with rigid transformations, i.e., $\indic_{\tau \shape} (\bx) = \indic_{\shape}(\tau^{-1} \bx)$, meaning that a membership query for a given point against the displaced solid can be computed by displacing the query point along the inverse trajectory and testing its membership against the stationary solid.

Let $\motion_{i,j} = \motion_i^{-1} \motion_j$ stand for the relative motion of $\shape_j$ as observed from a frame of reference attached to $\shape_i$, noting that by kinematic inversion, $\motion_{j,i} \triangleq \motion_{i,j}^{-1} = \motion_j^{-1} \motion_i$ would represent the relative motion of $\shape_i$ as observed from a frame of reference attached to $\shape_j$.
\begin{align}
    \motion_{i,j} \triangleq \motion_i^{-1} \motion_j = \big\{\tau_{i,j}(t) ~|~ 0 \leq t \leq 1 \big\},~\forall i \neq j
\end{align}
where $\tau_{i,j}(t) = \tau_i^{-1}(t) \tau_j(t)$. 
The displaced solids at any given time $t \in [0, 1]$ in the relative frames are:
\begin{align}
    \shape_{i,j}(t) \triangleq \tau_{j,i}(t)\shape_i = \big\{ \tau_{j,i}(t) \bx ~|~ \bx \in \shape_i \big\}, ~\forall i \neq j \label{eq_S12}
\end{align}

To quantify the contribution of every point $\bx \in \R^\dimm$ attached to $\shape_i$ (resp. $\shape_j$) to its collision with $\shape_j$ (resp. $\shape_i$), we can measure the duration of its trajectory that collides with $\shape_j$ (resp. $\shape_i$):
\begin{align}
    \field_{\shape_i}(\bx) &\triangleq \int_0^1 \indic_{\shape_{j,i}(t)}(\bx) ~dt = \int_0^1 \indic_{\shape_j}(\tau_{j,i}\bx) ~dt,~\forall i \neq j \label{eq_f1_}
\end{align}
%
% as illustrated in Fig. \ref{PointCorrelationFig}.
To eliminate the contribution of the points that are outside each shape, we can multiply by the indicator functions of each shape:
\begin{align}
    \overline{\field}_{\shape_i}(\bx) &\triangleq  \int_0^1 \indic_{\shape_j}(\tau_{j,i}\bx) \indic_{\shape_i}(\bx) ~dt,~\forall i\neq j  \label{eq_f1} 
\end{align}
%
%
    % \begin{figure}[h]
    % 	\centering
    % 	\includegraphics[width=0.45\textwidth]{fig/PointCorrelationFig}
    % 	\caption{Local measure of collision, $\overline{\field}_{\shape_1}(\bx)$, of two query points, $\mathbf{x_1}$ and $\mathbf{x_2}$. The relative trajectory of each query points is determined by the relative configurations, $\tau_{1,2}(t)$ of $S_{2}$ to $S_{1}$.} \label{PointCorrelationFig}
    % \end{figure}
%    

To derive global measures (a single value for each solid) from the above local measures, we can integrate them over the respective solids:
\begin{align}
    \gield^{}_{\shape_i} &\triangleq \int_{\shape_i} \field_{\shape_i}(\bx) ~d\mu^\dimm[\bx] = \int_{\Omega_i} \overline{\field}_{\shape_i}(\bx) ~d\mu^\dimm[\bx], ~ \forall i
     \label{eq_g1}
\end{align}

The goal of collision-free co-design is to find set of collision-free solids $\shape_i \subseteq \Omega_i,~\forall i$, in a sense that we shall define precisely below, such that $\gield^{}_{\shape_i} = \gield^{}_{\shape_j} = 0,~ \forall i \neq j$, which is true iff for all $\bx \in \R^\dimm$, $\overline{\field}_{\shape_i}(\bx) = \overline{\field}_{\shape_j}(\bx) = 0,~ \forall i \neq j$.

\subsection{Discretized Formulation} \label{sec_comp}

\begin{figure} [t!]
    \centering
    \begin{subfigure}[t]{0.4\linewidth}
    \centering
        \includegraphics[width=\linewidth]{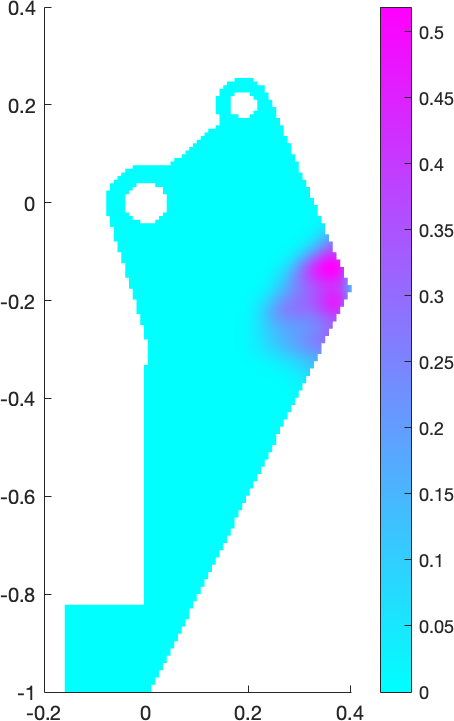}
        \caption{$\gield^{}_{\shape_{1,2}}$}
        \label{fig_GripperCollision-12}
    \end{subfigure}
    \begin{subfigure}[t]{0.4\linewidth}
    \centering
        \includegraphics[width=\linewidth]{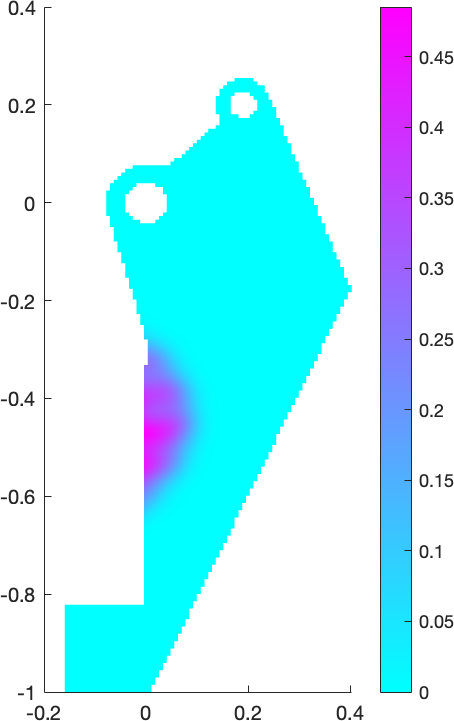}
        \caption{$\gield^{}_{\shape_{1,3}}$}
        \label{fig_GripperCollision-13}
    \end{subfigure}
    \centering
    \caption{Pair-wise collision fields for the gripper and the two cams.}
    \label{fig_GripperCamsPairwiseCollision}
\end{figure}

The collision measures discussed in Section \ref{sec_set} are representation-agnostic and various representation schemes (e.g., B-reps, mesh, and voxels) can be used as long as they support evaluation $\dimm-$integrals presented in \eq{eq_f1_}-\eq{eq_g1}. 
Since we plan to use the same representation TO, where we discretize the domain into uniform grid elements. Similar to the formulation presented in \cite{morris2022co}, we employ an asymmetric discretization strategy where the stationary solid (i.e., the one to which the frame of reference is attached) is discretized via a finite volume scheme, while the moving solid (i.e., the one whose motion is observed) is discretized via a finite sample scheme. In other words, given a sufficiently fine discretization (i.e., edge length of $\epsilon > 0$), we use the the primal grid nodes (i.e., vertices of the finite elements $v_j$) for the moving part $j$ and the dual grid cells (i.e., finite elements $e_i$) for the stationary part $i$. Let $\bx_i \in \R^\dimm$ be the coordinates of the i\th vertex (i.e., $0-$cell) on the grid and $\cell_i \subset \R^\dimm$ denote the dual $\dimm-$cell (e.g., congruent quadrilateral elements in 2D and hexahedral voxels in 3D), we can define: 
\begin{equation}
    \cell_i \triangleq \big\{ \bx_i + \bx ~ \big|~ \bx \in C \big\}, \quad \cell \triangleq \big[-\nicefrac{\epsilon}{2}, +\nicefrac{\epsilon}{2} \big]^\dimm
\end{equation} 
\Cref{fig_GripperCamsPairwiseCollision} illustrates the pair-wise collision fields for the gripper with each of the two cams.

The finite approximations of the collision measure in \eq{eq_g1} can be written as matrix equations:
\begin{equation}
    \gield^{}_{\shape_{i,j}} \approx \big[\rho^{e}_{i} \big]^\mathrm{T} \big[ \weight^{}_{i,j} \big] \big[ \rho^{v}_{j} \big], 
    \label{eq_g1_matrix}
\end{equation}
The two arrays $[\rho^{e}_{i}]_{n^e_i\times 1}$ and $[\rho^{v}_{j}]_{n^v_j \times 1}$ are discrete representations of the two solids, i.e., the {\it design variables}. $n^e_i$ and $n^v_j$ denote the number of elements in solid $i$ and the number of vertices in solid $j$. 
The collision weight matrix (CWM) $[ \weight_{i,j}]_{n^e_i \times n^v_j}$ essentially captures the {\it pairwise correlations} between primal grid nodes of a moving grid and dual grid cells of a stationary grid, which depend solely on the relative motion of the grids and the grid structure. 

Assuming small deformations, the CWM {\it only} depends on the initial designs and needs to be computed only once. We leverage this property in the iterative co-design optimization in Section \ref{sec_method} to ensure scalability of the approach as computing the collision measures a few hundred times for arbitrarily complex shapes and motions can become computationally prohibitive. The CWM computes the aggregate collision between two discretized models over time. The time integral can be approximated using a Reimann sum:
\begin{align}
    \weight^{i, j} &\approx \epsilon^\dimm \delta_t \sum_{k = 1}^K \indic_{\cell_{i}} \left( \tau_{i,j}(t_k)\bx^{}_{j} \right), \label{eq_w12_approx} 
\end{align}
where we consider uniform discretization of the time period $[0, 1]$ into $K$ steps,  $t_k \triangleq (k-1)\delta_t$ for $k = 1, 2, \ldots, K$, assuming each time step $\delta_t$ is small enough to capture the motion trajectories accurately enough. 

The collision measure of \eq{eq_g1_matrix} can be generalized for multiple parts as:
\begin{align}
    \Gield^{}_{i} &\approx \sum_{i \neq j} \big[\rho^{e}_{i} \big]^\mathrm{T} \big[ \weight^{}_{i,j} \big] \big[ \rho^{v}_{j} \big], \label{eq_g1_matrix_multiple}
\end{align}

\begin{figure} [t!]
    \centering
    \begin{subfigure}[t]{0.8\linewidth}
    \centering
        \includegraphics[width=0.5\linewidth]{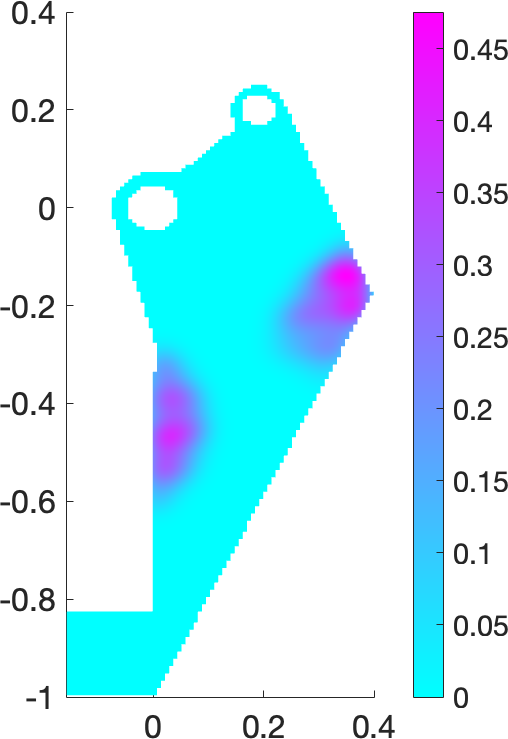}
        \caption{$\Gield^{}_{\shape_{1}}$}
        \label{fig_gripperOverallCollision}
    \end{subfigure}
    
    \begin{subfigure}[t]{.45\linewidth}
        \includegraphics[width=\linewidth]{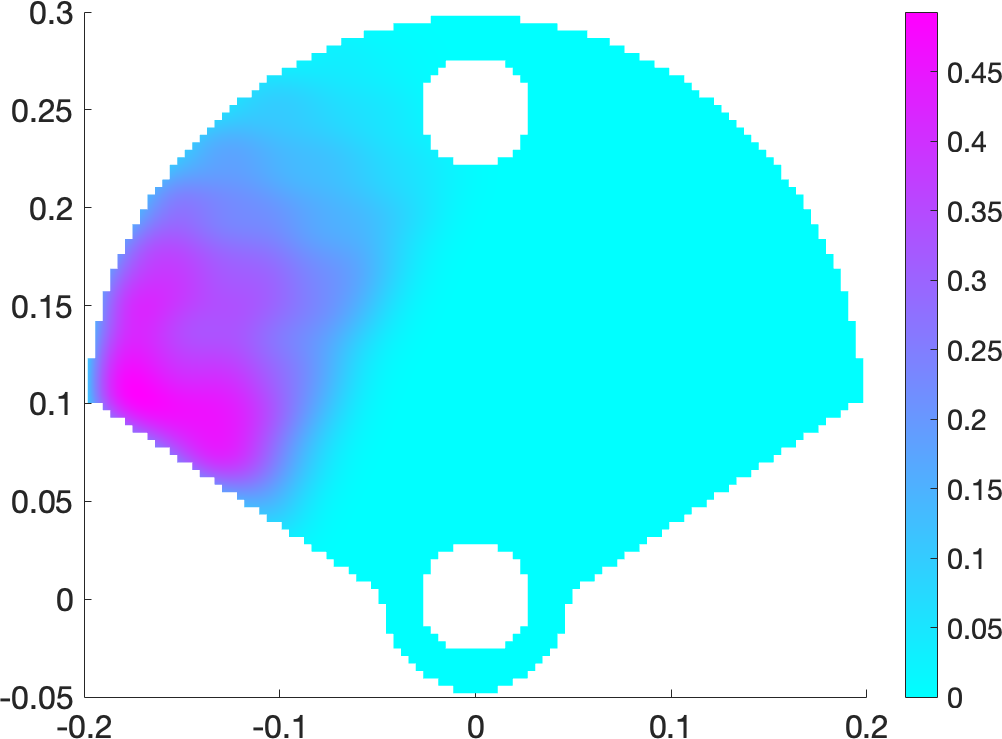}
        \caption{$\Gield^{}_{\shape_{2}} = \gield^{}_{\shape_{2,1}}$}
        \label{fig_camBC_1_OverallCollision}
    \end{subfigure}
        \begin{subfigure}[t]{.45\linewidth}
        \includegraphics[width=\linewidth]{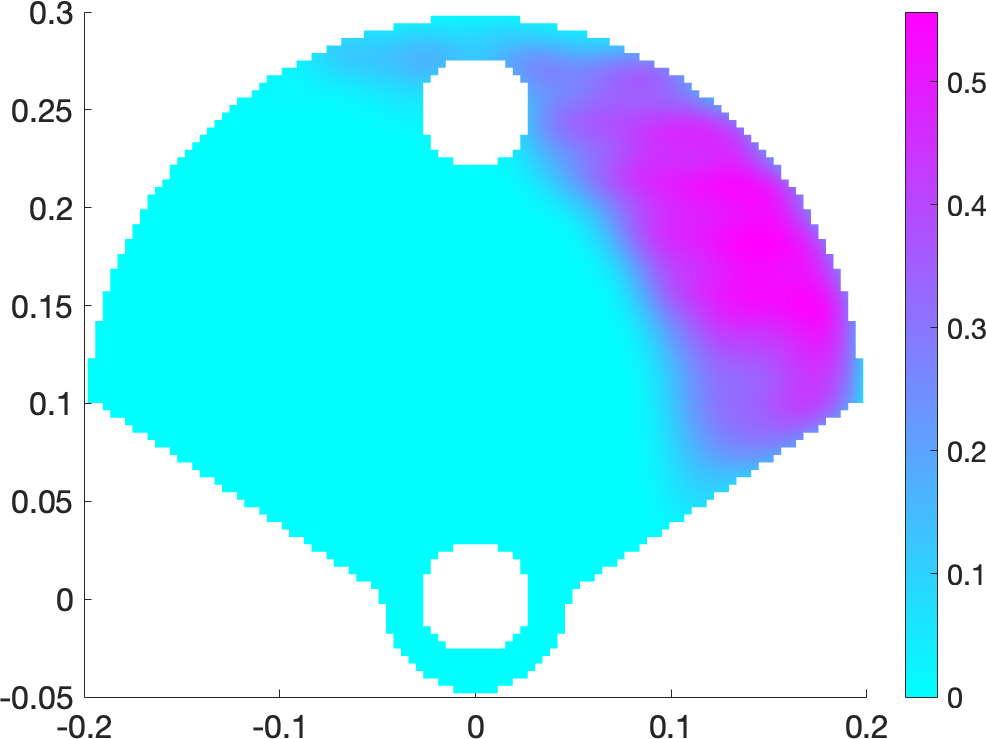}
        \caption{$\Gield^{}_{\shape_{3}} = \gield^{}_{\shape_{3,1}}$}
        \label{fig_camBC_2_OverallCollision}
    \end{subfigure}
    \centering
    \caption{Overall collision fields over the Gripper and two cams. Since the two cams only collides with the gripper, their overall collision field is similar to the their pair-wise collision fields with the gripper.}
    \label{fig_GripperCamsOverallCollision}
\end{figure}

\Cref{fig_GripperCamsOverallCollision} illustrates the overall collision fields for the gripper ($\Gield^{}_{\shape_{1}}$) and the two cams ($\Gield^{}_{\shape_{2}}$ and $\Gield^{}_{\shape_{3}}$). In this scenario, there is no collision between the cams and the overall collision is identical to their pair-wise collision field with only the gripper.

\subsection{Collision Sensitivity Analysis}

To enable efficient co-design optimization $[\rho^{e}_{i}]_{n^e_i\times 1}$  while avoiding collisions, the collision measures of \eq{eq_g1_matrix_multiple} must be differentiated with respect to the design variables $\big[\rho^{e}_{i} \big]$. The resulting discrete sensitivity fields are computed using a chain rule:
\begin{align}
    \left[ \frac{\partial \Gield^{}_{i}}{\partial \rho^{e}_{i}} \right] &\approx \sum_{i \neq j} \big[ \weight_{i,j} \big] \big[ \rho^{v}_{j} \big], \label{eq_dg_matrix}
\end{align}

Since we aim to remove material from regions with higher collision measure values, we define:
\begin{align}
    \TS_ {\Gield^{}_{i}} := 1- \left[ \frac{\partial \Gield^{}_{i}}{\partial \rho^{e}_{i}}\right]. \label{eq_TSg_matrix}
\end{align}

\begin{figure} [t!]
    \centering
    \begin{subfigure}[t]{0.8\linewidth}
    \centering
        \includegraphics[width=0.5\linewidth]{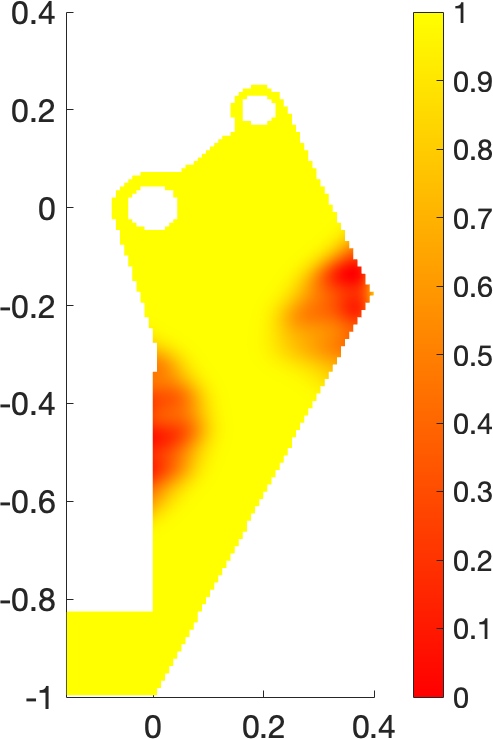}
        \caption{$\TS_ {\Gield^{}_{1}}$}
        \label{fig_gradGripper}
    \end{subfigure}
    
    \begin{subfigure}[t]{.45\linewidth}
        \includegraphics[width=\linewidth]{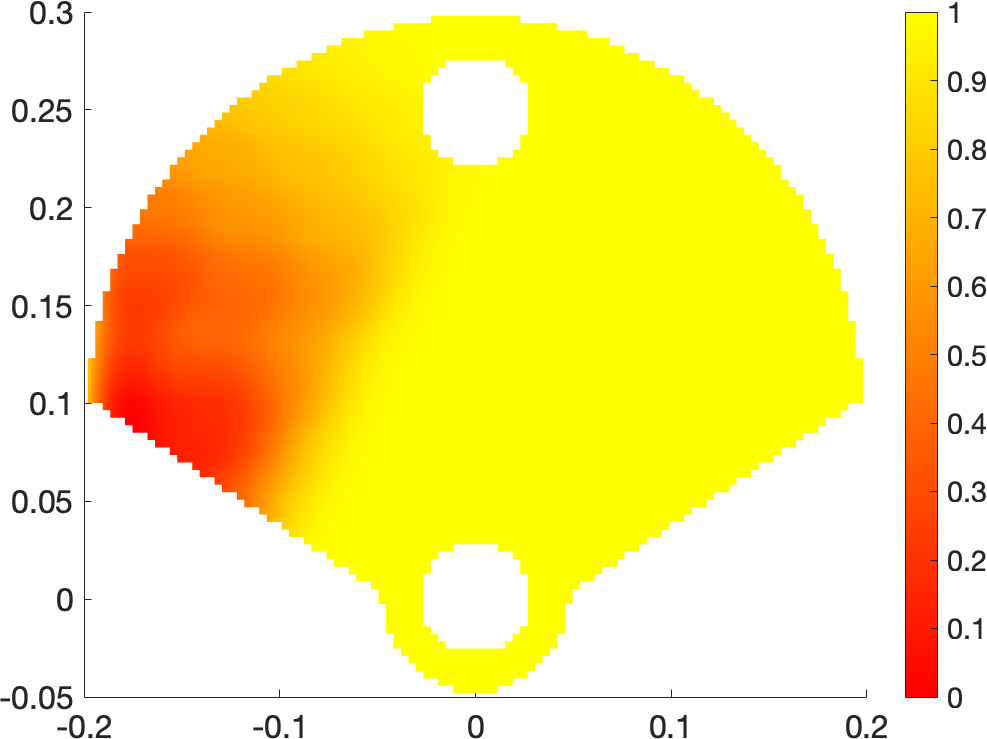}
        \caption{$\TS_ {\Gield^{}_{2}}$}
        \label{fig_gradCam1}
    \end{subfigure}
        \begin{subfigure}[t]{.45\linewidth}
        \includegraphics[width=\linewidth]{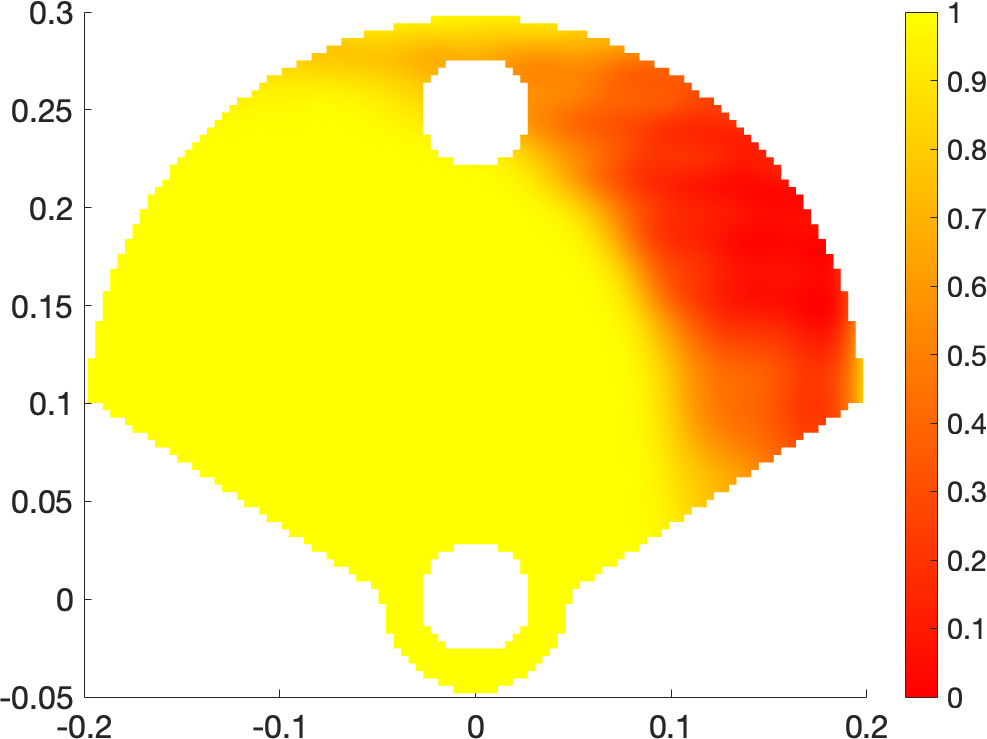}
        \caption{$\TS_ {\Gield^{}_{3}}$}
        \label{fig_gradCam2}
    \end{subfigure}
    \centering
    \caption{Collision gradient fields over the Gripper and two cams.}
    \label{fig_GripperCamsGrads}
\end{figure}

\Cref{fig_GripperCamsGrads} illustrates the overall sensitivity fields for the assembly of \Cref{fig_GripperCamsBC}.

\section{Co-Design Optimization} \label{sec_method}

In this section, we present the TO formulation to enable co-design of moving components of an assembly. To this end, we employ the Pareto-tracing TO (PareTO) \cite{suresh2010199} to explore the trade-offs between competing objectives, here compliance and collision avoidance. At every step, we conduct a fixed-point iteration to find a local optimum at each volume fraction. The optimization terminates after all collisions are resolved. 

Mathematically, we formulate the optimization problem as finding 
$\big[\rho^{e}_{i}\big] \subseteq \Omega_i,~i=1,\cdots,N $ 
to:

\begin{equation}
	\text{Find}~ \big[\rho^{e}_{i}\big] \subseteq \Omega_i:
	\begin{dcases} 
		\text{select target}~ \bar{V}_i^\mathrm{targ} \in (0, 1],~ \forall i \\
		\text{ILI:}
		\begin{dcases}
                \mathop{\text{minimize}}\limits_{ \rho^{e}_{i}} \quad
    		f( \rho^{e}_{i}),\\ %\label{generalForm} \\
    		\text{s.t.}\quad 
                 \big[\bK_i\big] \big[\bu_i\big] =  \big[\bff_i\big],\\% \label{generalForm_fea}\\
                 \quad \quad ~\bar{V_i} = \bar{V}_i^\mathrm{targ},\\ %\label{volCons}\\
                   \quad \quad ~{\Gield}_{i} = 0, ~ \forall i.  %\label{generalForm_const1} 
		\end{dcases}
	\end{dcases} \label{eq_ParetoProblem}
\end{equation} 
Generally, we begin the optimization process with all solid designs are gradually reduce the target volume fraction through $\bar{V}_i^\mathrm{targ} \leftarrow \bar{V}_i^\mathrm{targ} - \delta_{v_i}$. To be able to effectively explore the design spaces (from extreme cases where material is removed only from one part to intermediate scenarios where material is removed from all part with various levels of aggressiveness), we define $\delta_{v_i} \triangleq \gamma \delta_v^{max}$; where $\delta_v^{max}$ is the maximum allowable volume decrement for all parts and the hyper-parameter $0 \le\gamma\le 1$ is used to control the decrement aggressiveness for each component.  

The inner-loop optimization can be expressed as local minimization of the
Lagrangian defined as:
\begin{equation}
\begin{split}
	\mathcal{L}_i &:= [\bff_i]^\mathrm{T} [\bu_i] + [\mu_i]^\mathrm{T} \Big(
	[\bK_i][\bu_i] - [\bff]\Big) \\
       & +\lambda_{v_i}(\bar{V}_i - \bar{V}_i^\mathrm{targ})  + \lambda_{g_i}{\Gield}_{i}. \label{eq_Lag}
\end{split}
\end{equation} 
The Karush--Kuhn--Tucker (KKT) conditions \cite{wright1999numerical} for this
problem are given by $\nabla_{\rho^{e}_{i}} \mathcal{L}= 0$ in which the gradient is
defined by partial differentiation with respect to the independent variables;
namely, the design variables used to represent $\big[\rho^{e}_{i}\big]$ and the Lagrange
multipliers $\mu_i$, $\lambda_{v_i}$ and $\lambda_{g_i}$. The latter simply encodes the constraints into $\nabla_{\rho^{e}_{i}} \mathcal{L}_= 0$:
\begin{align}
	\frac{\partial~}{\partial \mu_i} \mathcal{L}_i &=
	[\bK_i][\bu_i] - [\bff] := [0],\\
	\frac{\partial~}{\partial \lambda_{v_i}} \mathcal{L}_i &= (\bar{V}_i -
	\bar{V}_i^\mathrm{targ}) := 0, \\
 	\frac{\partial~}{\partial \lambda_{g_i}} \mathcal{L}_i &=
	{\Gield}_{i} := 0,
\end{align}
Next, let the prime symbol $(\cdot)'$ represent the generic (linear)
differentiation of a function with respect to $\rho^{e}_{i}$, we obtain (via chain
rule):
\begin{equation}
\begin{split}
	\mathcal{L}_i' &= [\bff_i]^\mathrm{T} [\bu_i'] + [\mu_i]^\mathrm{T} \Big([\bK_i][\bu_i]\Big)' \\
  & + \lambda_{v_i}\bar{V}_i' + \lambda_{g_i}{\Gield}_{i}',
 \label{eq_chain_rule} 
\end{split}
\end{equation} 
Using the adjoint method \cite{bendsoe2003topology}, we have $[\mu_i] := -[\bK_i]^{-1}[\bff_i]$. Thus, \eq{eq_chain_rule} reduces to:
\begin{equation}
\begin{split}
 \mathcal{L}_i' = \lambda_{v_i} \bar{V}_\Omega'  \underbrace{-[\bu_i]^\mathrm{T} [\bK_i'][\bu_i]}_{\text{compliance sensitivity}} + \lambda_{g_i}\underbrace{\Gield_{i}'}_{\text{collision sensitivity}}
 \end{split}\label{eq_Lag_prime}
\end{equation}

\begin{figure} [t!]
    \centering
\includegraphics[width=\linewidth]{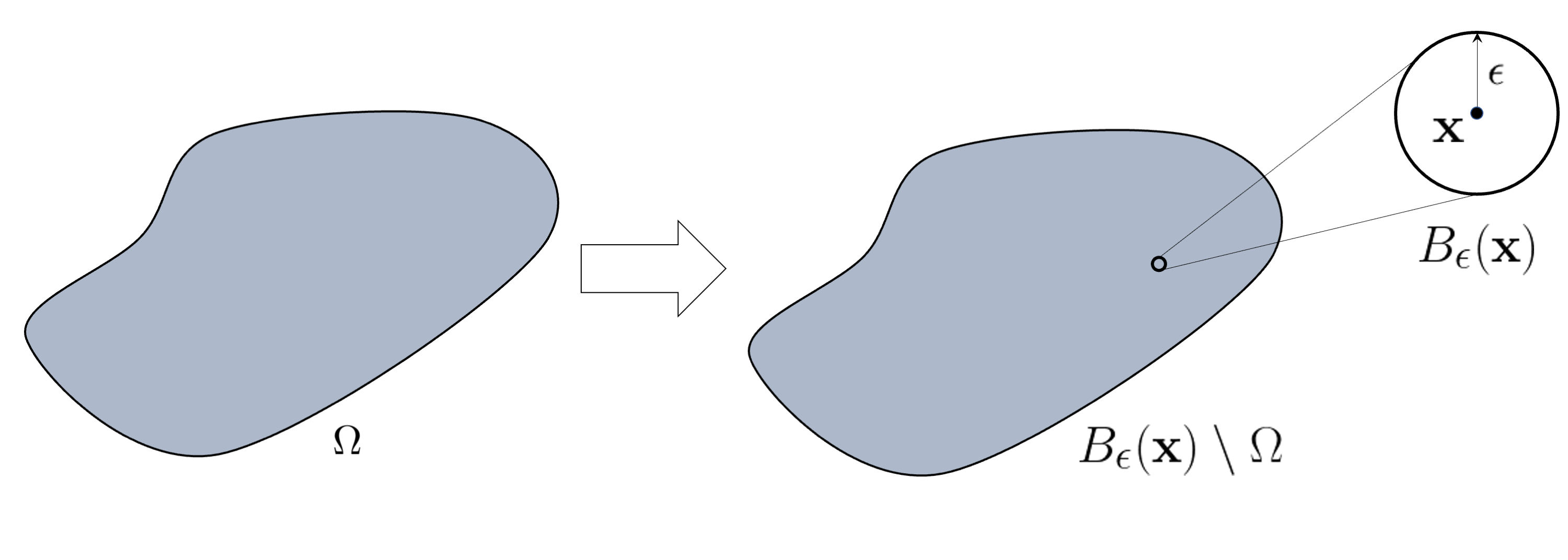}
        \caption{Computing the topological sensitivity by hypothetically perturbing the design $\Omega$ at point $\bx$ via a small inclusion of radius $\epsilon$.}
    \label{fig_TS}
\end{figure}

In this work, we use the TSF interpretation for the compliance sensitivity in \eq{eq_Lag_prime}. The topological sensitivity ${\mathcal T(\bx)}$ is defined as the ratio of the first-order change in the objective $f$ to the area (or volume) of the \textit{hypothetical} infinitesimal hole $B_\epsilon(\bx)$ in the design at point $\bx$, as illustrated in \ref{fig_TS}. Mathematically,
\begin{equation} \label{eq_TSF2D}
	{\rm {\mathcal T}}_\bx(f){\rm \; }\equiv \mathop{\lim }\limits_{\epsilon\to 0^+} \frac{f (\Omega- B_\epsilon(\bx))- f(\Omega) }{-{\vol}(B_\epsilon(\bx))},
\end{equation}
where $\vol(\cdot)$ denotes the volume (or area in 2D) of the inclusion. 

The closed-form expression for the topological derivative of compliance for plane-stress problems in 2D is \cite{novotny2012topological}:
\begin{equation} \label{eq_CompTSF2D_plainStress}
	{\rm {\mathcal T}}_\bx(f){\rm \; }=\frac{4}{1+\nu } \sigma:\varepsilon-\frac{1-3\nu }{1-\nu ^{2} } tr(\sigma )tr(\varepsilon),
\end{equation}
where $\nu$, $\sigma$, and $\varepsilon$ respectively denote the Poisson's ratio, stress tensor, and strain tensor at every point $\bx$.
Similar closed-form expressions have also been derived for 3D linear elasticity problems \cite{novotny2007topological}.

Neglecting the topological sensitivity for volume (constant everywhere), we have the following expression for the overall TSF for each part with $\big[\rho^{e}_{i}\big]$ as design variables:
\begin{equation}
\begin{split}
 \hat{\TS}_i := \mathcal{T}(\mathcal{L}_i) = \mathcal{T}_i+ \lambda_{g_i}  \TS_ {\Gield^{}_{i}}
 \end{split}\label{eq_TSF}
\end{equation}
At every step, we perform FEA, compute sensitivity field, and reject a few elements with \textit{lowest} compliance sensitivity value and repeat until we satisfy the collision-free constraint. Consider now a domain $S_i^{\tau_i}$ as the set of all points in part $i$ where the sensitivity field exceeds the value $\tau_i$, defined per:
\begin{equation} \label{eq_levelset}
	S_i^{\tau} \triangleq \{ \bx ~\big\lvert~   \hat{\TS}_\bx >\tau_i\}
\end{equation}

The value of $\lambda_{g_i}$ can be adaptively adjusted throughout the optimization considering the values of compliance and violation of collision constraints. Here, we prescribe the value of $\lambda_{g_i}$ to explore the feasible design space and provide more insight on the impact of collision-avoidance constraint.  

\begin{figure} [t!]
    \centering
    \begin{subfigure}[t]{0.6\linewidth}
    \centering
        \includegraphics[width=\linewidth]{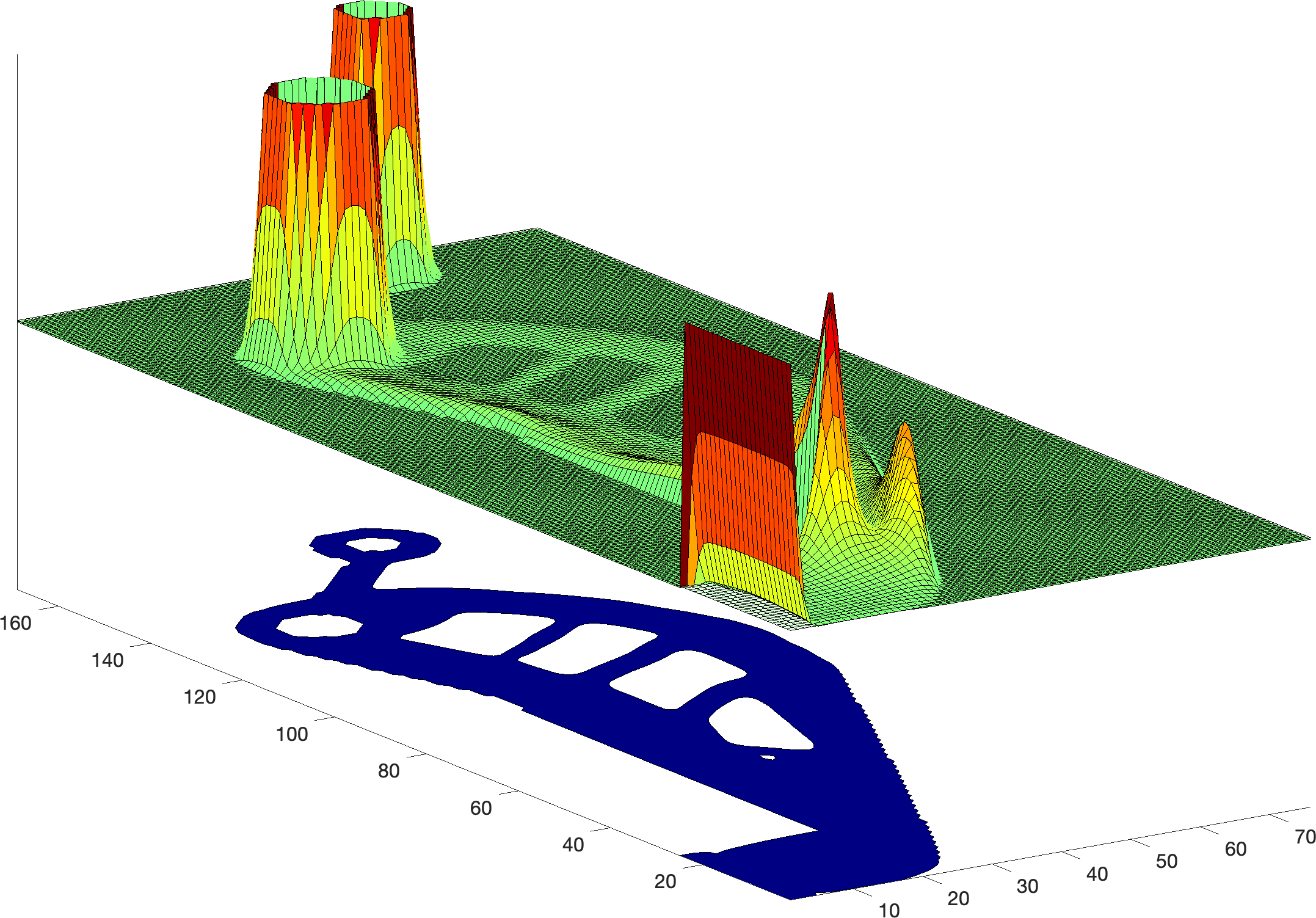}
        \caption{Gripper compliance TSF}
        \label{fig_GripperTSFWt0}
    \end{subfigure}
    \begin{subfigure}[t]{0.35\linewidth}
    \centering
        \includegraphics[width=0.8\linewidth]{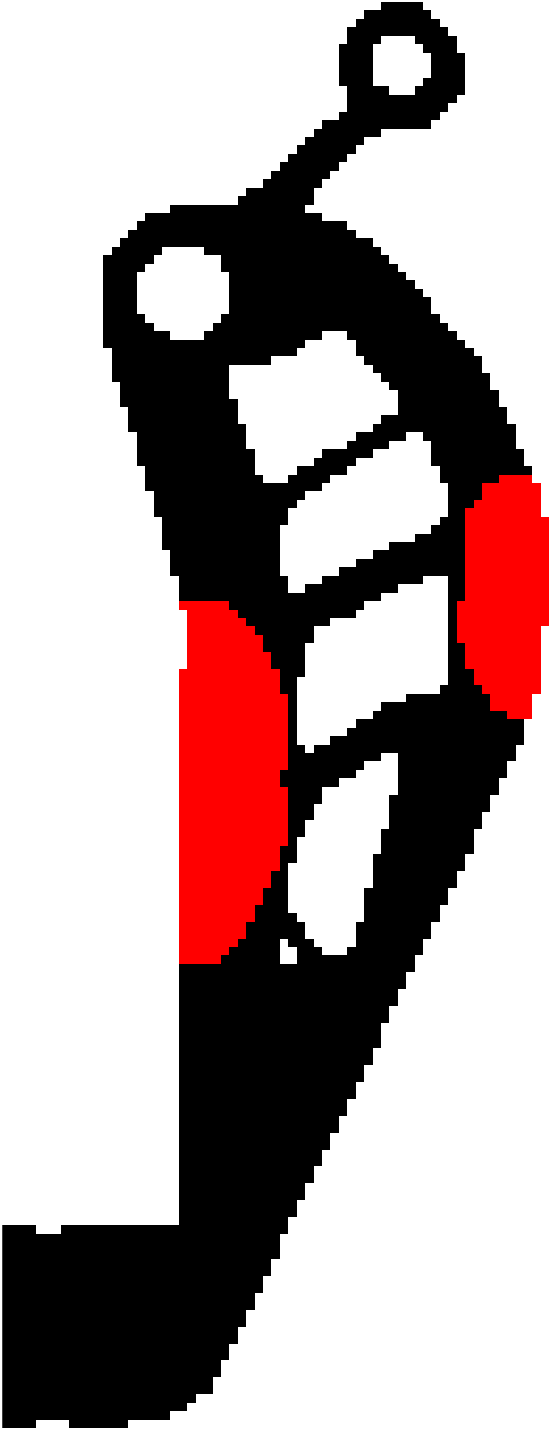}
        \caption{Optimized gripper at 0.5 volume fraction with \textcolor{red}{\textbf{colliding regions}} shown in red.}
        \label{fig_GripperOptWt0}
    \end{subfigure}

     \begin{subfigure}[t]{.6\linewidth}
        \includegraphics[width=\linewidth]{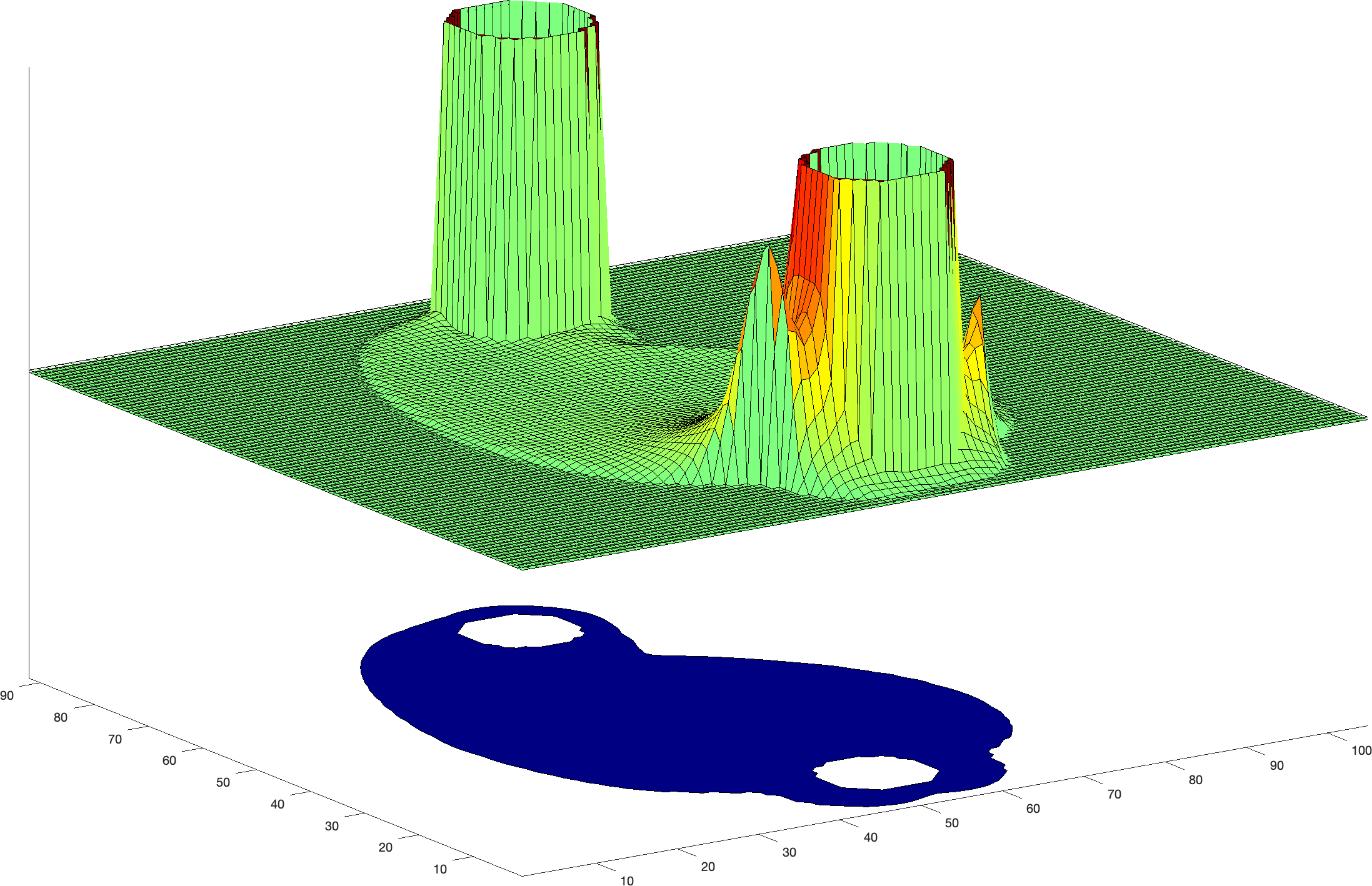}
        \caption{Cam \#1 compliance TSF.}
        \label{fig_cam1TSFWt0}
    \end{subfigure}
    \begin{subfigure}[t]{.35\linewidth}
        \includegraphics[width=0.9\linewidth]{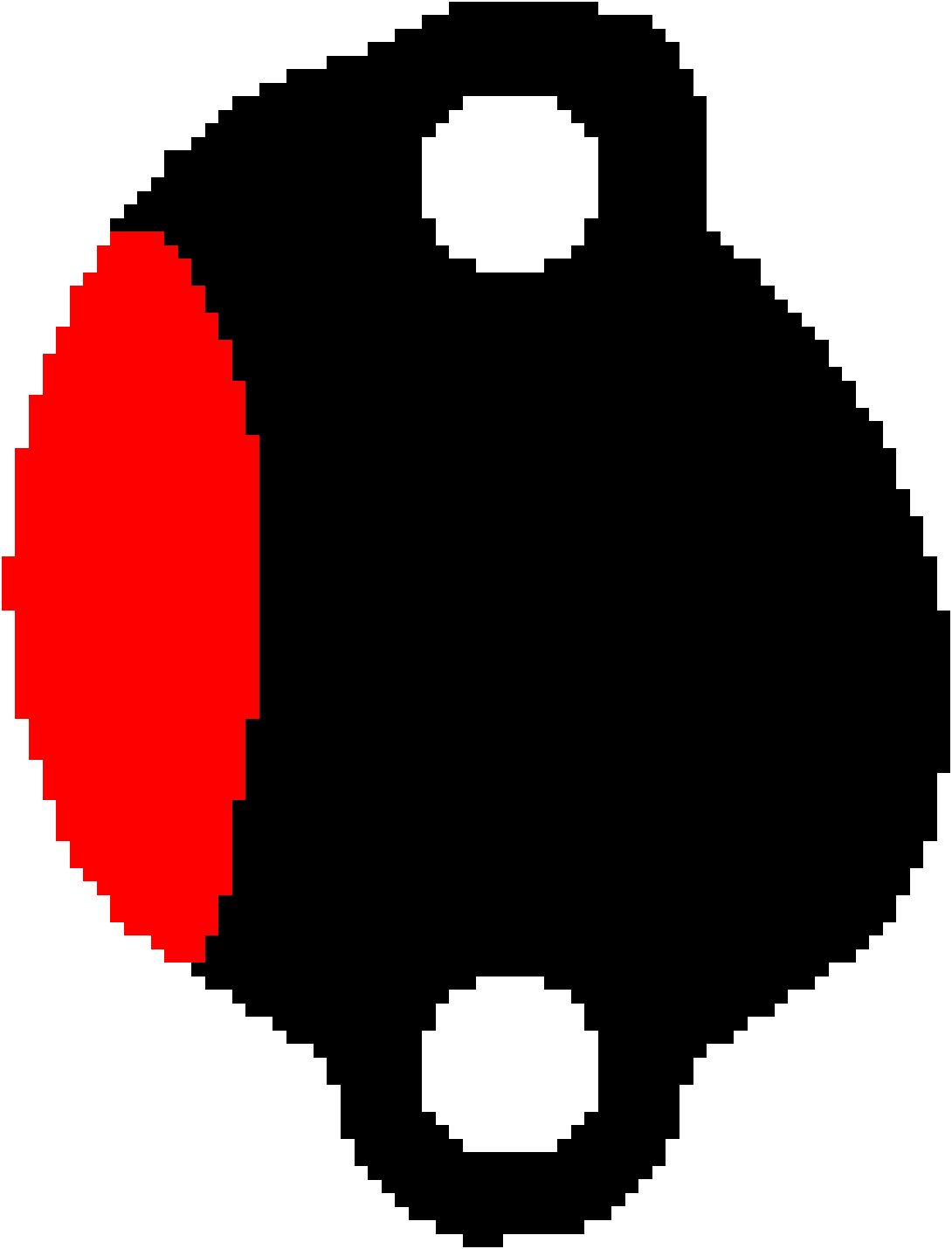}
        \caption{Cam \#1 colliding regions.}
        \label{fig_cam1OptWt0}
    \end{subfigure}

    \begin{subfigure}[t]{.6\linewidth}
        \includegraphics[width=\linewidth]{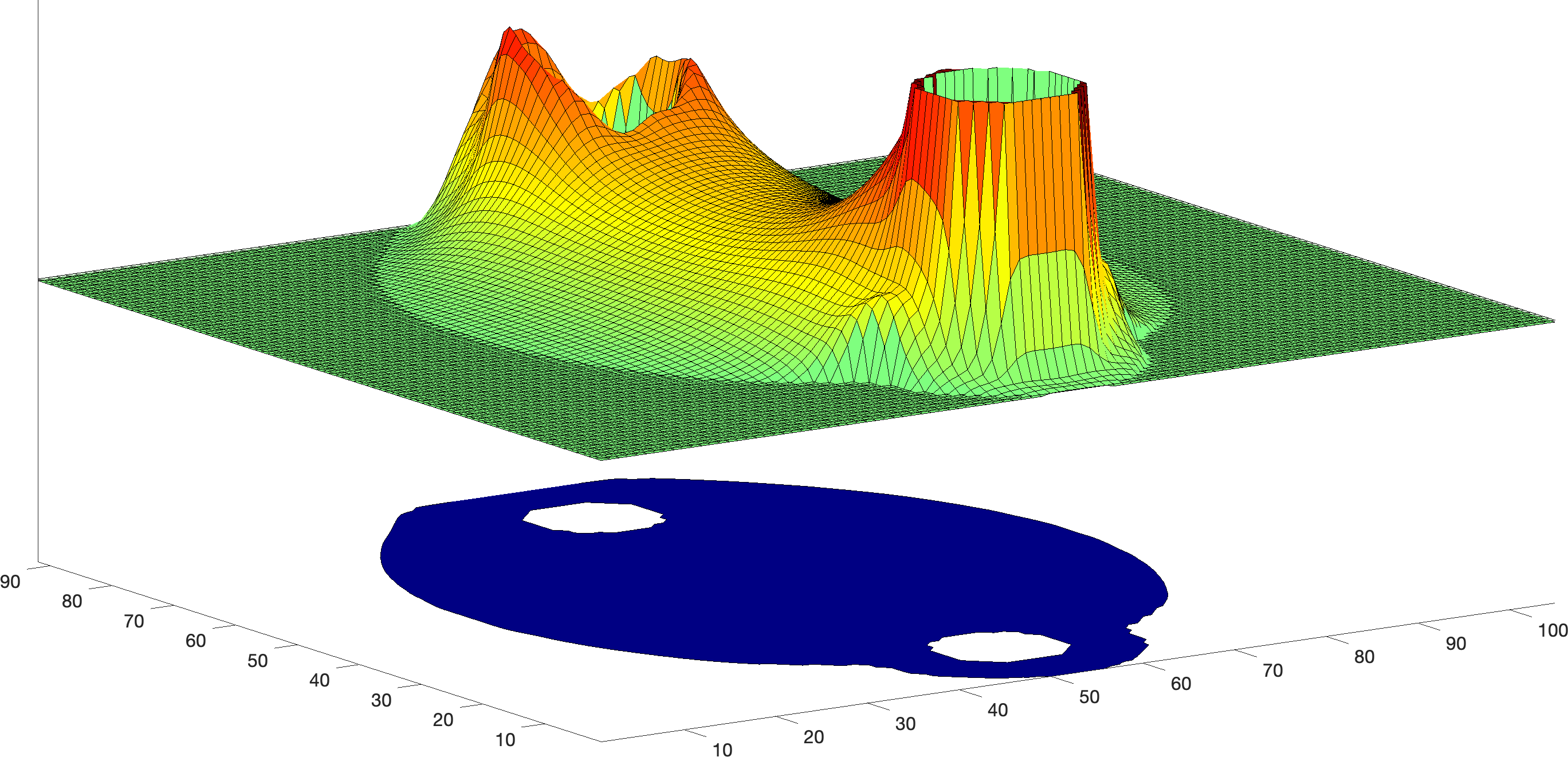}
        \caption{Cam \#2 compliance TSF.}
        \label{fig_cam2TSFWt0}
    \end{subfigure}
    \begin{subfigure}[t]{.35\linewidth}
        \includegraphics[width=0.85\linewidth]{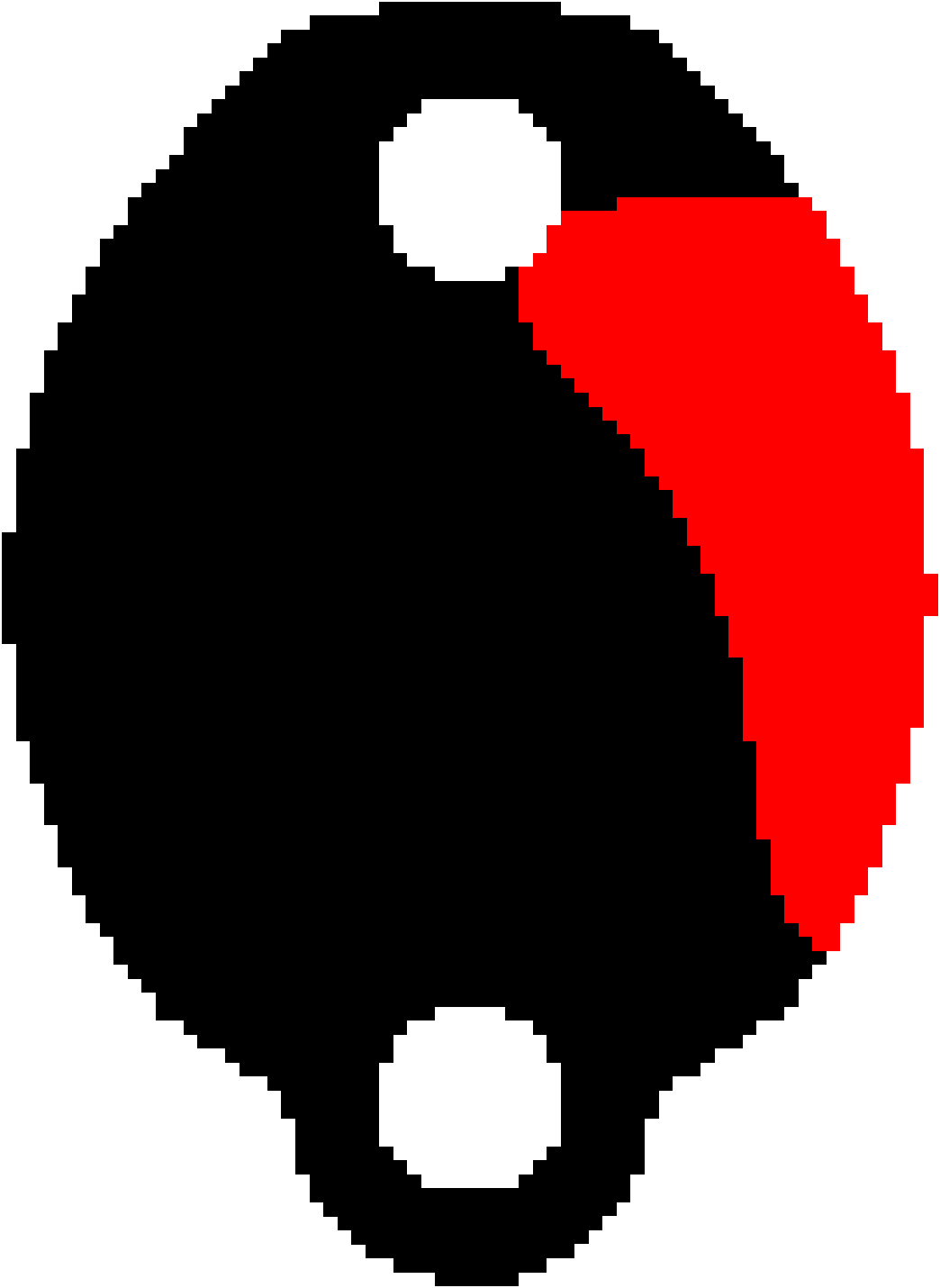}
        \caption{Cam \#2 colliding regions.}
        \label{fig_cam2OptWt0}
    \end{subfigure}
    \centering
    \caption{Optimized designs at 0.7 volume fraction for all components \textit{without} considering collision avoidance constraint ($\lambda_{g_i} = 0, \forall i$). }
    \label{fig_GripperCamsOptWt0}
\end{figure}

\Cref{fig_GripperCamsOptWt0} illustrates the optimized designs for the gripper and the two cam structures at volume fraction $v_i=0.5$ (for all) \textit{without} incorporating the collision avoidance constraint. We consider $\delta_v^{max} = 0.01$ and $\gamma_i=1$ for all parts. Observe that in by solely optimizing the components with respect to compliance, the parts collide with each other on multiple occasions throughout their prescribed trajectories (red regions in Figures \ref{fig_GripperOptWt0}, \ref{fig_cam1OptWt0}, and \ref{fig_cam2OptWt0}).

% Part1, Objective: 1.05, Volume: 0.70, Collision Volume: 0.44, Deformation: 0.00013153, von Mises: 836.4172
% Part2, Objective: 1.01, Volume: 0.70, Collision Volume: 0.33, Deformation: 1.1177e-06, von Mises: 13.8609
% Part3, Objective: 1.00, Volume: 0.70, Collision Volume: 0.37, Deformation: 6.0449e-08, von Mises: 1.7304

\Cref{alg_PareTO} provides a description of the approach.
\begin{algorithm}[ht!] 
	\caption{Co-Design via Collision-Aware PareTO }
	\begin{algorithmic}[1]
		\Procedure{PareTO}{$ \Omega_i, V^*, \delta v, \gamma_i$}
		\State $[\rho^e_i] \leftarrow \Omega_i$   \Comment{Initialize at volume fraction 1.0}
		\State $V_i \leftarrow$ \Call{EvaluateVolume}{$[\rho^e_i]$}
            \State $\lambda_{g_i} \leftarrow \gamma_i \delta v$
		\While{$V_i > V^*$ \text{and} $ \Gield_i > 0, ~\forall i$}
		\State $ v_i \leftarrow V_i - \delta v_i$ 
    	\State $\bu_i \leftarrow$ \Call{SolveFEA}{$[\rho^e_i],\bK_i,\bff_i$}  
            \State $f \leftarrow$ \Call{EvaluateCompliance}{$[\rho^e_i],\bu$}
            \State $\Gield_i \leftarrow$ \Call{EvaluateCollision}{$[\rho^e_i],[\rho^v_j]~ \forall~i,j$} 
            
		\State $ \delta f \leftarrow 1$
		\While{$ \delta f > \epsilon$ }
  		
            \State $\Gield_i \leftarrow$ \Call{EvaluateCollision}{$[\rho^e_i],[\rho^v_j],~ \forall~i,j$} 
		\State $\mathcal{T} \leftarrow$ \Call{ComplianceGradient}{$[\rho^e_i],\bu_i$} 
            \State $ \TS_ {\Gield^{}_{i}}
            \leftarrow$ \Call{CollisionGradient}{$[\rho^e_i],[\rho^v_j]$}
            
            \State $\hat{\TS} \leftarrow \mathcal{T}_i+ \lambda_{g_i} \left[ \frac{\partial \Gield^{}_{i}}{\partial \rho^{e}_{i}} \right]$
            
		\State $\tau_i \leftarrow$ \Call{FindThreshold}{$[\rho^e_i],\hat{\TS},\delta v$} 
		\State $[\rho^e_i]^* \leftarrow$ \Call{ExtractLevelSet}{$\hat{\TS},\tau $}
		\State $\bu_i \leftarrow$ \Call{SolveFEA}{$[\rho^e_i]^{*},\bK_i,\bff_i$}  
		\State $ f^{*} \leftarrow$ \Call{EvaluateCompliance}{$[\rho^e_i]^*,\bu$}
		\State $ \delta f \leftarrow |f^* - f|$
		\State $f \leftarrow f^*$
		\EndWhile
		\State $[\rho^e_i] \leftarrow [\rho^e_i]^*$
		\State $V_i \leftarrow$ \Call{EvaluateVolume}{$[\rho^e_i]$}
		\EndWhile
		\State \textbf{return} $[\rho^e_i]$
		\EndProcedure
	\end{algorithmic} \label{alg_PareTO}
\end{algorithm}
\section{Results} \label{sec_results}
In this section, we demonstrate the effectiveness of the proposed approach through a few examples with various design complexities and motions. For all examples, we assume Young's modulus $E = 1~GPa$, Poisson's ratio $\nu = 0.3$, and maximum volume decrement $\delta v = 0.025$. All examples are run on a MacBook Pro M1 Max with 32 GB of memory.

\subsection{Cam and Follower}

\begin{figure} [h!]
    \centering
    \includegraphics[width=0.4\linewidth]{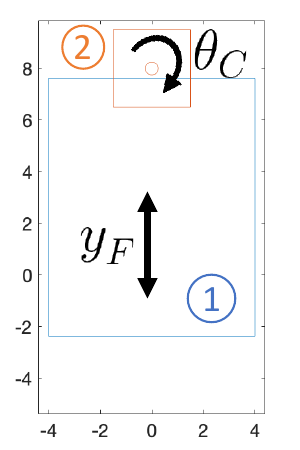}
    \caption{Cam and follower initial configuration.}
    \label{fig_camFollowerConfig}
\end{figure}

\begin{figure} [h!]
    \centering
    \begin{subfigure}[t]{.45\linewidth}
        \includegraphics[width=0.9\linewidth]{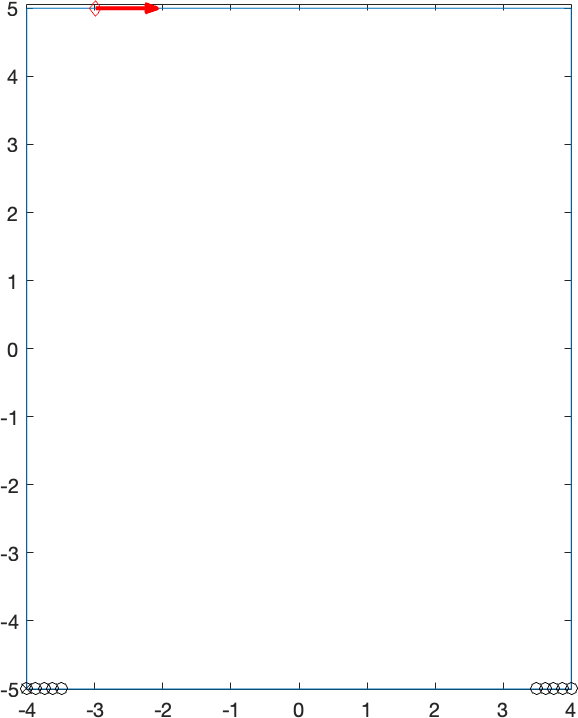}
        \caption{Follower.}
        \label{fig_followerBC}
    \end{subfigure}
    \begin{subfigure}[t]{.3\linewidth}
        \includegraphics[width=0.9\linewidth]{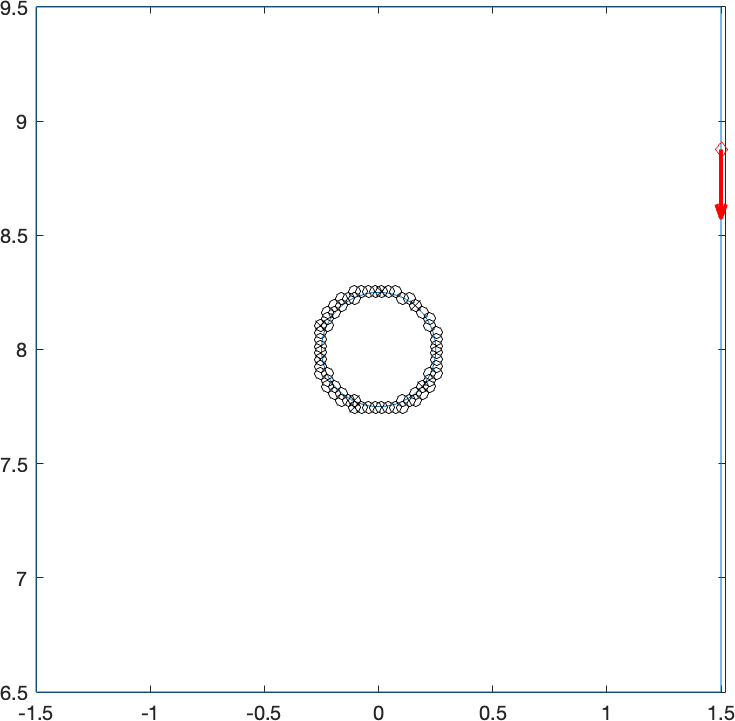}
        \caption{Square cam.}
        \label{fig_squareCamBC}
    \end{subfigure}
    \centering
    \caption{Cam and follower boundary conditions.}
    \label{fig_camFollowerBC}
\end{figure}

\begin{figure} [h!]
    \centering
    \begin{subfigure}[t]{.45\linewidth}
        \includegraphics[width=\linewidth]{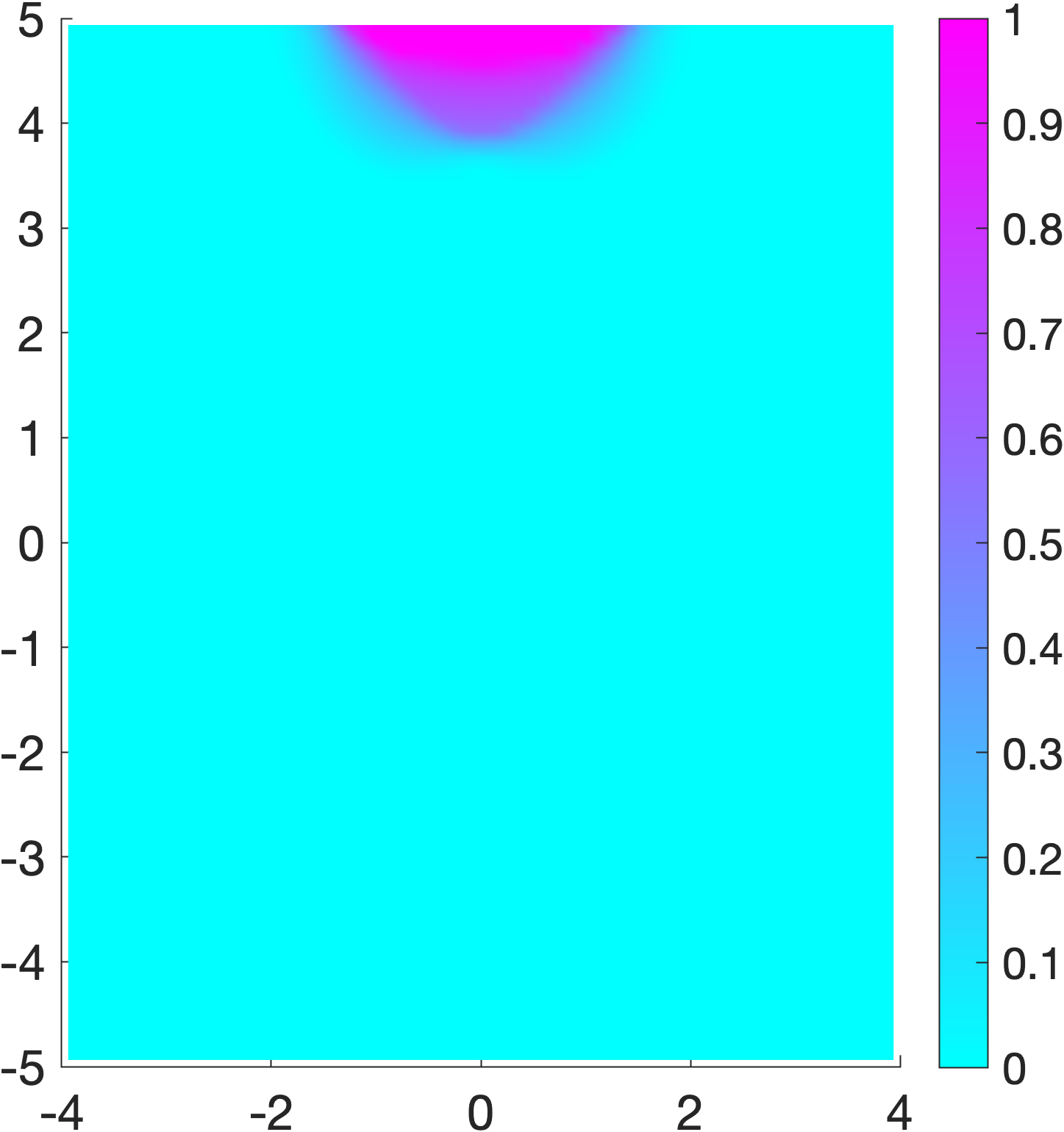}
        \caption{Follower.}
        \label{fig_followerCollField0}
    \end{subfigure}
    \begin{subfigure}[t]{.3\linewidth}
        \includegraphics[width=\linewidth]{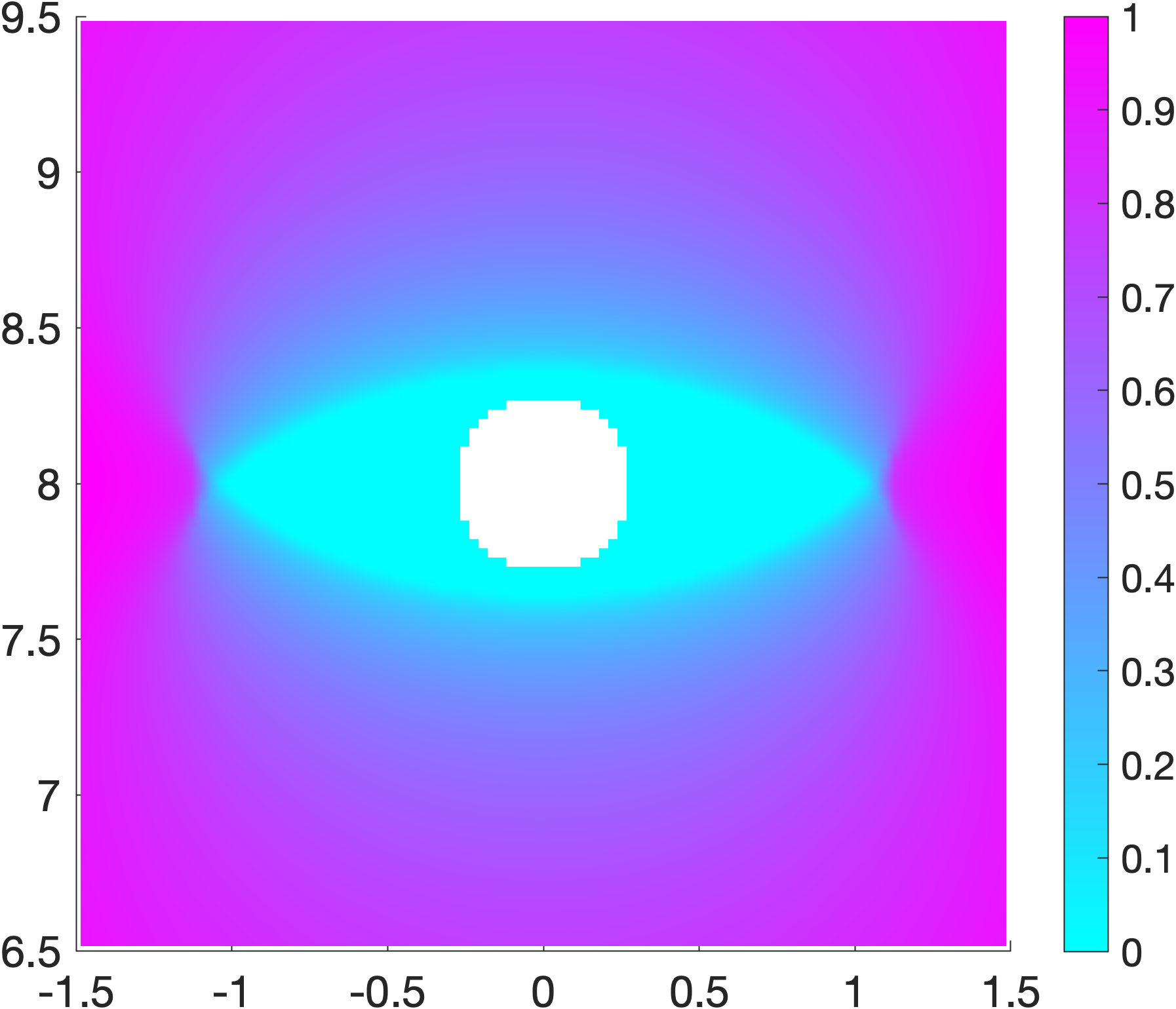}
        \caption{Square cam.}
        \label{fig_camCollField0}
    \end{subfigure}
    \centering
    \caption{Initial collision measure fields for the cam-follower system.}
    \label{fig_camFollowerCollFields0}
\end{figure}

\begin{figure} [h!]
    \centering
    \begin{subfigure}[t]{.45\linewidth}
        \includegraphics[width=\linewidth]{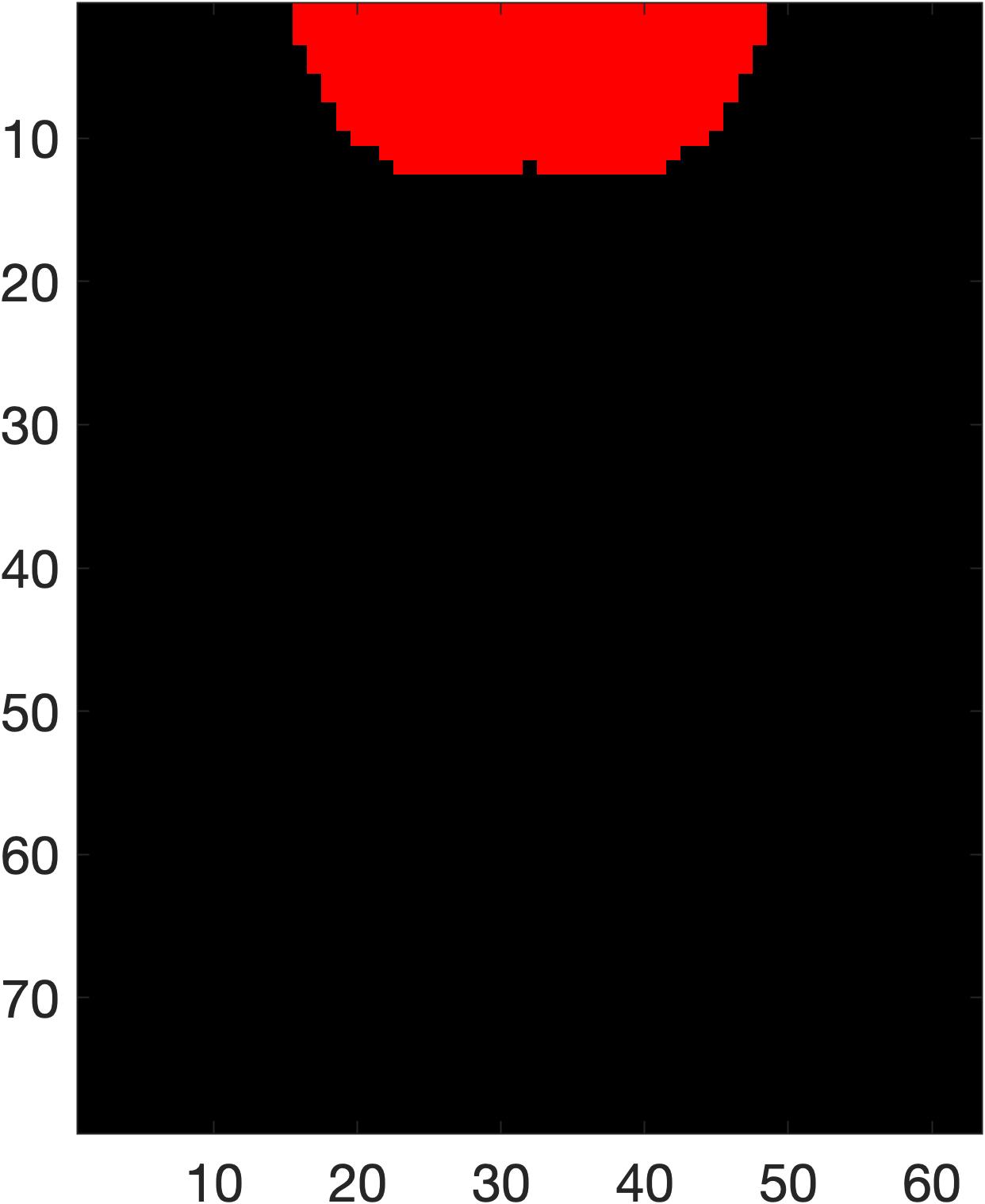}
        \caption{Follower.}
        \label{fig_followerColl0}
    \end{subfigure}
    \begin{subfigure}[t]{.3\linewidth}
        \includegraphics[width=\linewidth]{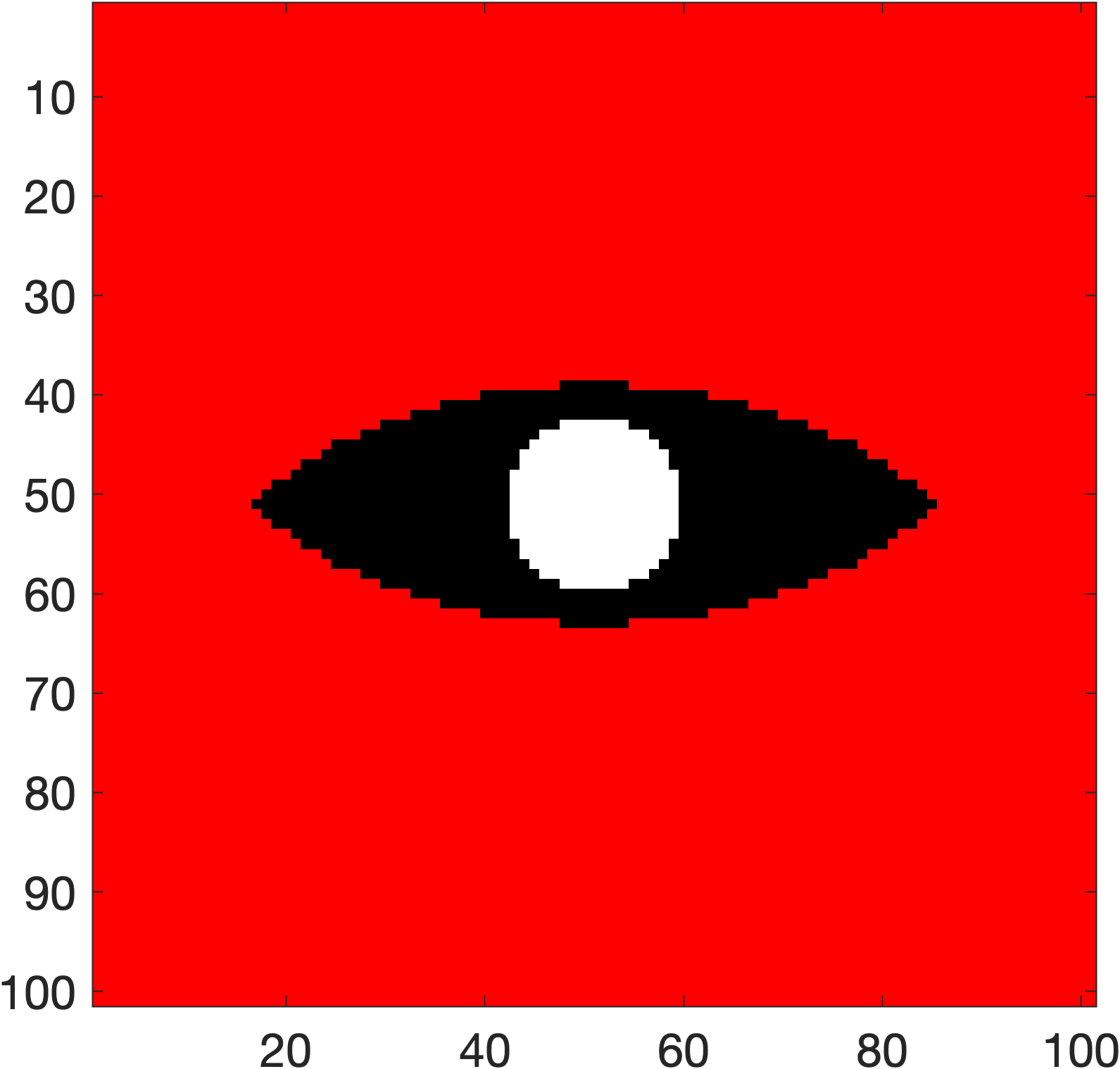}
        \caption{Square cam.}
        \label{fig_camColl0}
    \end{subfigure}
    \centering
    \caption{Initial colliding regions for the cam-follower system.}
    \label{fig_camFollowerCollRegions}
\end{figure}

\begin{figure} [h!]
    \centering
    \begin{subfigure}[t]{.45\linewidth}
    \centering
        \includegraphics[width=0.9\linewidth]{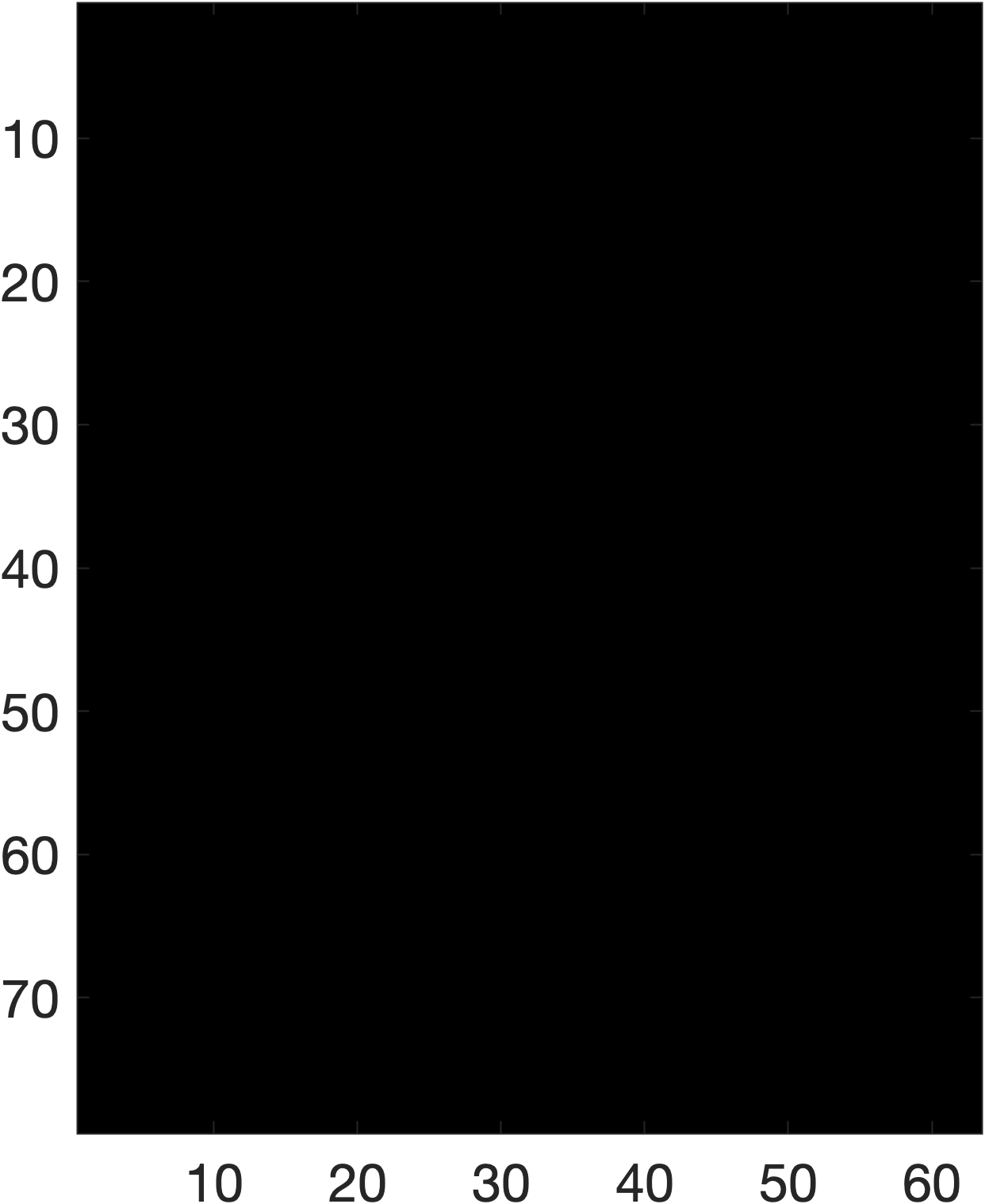}
        \caption{Follower with $\lambda_{g_1} = 1$ and $\gamma_1 = 0$.}
        \label{fig_followerOpt_a}
    \end{subfigure}\hfill
    \begin{subfigure}[t]{0.45\linewidth}
    \centering
        \includegraphics[width=0.7\linewidth]{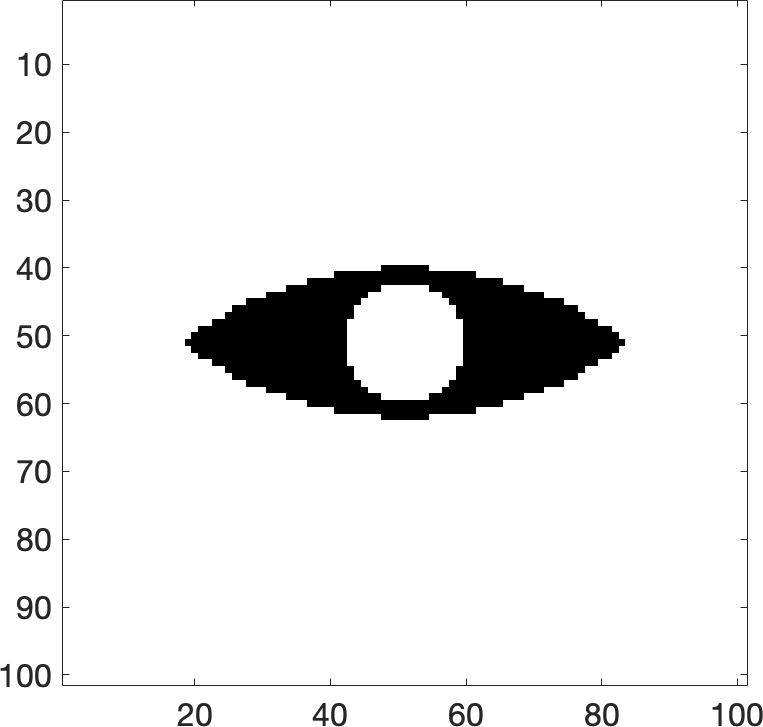}
        \caption{Cam with $\lambda_{g_2} = 1$ and $\gamma_2 = 1$. No compliance is considered and the design is infeasible for the prescribed boundary conditions.}
        \label{fig_camOpt_b}
    \end{subfigure}

        \begin{subfigure}[t]{.45\linewidth}
        \centering
        \includegraphics[width=0.9\linewidth]{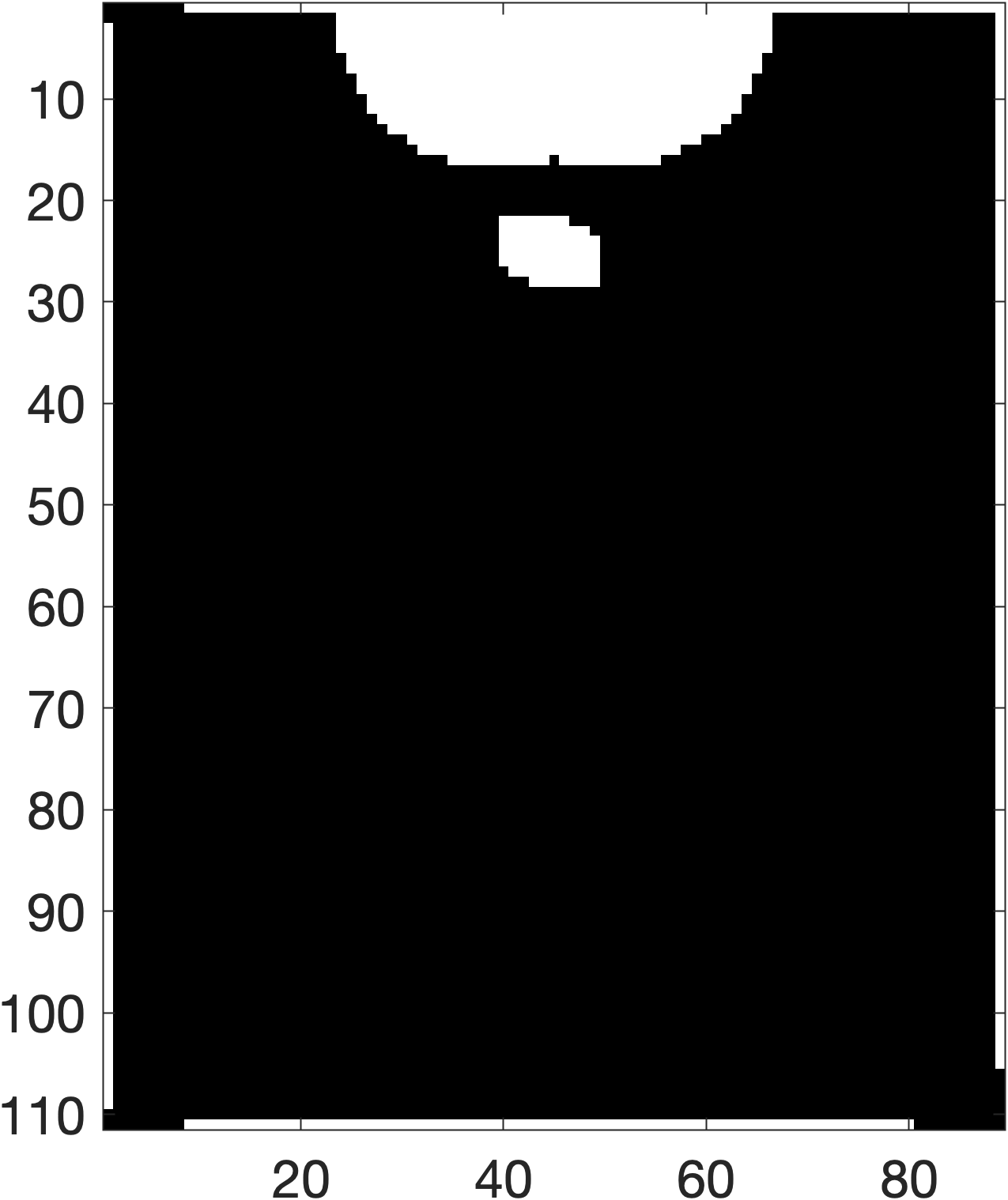}
        \caption{Follower with $\lambda_{g_1} = 0.2$ and $\gamma_1 = 1$.}
        \label{fig_followerOpt_c}
    \end{subfigure}\hfill
    \begin{subfigure}[t]{0.45\linewidth}
    \centering
        \includegraphics[width=0.7\linewidth]{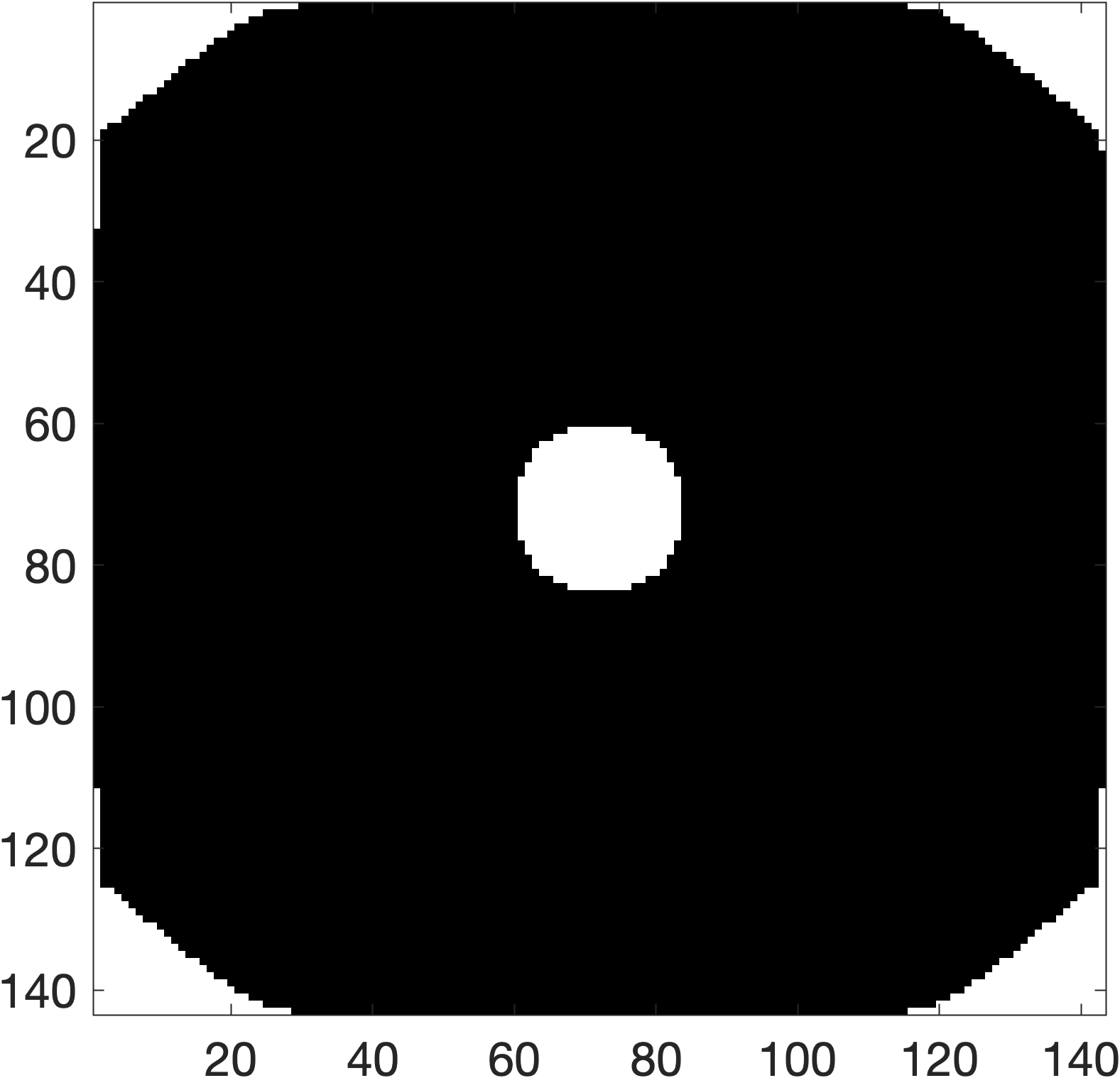}
        \caption{Cam with $\lambda_{g_2} = 0.2$ and $\gamma_2 = 0.5$.}
        \label{fig_camOpt_d}
    \end{subfigure}

            \begin{subfigure}[t]{.45\linewidth}
            \centering
        \includegraphics[width=0.9\linewidth]{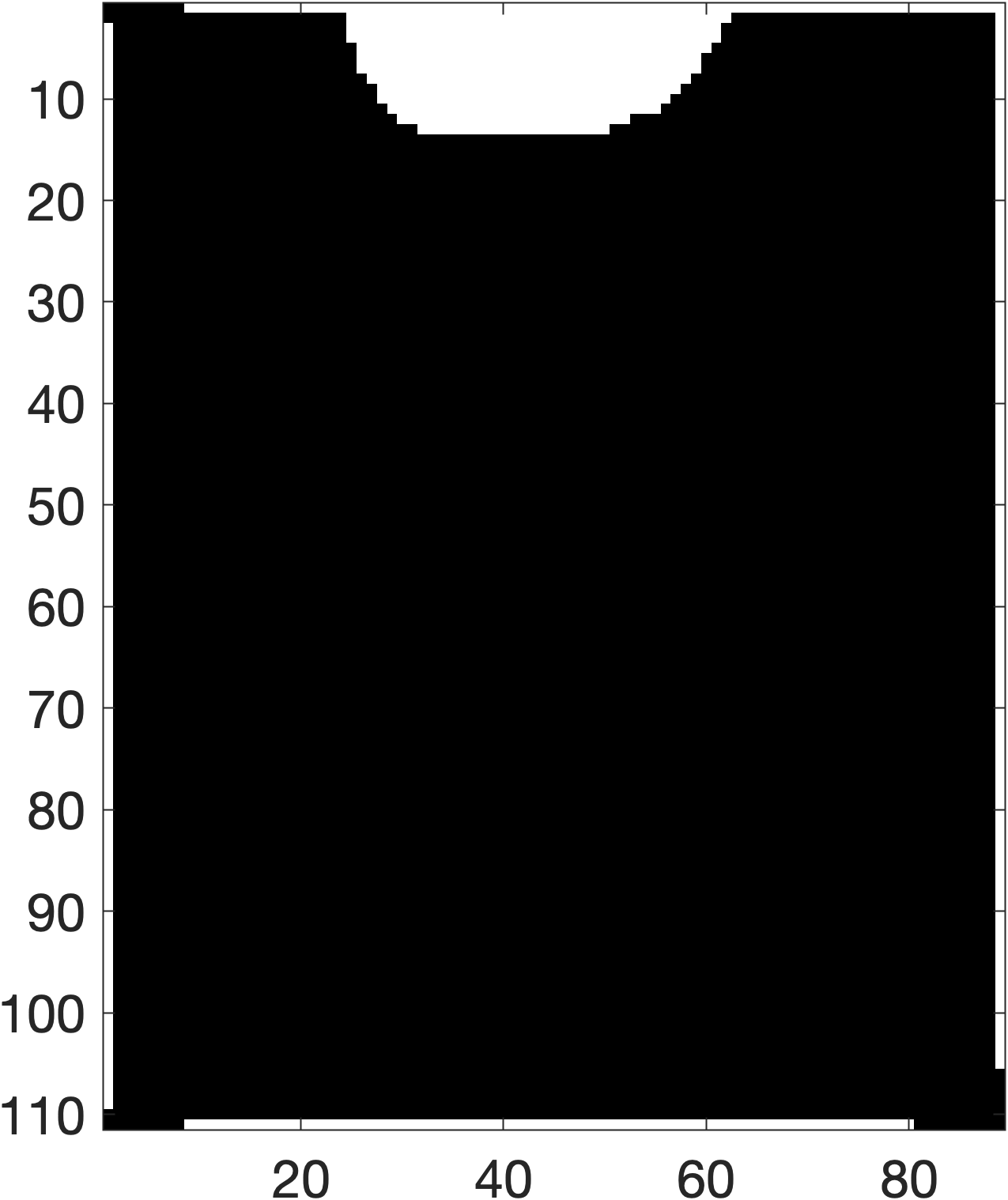}
        \caption{Follower with $\lambda_{g_1} = 0.2$ and $\gamma_1 = 0.25$.}
        \label{fig_followerOpt_e}
    \end{subfigure}\hfill
    \begin{subfigure}[t]{0.45\linewidth}
    \centering
        \includegraphics[width=0.7\linewidth]{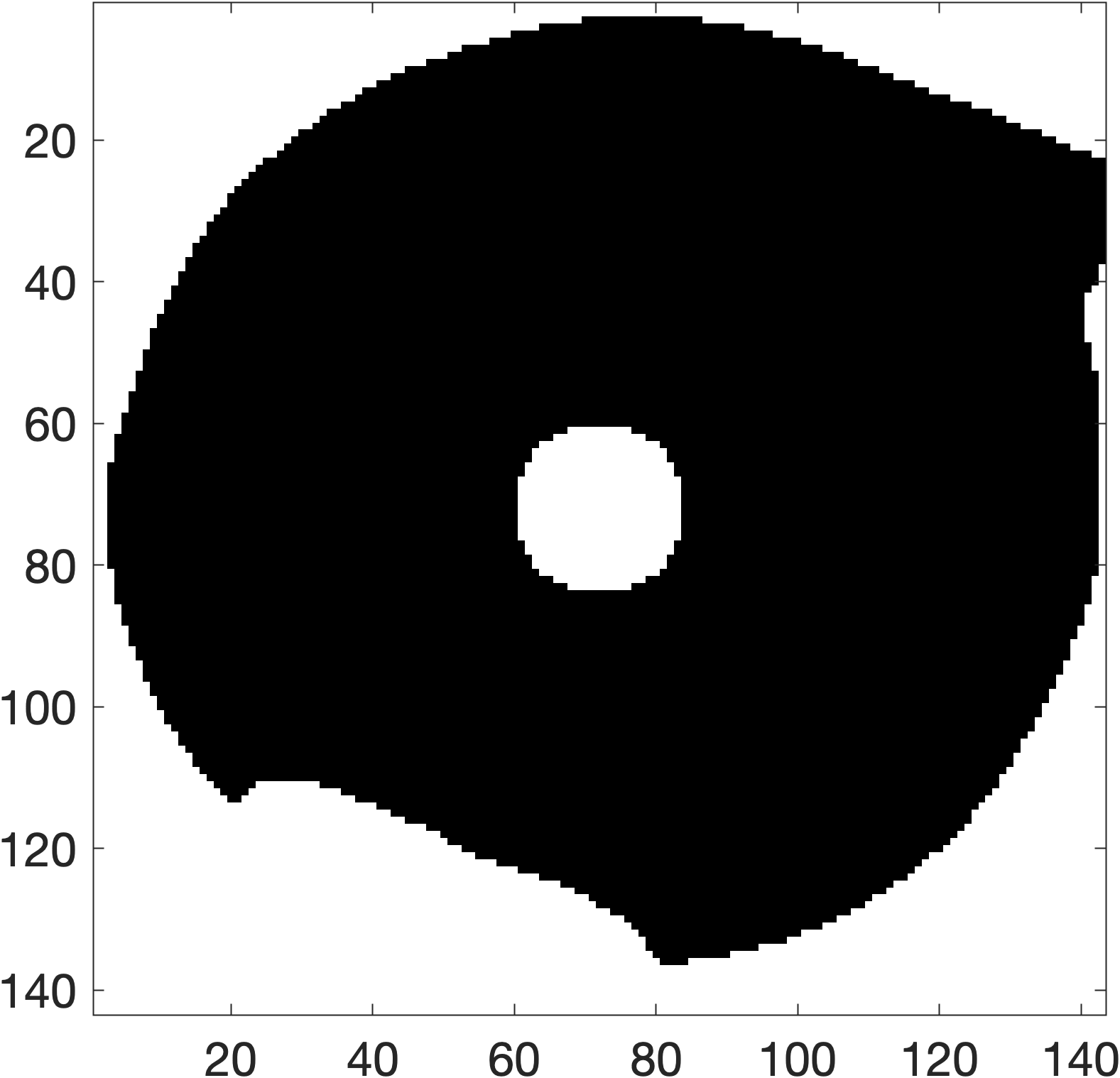}
        \caption{Cam with $\lambda_{g_2} = 0.2$ and $\gamma_2 = 1$.}
        \label{fig_camOpt_f}
    \end{subfigure}
    
    \centering
    \caption{Co-optimized cam-follower system with different parameters.}
    \label{fig_camFollowerOptimized}
\end{figure}

\begin{figure} [h!]
    \centering
\includegraphics[width=\linewidth]{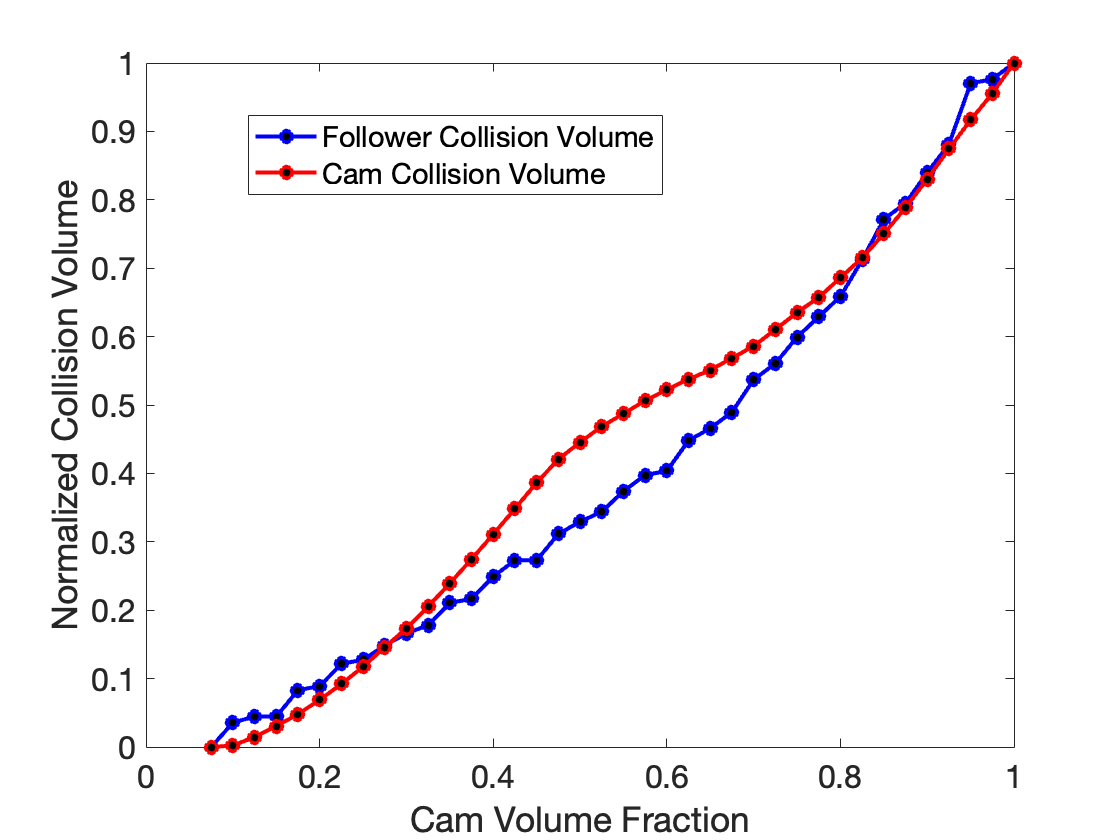}
        \caption{The evolution of collision volume between the cam and the follower without considering compliance ($\lambda_{g_1}=\lambda_{g_2} = 0$) and only removing material from the cam ($\gamma_{1}= 0$ and $\gamma_{2} = 1$). The final cam geometry is similar to follower unsweep. }
    \label{fig_camFollowerConvergenceUnsweep}
\end{figure}

\begin{figure*} [h!]
    \centering
        \begin{subfigure}[t]{.3\linewidth}
        \includegraphics[width=0.9\linewidth]{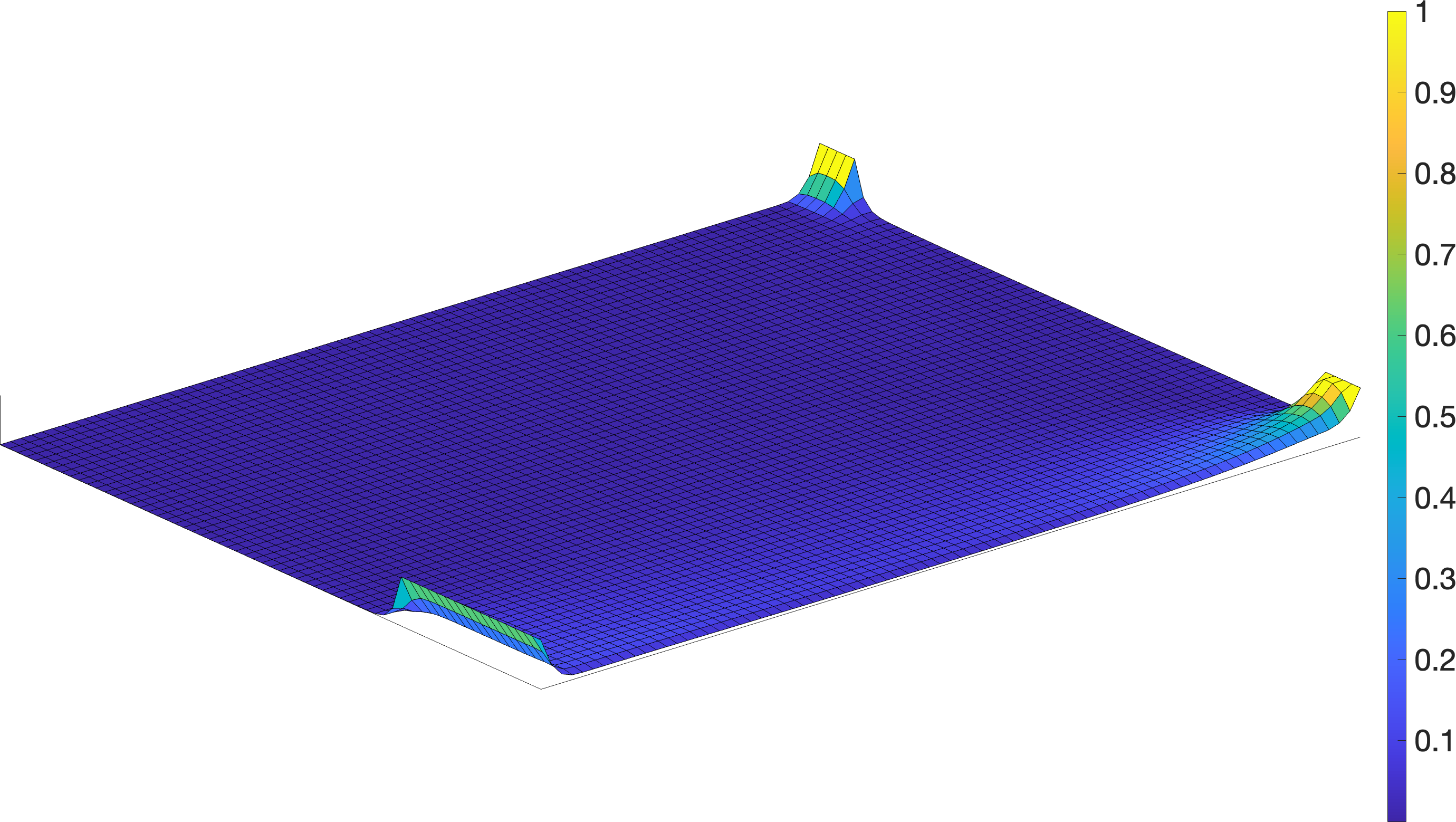}
        \caption{Follower compliance TSF. }
        \label{fig_followerCompTSF}
    \end{subfigure}
        \begin{subfigure}[t]{.3\linewidth}
        \includegraphics[width=0.9\linewidth]{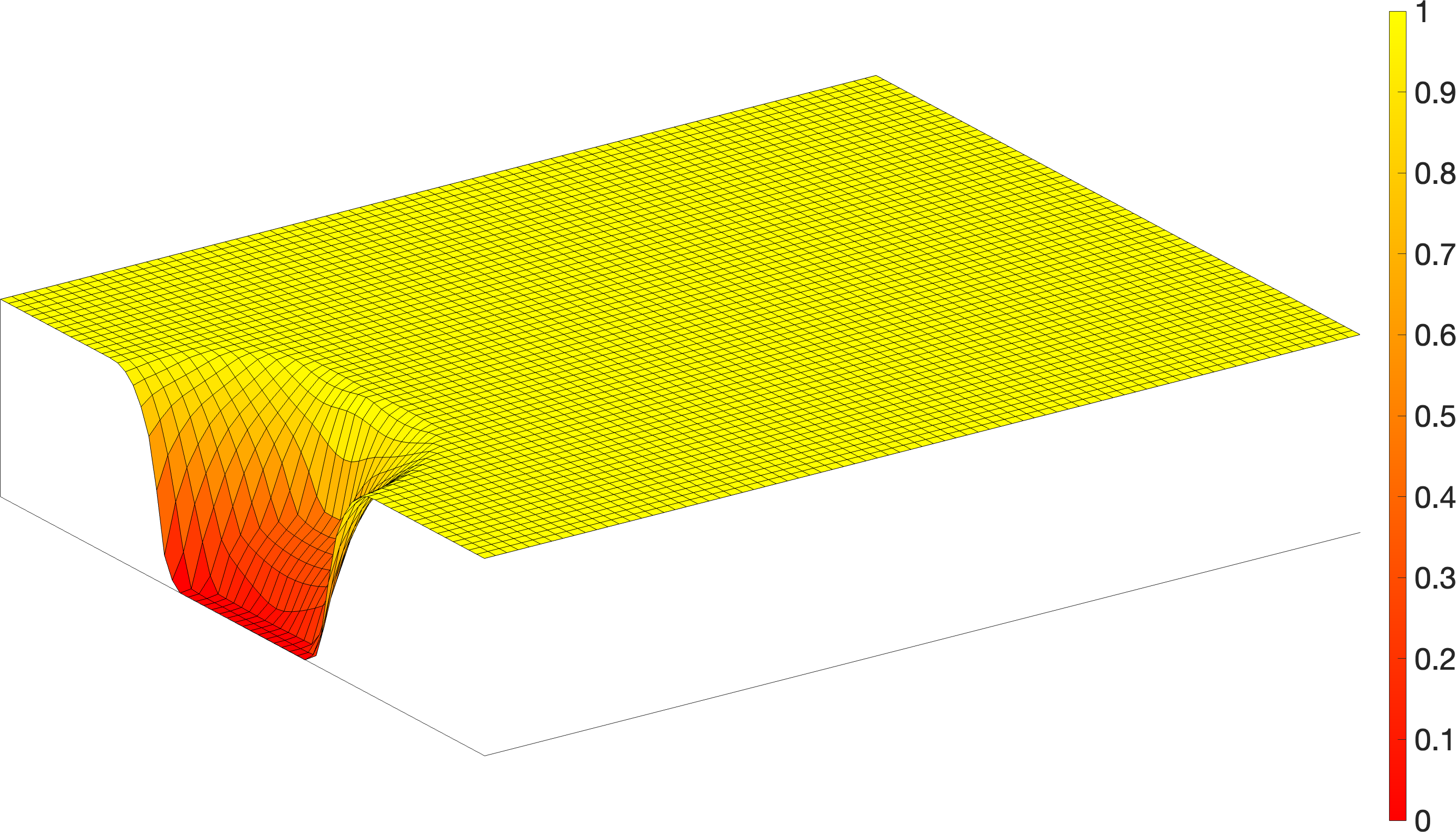}
        \caption{Follower collision gradient. }
        \label{fig_followerCollGrad}
    \end{subfigure}
    \begin{subfigure}[t]{.3\linewidth}
        \includegraphics[width=0.9\linewidth]{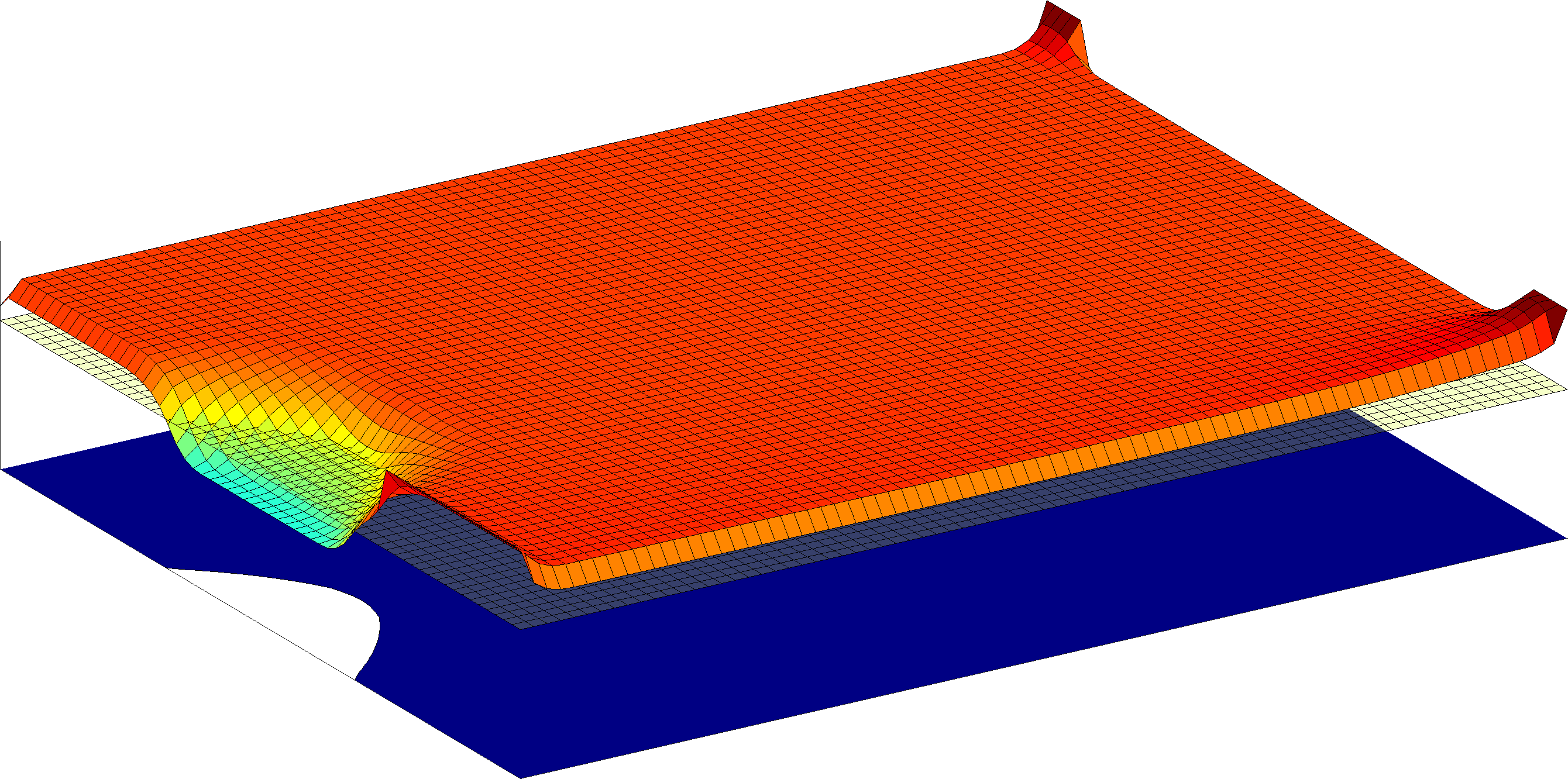}
        \caption{Follower augmented TSF with $\lambda_{g_1} = 0.5$. }
        \label{fig_followerAugTSF}
    \end{subfigure}

          \begin{subfigure}[t]{.3\linewidth}
        \includegraphics[width=0.9\linewidth]{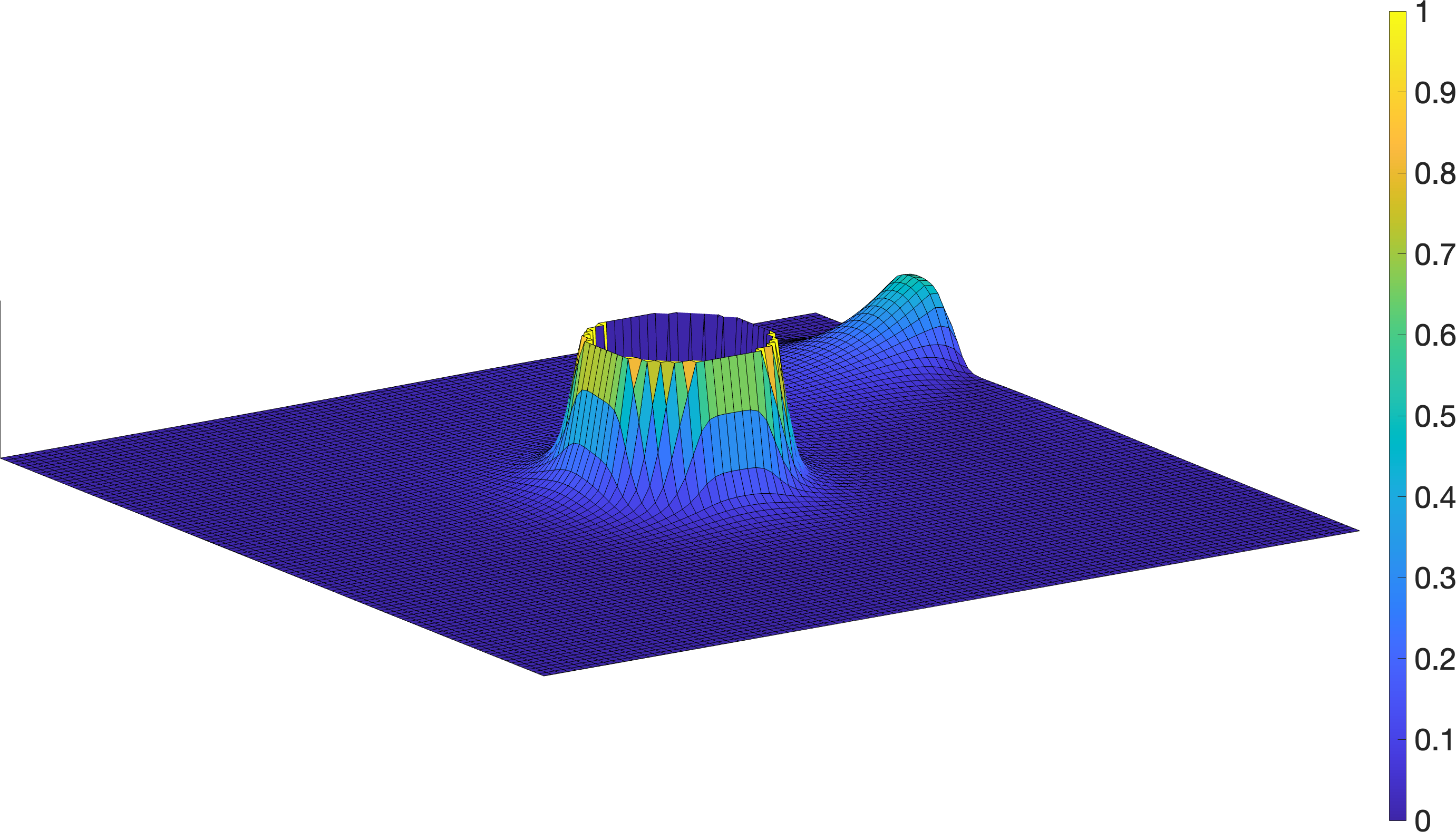}
        \caption{Cam compliance TSF. }
        \label{fig_camCompTSF}
    \end{subfigure}
        \begin{subfigure}[t]{.3\linewidth}
        \includegraphics[width=0.9\linewidth]{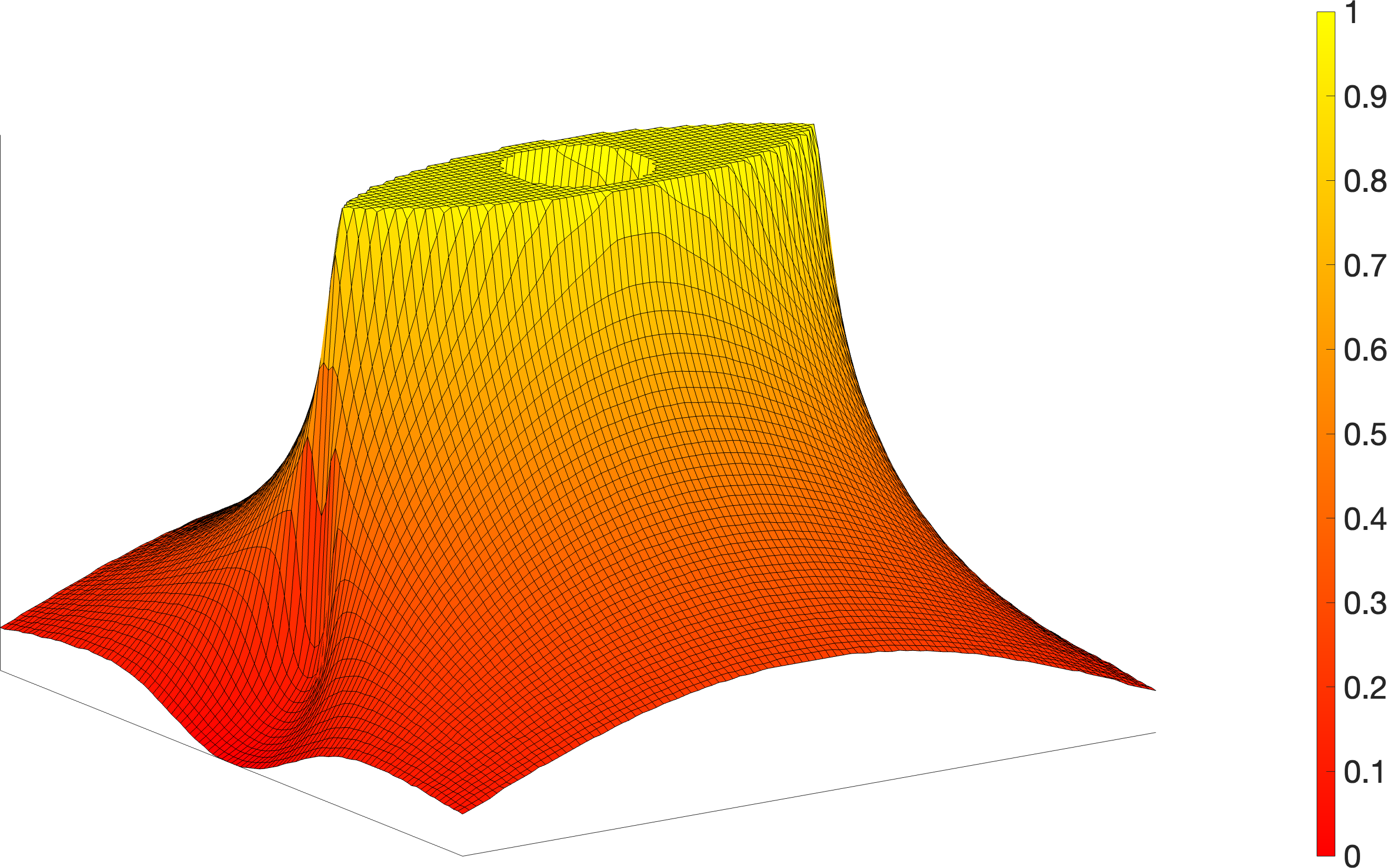}
        \caption{Cam collision gradient. }
        \label{fig_camCollGrad}
    \end{subfigure}
    \begin{subfigure}[t]{.3\linewidth}
        \includegraphics[width=0.9\linewidth]{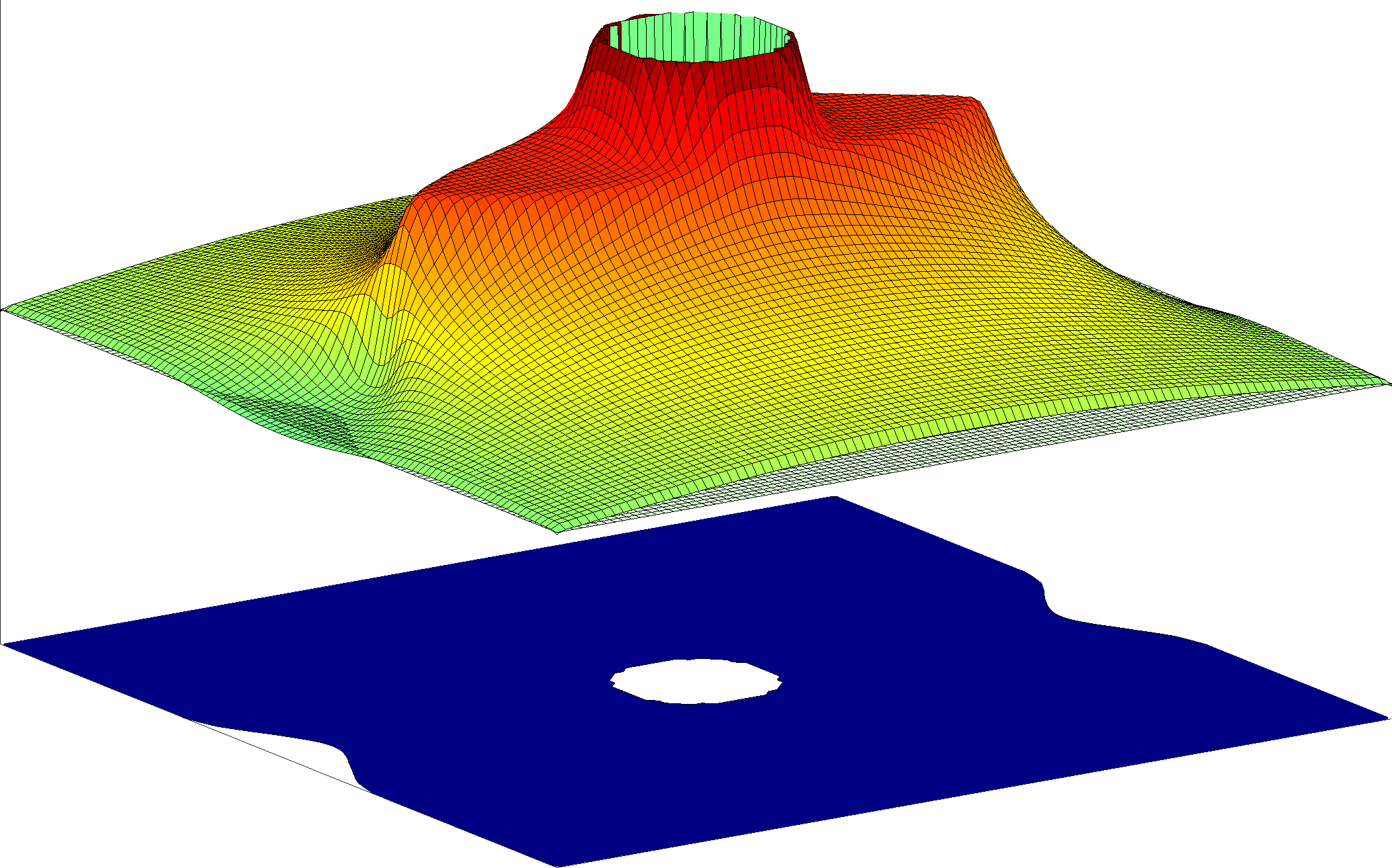}
        \caption{Cam augmented TSF with $\lambda_{g_2} = 0.5$.}
        \label{fig_camAugTSF}
    \end{subfigure}
    \centering
    \caption{Cam-follower augmented TSFs with $\lambda_{g_i}=0.5$ with the level-set at 0.025 decrement.}
    \label{fig_camFollowerTSFs}
\end{figure*}

The first example is the cam-follower system of \Cref{fig_camFollowerConfig} with the dimensions and initial positions shown in \Cref{fig_camFollowerCollFields0}.
The cam is prescribed to rotate $2\pi$ radians about the center of its circular cutout hole, which is located at $O_{\mathrm{C}} = [0,8]$ of the space, while the follower moves in the vertical direction as a function of the angular position of the cam, $\theta_{\mathrm{C}}$. The vertical position $y_{\mathrm{F}}$ of the center of the follower, as a function of $\theta_{C}$ provided by the following formula:
\begin{equation}
    \label{followerEquation}
    y_{\mathrm{F}} = {{3L}\over{4}} + {{L}\over{8}}{\cos(2\theta_{\mathrm{C}})}.
\end{equation}
The temporal resolution for collision analysis is 1000 time steps.
 The follower and the cam are discretized into 10,000 and 20,000 bilinear quadrilateral finite elements, respectively. The boundary conditions (BC) for the follower is illustrated in \Cref{fig_followerBC}, where we  assume fixed degrees of freedom (DOF) along the y axis at the bottom-left corner and fixed DOFs in both x and y at the bottom-right corner. An external force $f^F_{ext}=[1,0]$ is applied at the top-left corner. \Cref{fig_squareCamBC} illustrates the BC for the square cam, where the circular cutout hole is assumed fixed and an external force $F^C_{ext} = [0,-1]$ is applied at the right-top corner. 
 \Cref{fig_camFollowerCollFields0} shows the initial collision measure fields for the two components under the prescribed trajectories, which results in the initial collision regions of \Cref{fig_camFollowerCollRegions}. 
 \Cref{fig_camFollowerTSFs} shows the compliance TSF, collision gradient field, and the augmented sensitivity fields for both the cam and the follower with $\lambda_{g_1} = \lambda_{g_2} = 0.5$. Observe that removing material solely based on the collision gradient would removed critical regions where Dirichlet or Neumann BC are applied as demonstrated in Figures \ref{fig_followerOpt_a} and \ref{fig_camOpt_b}, where we only optimize the cam geometry \textit{without} considering compliance. In other words, the optimization reduces to finding the follower unsweep as the cam shape, which is \textit{not} a valid design from the compliance perspective ($f_C \rightarrow \infty$). The evolution of collision volume for the follower and the cam are shown in \Cref{fig_camFollowerConvergenceUnsweep}.\\
 On the other hand, TSF alone does not capture information about collision and is unlikely to produce a collision-free assembly. However, the augmented sensitivity field encapsulates information about both constraints and will successfully result in collision-free and physically valid assemblies. Figures \ref{fig_followerOpt_c} and \ref{fig_camOpt_d}  illustrate the co-optimized cam-follower assembly for $\lambda_{g_1} = \lambda_{g_2} = 0.2$, $\gamma_1 = 1$, and $\gamma_2 = 0.5$. Alternatively, Figures \ref{fig_followerOpt_e} and \ref{fig_camOpt_f}  show the co-optimized parts for $\lambda_{g_1} = \lambda_{g_2} = 0.2$, $\gamma_1 = 0.25$, and $\gamma_2 = 1$. \Cref{table_camFollowerResults} summarizes the results for the optimized cam-follower systems with different parameters $\lambda_{g_i}$ and $\gamma_i$. 
 
\begin{table*}[!ht]
\centering
\caption{Impact of $\gamma_i$ and $ \lambda_{g_i}$ on compliance and collision for the cam-follower system.}
\begin{tabular}{c|cc|cc|cc|cc}
\hline \hline 
$\lambda_{g_i}$ & {$\gamma_{1}$}& {$\gamma_{2}$} &$v_1$&$v_2$& $f_1/f_1^0$ & $f_2/f_2^0$ & $\Gield_1$ & $\Gield_2$ \\ \hline
1 & 0 & 1 & 1.00 & 0.08 & 1.00 & $\infty$ & 0.00 & 0.00 \\
1 & 1 & 0 & 0.93 & 1.00 & 1.10 & 1.00 & 0.00 & 0.00 \\
0.05 & 1 & 1 & 0.90 & 0.90 & 1.02 & 1.01 & 0.00 & 0.00 \\
0.2 & 1 & 1 & 0.90 & 0.90 & 1.05 & 1.01 & 0.00 & 0.00 \\
0.5 & 1 & 1 & 0.90 & 0.90 & 1.05 & 1.01 & 0.00 & 0.00 \\
0.2 & 1 & 0.5 & 0.90 & 0.95 & 1.05 & 1.00 & 0.00 & 0.00 \\
0.2 & 0.5 & 1 & 0.91 & 0.83 & 1.05 & 1.01 & 0.00 & 0.00 \\
0.2 & 0.25 & 1 & 0.93 & 0.70 & 1.05 & 1.04 & 0.00 & 0.00 \\
0.2 & 1 & 0.25 & 0.90 & 0.98 & 1.05 & 1.00 & 0.00 & 0.00 \\
\end{tabular}

\label{table_camFollowerResults}
\end{table*}

%%%%%%
%%%%%%
\subsection{Three Squares Assembly}
\begin{figure} [h!]
    \centering
    \includegraphics[width=0.5\linewidth]{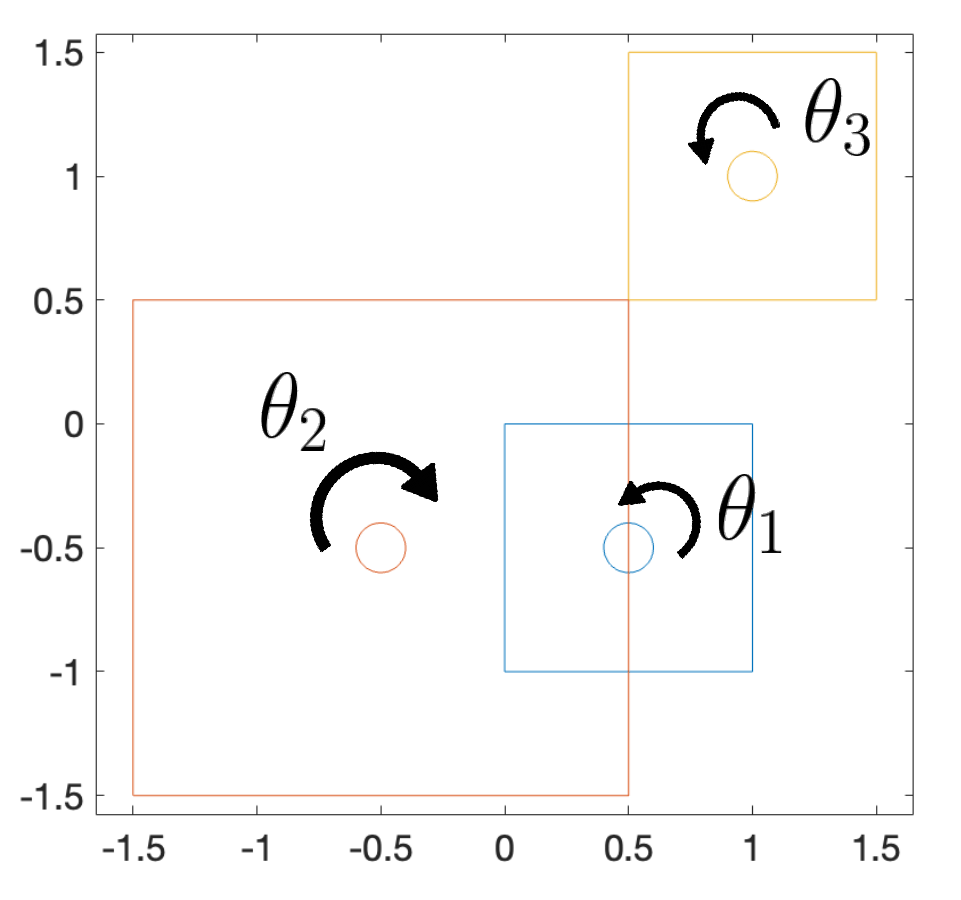}
    \caption{Cam and follower initial configuration.}
    \label{fig_threeSquareConfig}
\end{figure}

\begin{figure} [h!]
    \centering
    \begin{subfigure}[t]{.3\linewidth}
        \includegraphics[width=\linewidth]{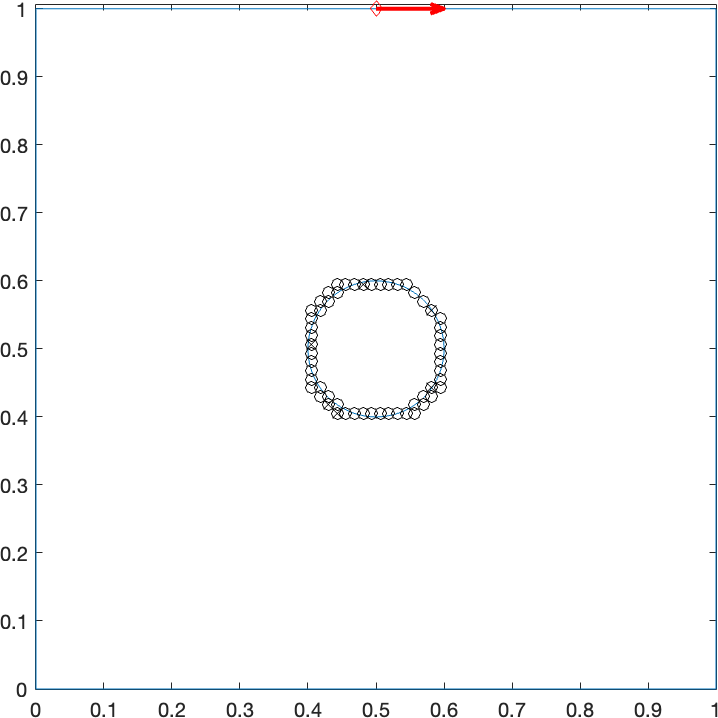}
        \caption{}

    \end{subfigure}
    \begin{subfigure}[t]{.3\linewidth}
        \includegraphics[width=\linewidth]{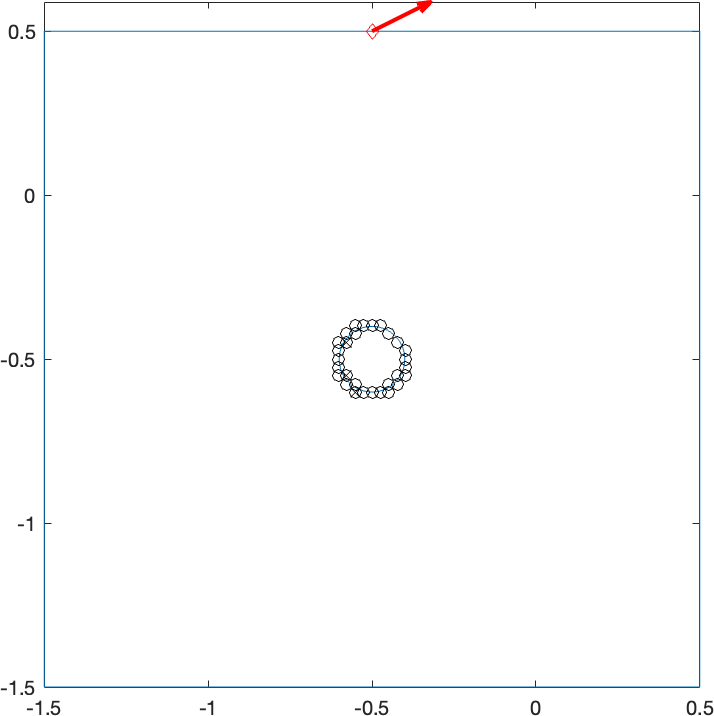}
        \caption{}

    \end{subfigure}
        \begin{subfigure}[t]{.3\linewidth}
        \includegraphics[width=\linewidth]{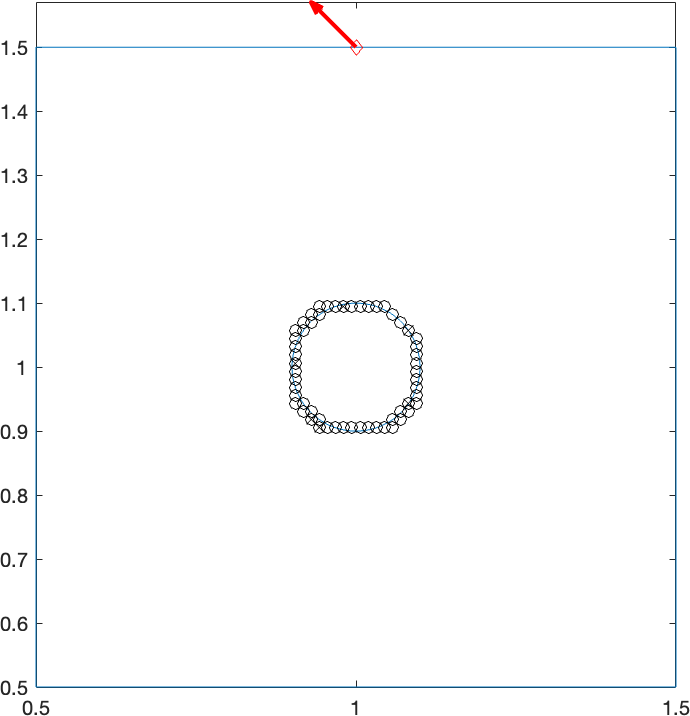}
        \caption{}
    \end{subfigure}
    \centering
    \caption{Three rotating-body boundary conditions.}
    \label{fig_threeSquareBC}
\end{figure}

\begin{figure} [h!]
    \centering
    \begin{subfigure}[t]{.3\linewidth}
        \includegraphics[width=\linewidth]{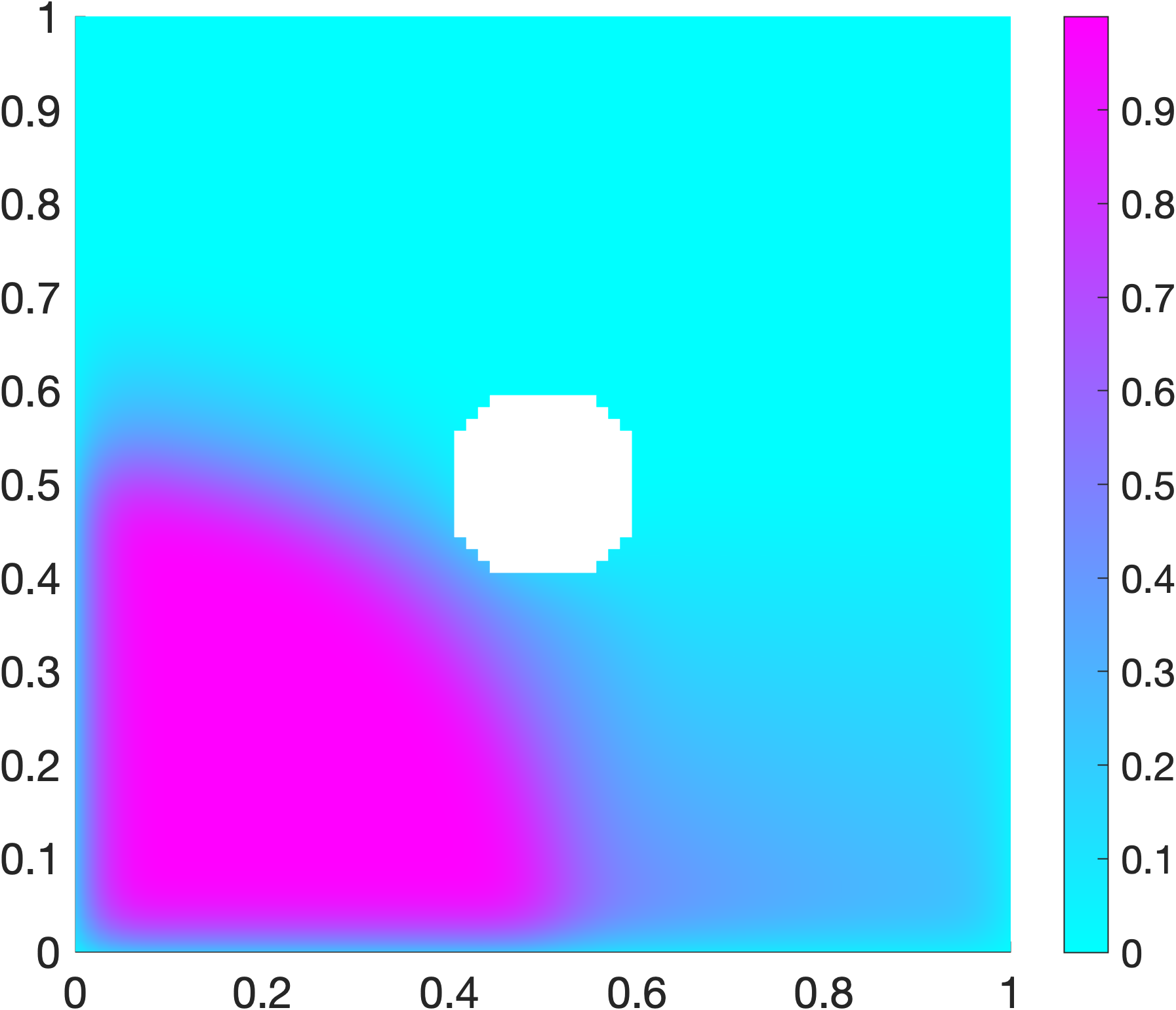}
        \caption{}
    \end{subfigure}
    \begin{subfigure}[t]{.3\linewidth}
        \includegraphics[width=\linewidth]{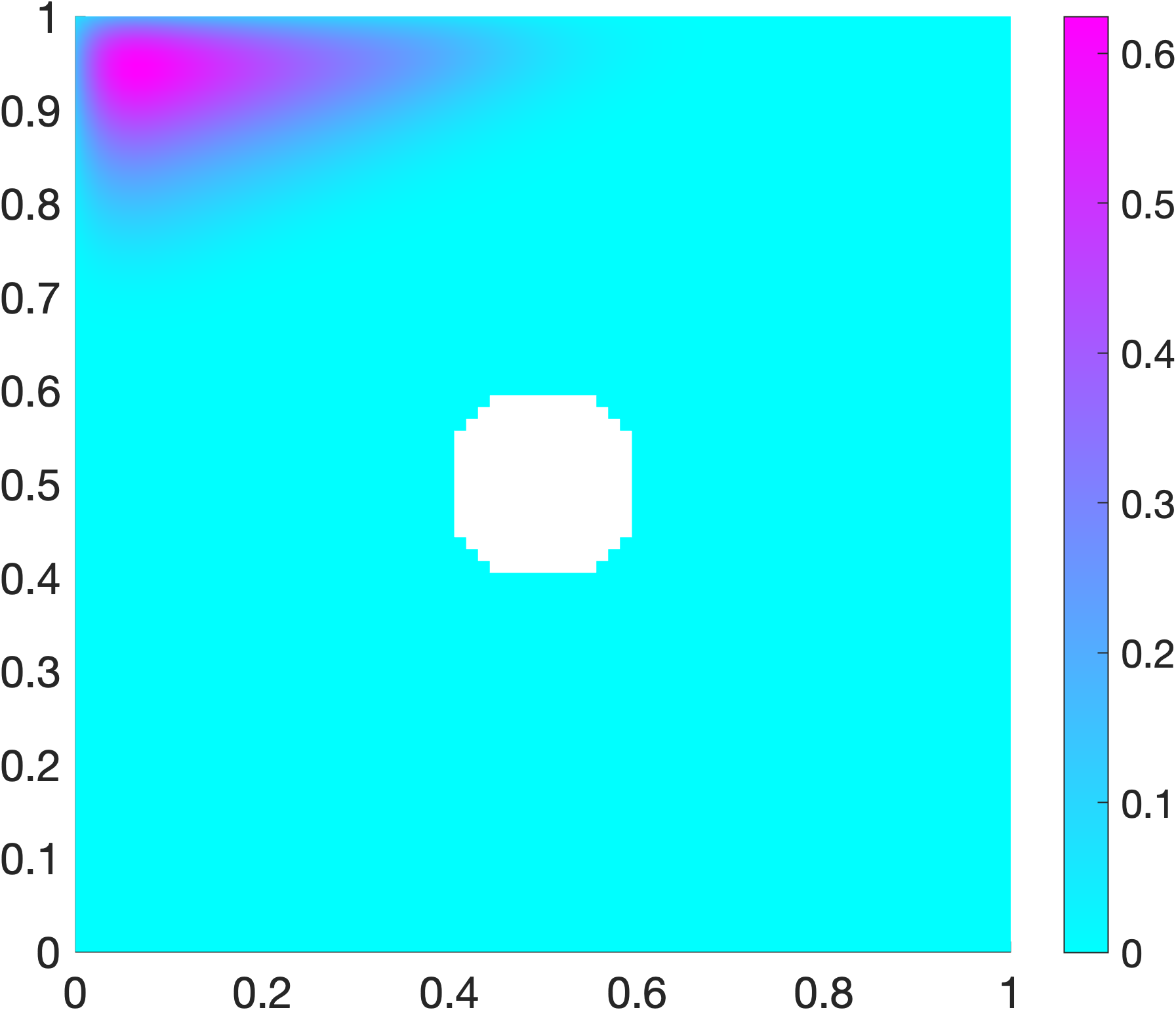}
        \caption{}
    \end{subfigure}
        \begin{subfigure}[t]{.3\linewidth}
        \includegraphics[width=\linewidth]{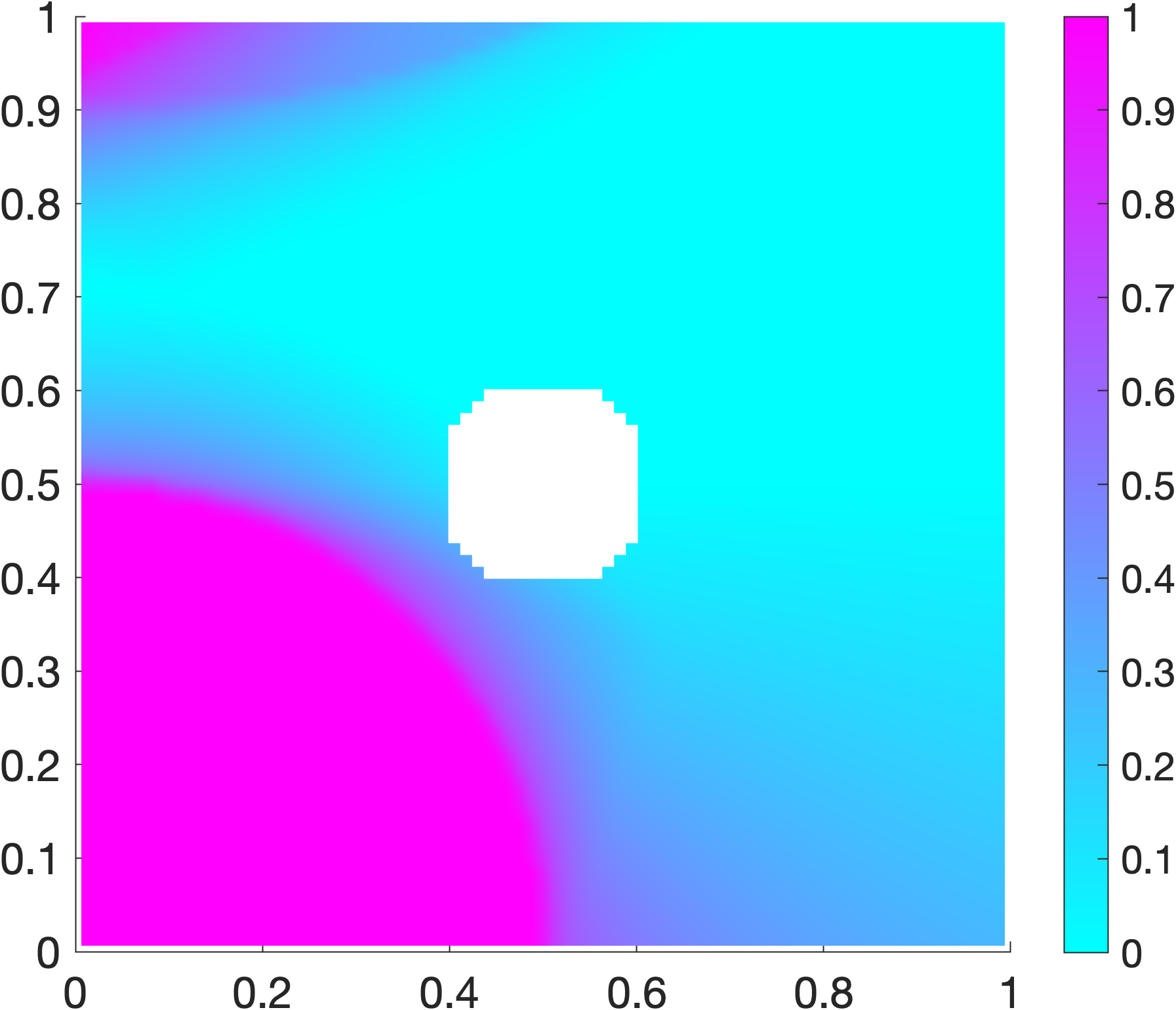}
        \caption{}
    \end{subfigure}

        \begin{subfigure}[t]{.3\linewidth}
        \includegraphics[width=\linewidth]{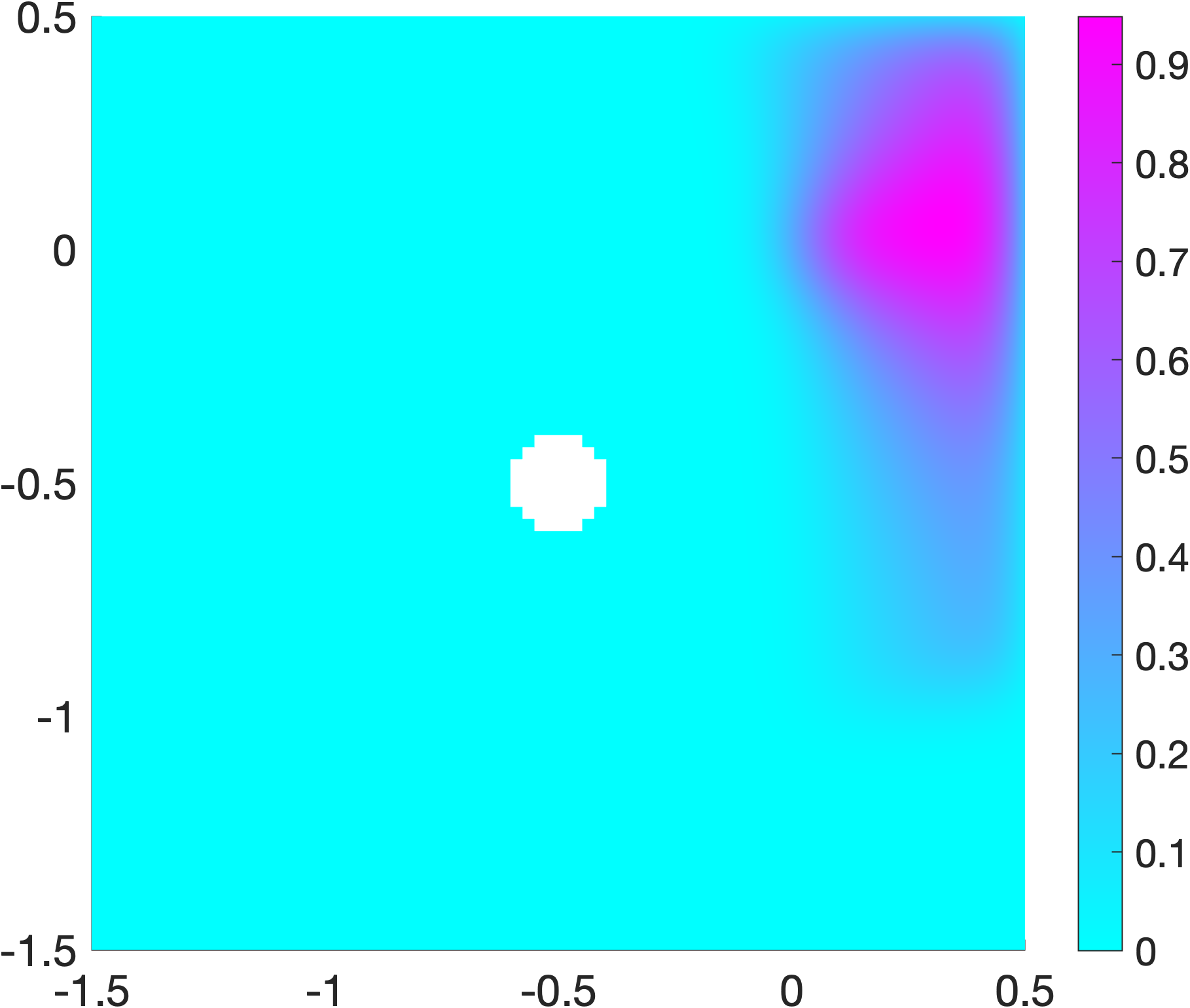}
        \caption{}
    \end{subfigure}
    \begin{subfigure}[t]{.3\linewidth}
        \includegraphics[width=\linewidth]{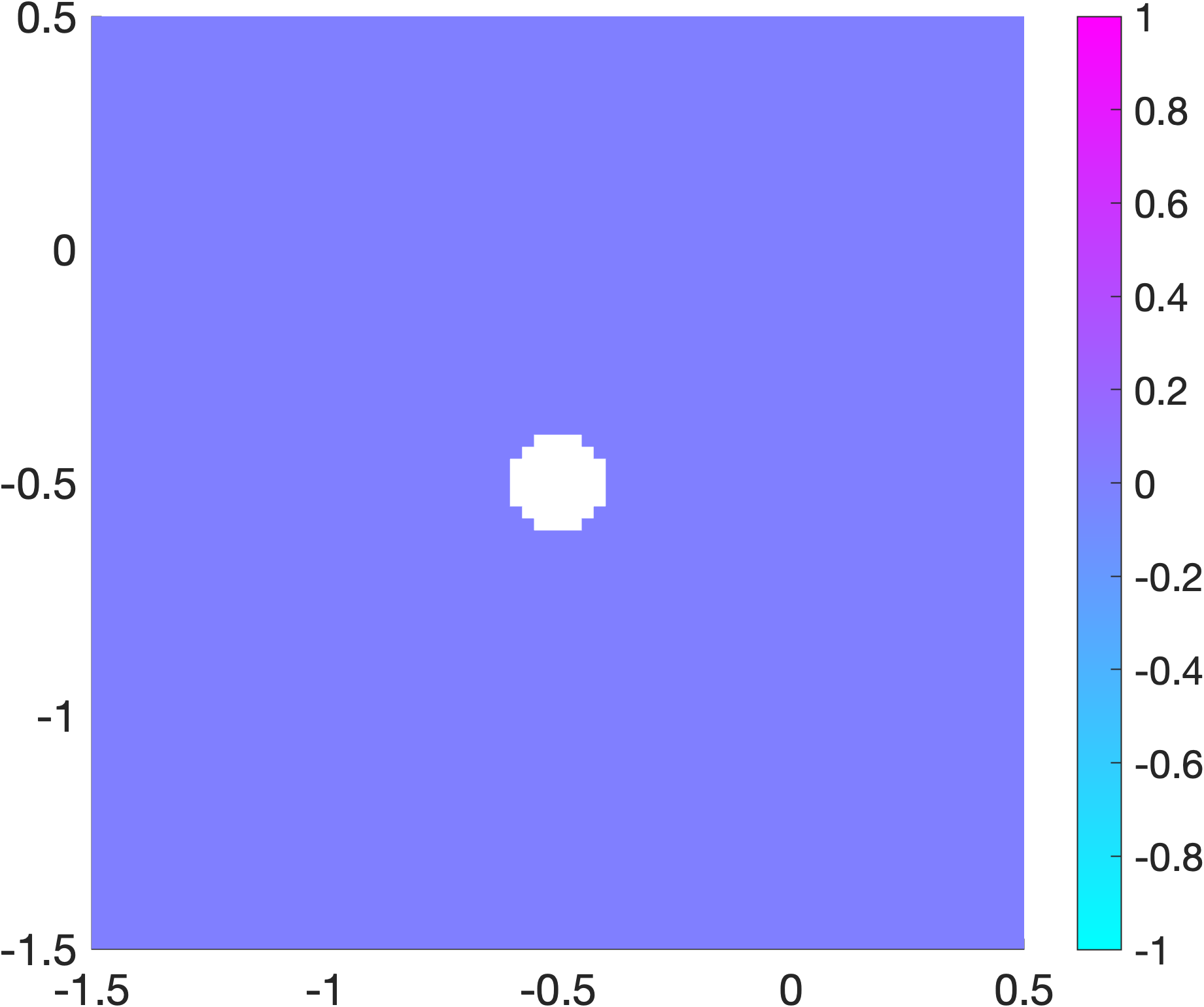}
        \caption{}
    \end{subfigure}
        \begin{subfigure}[t]{.3\linewidth}
        \includegraphics[width=\linewidth]{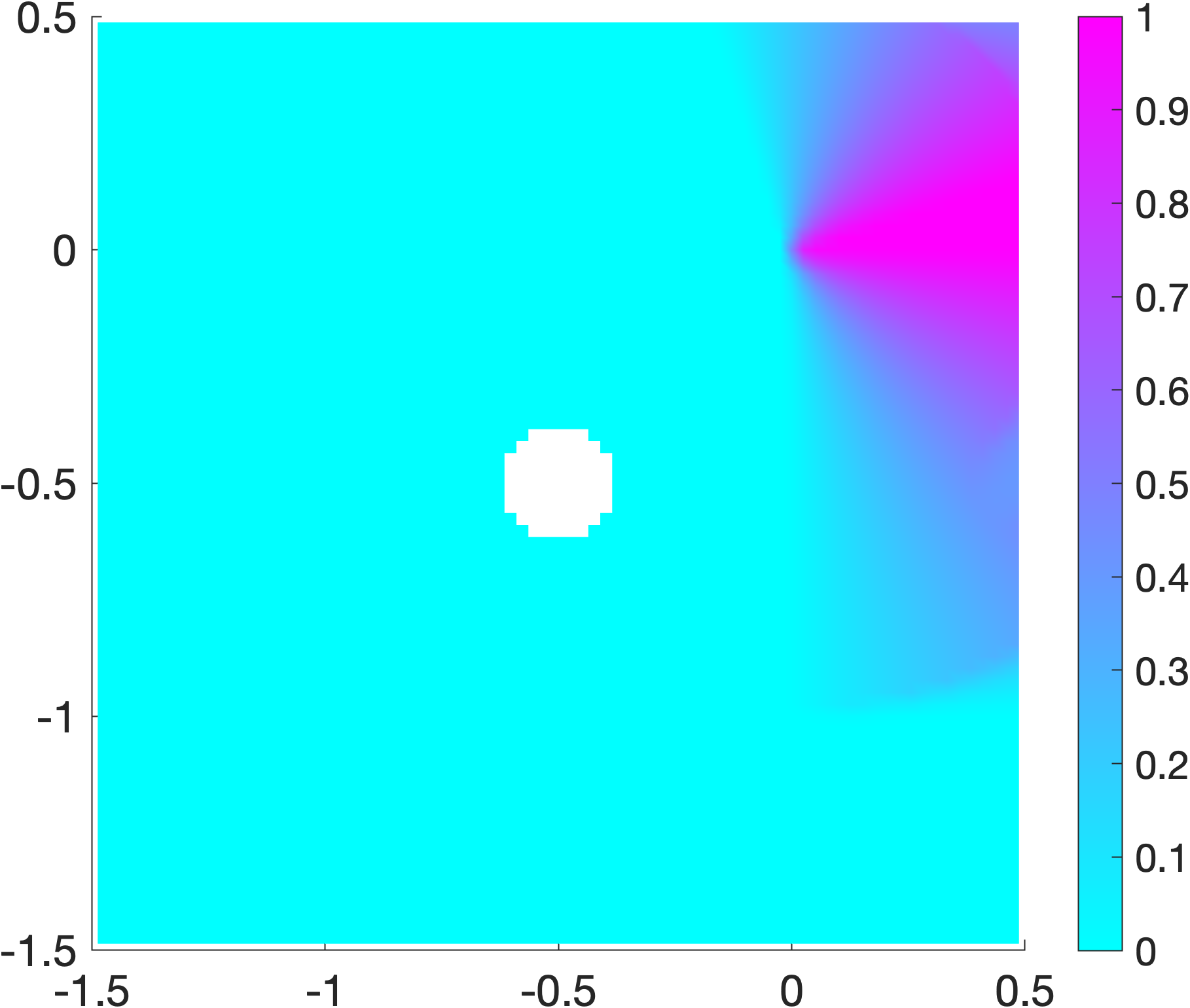}
        \caption{}
    \end{subfigure}

            \begin{subfigure}[t]{.3\linewidth}
        \includegraphics[width=\linewidth]{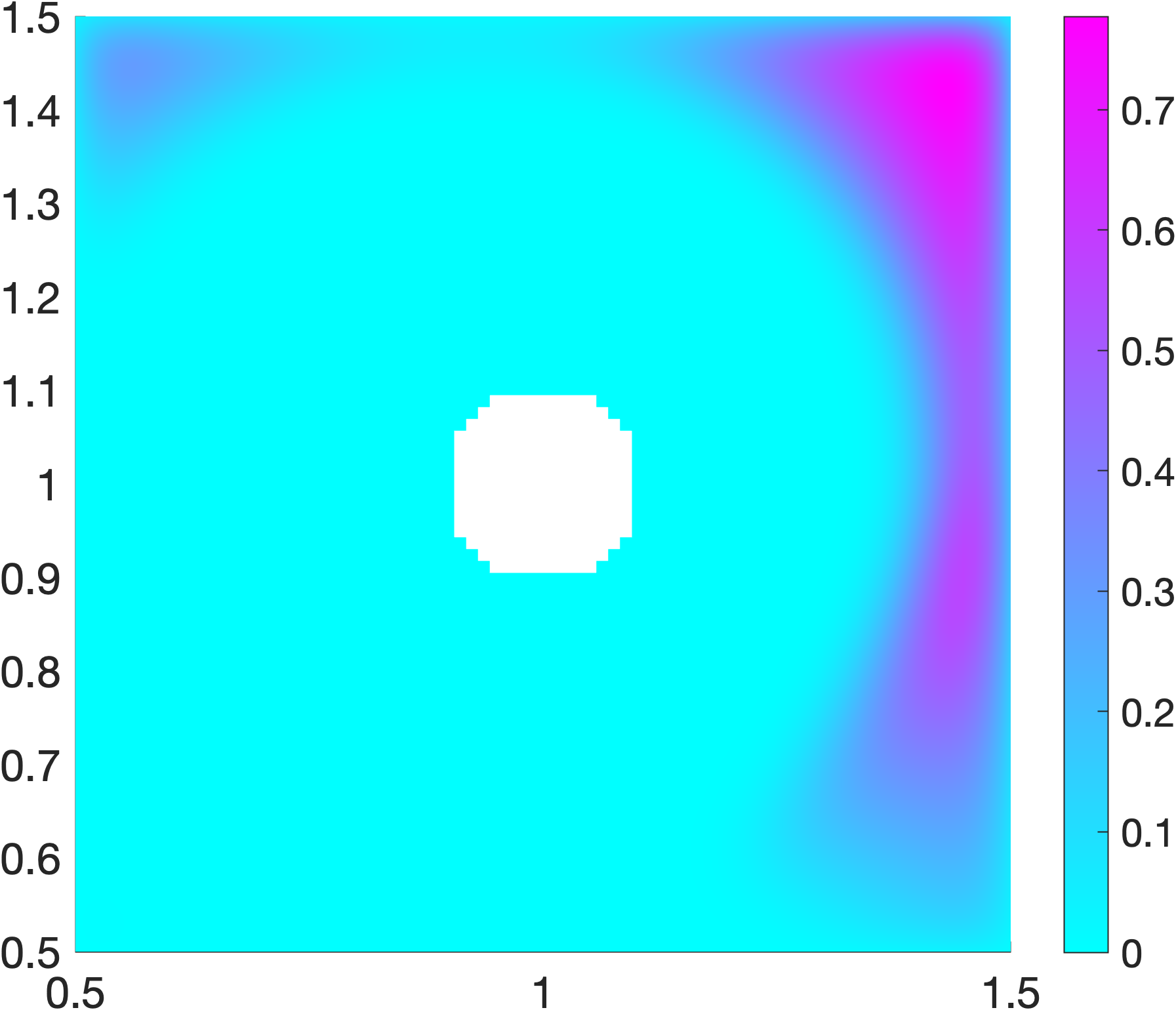}
        \caption{}
    \end{subfigure}
    \begin{subfigure}[t]{.3\linewidth}
        \includegraphics[width=\linewidth]{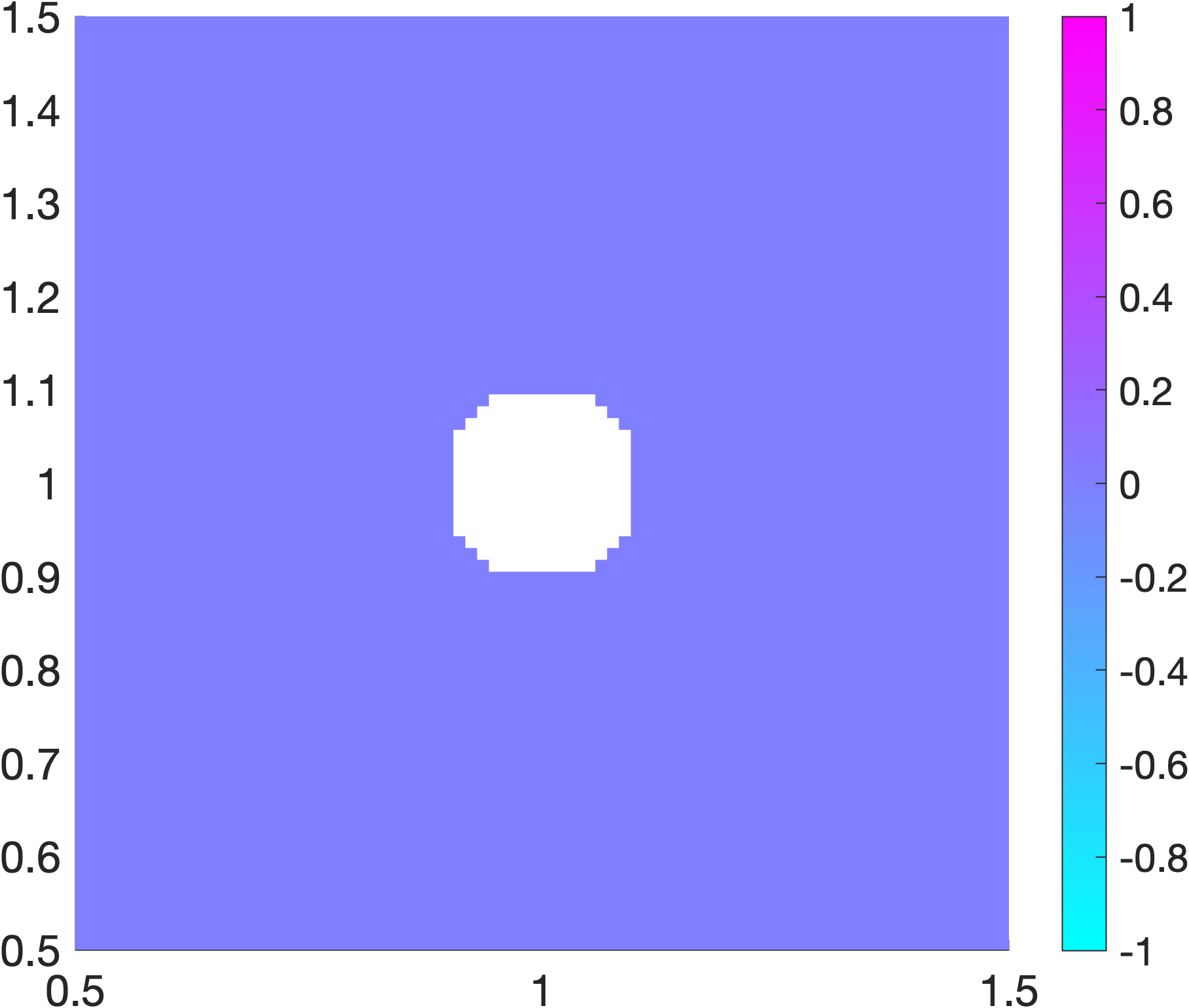}
        \caption{}
    \end{subfigure}
        \begin{subfigure}[t]{.3\linewidth}
        \includegraphics[width=\linewidth]{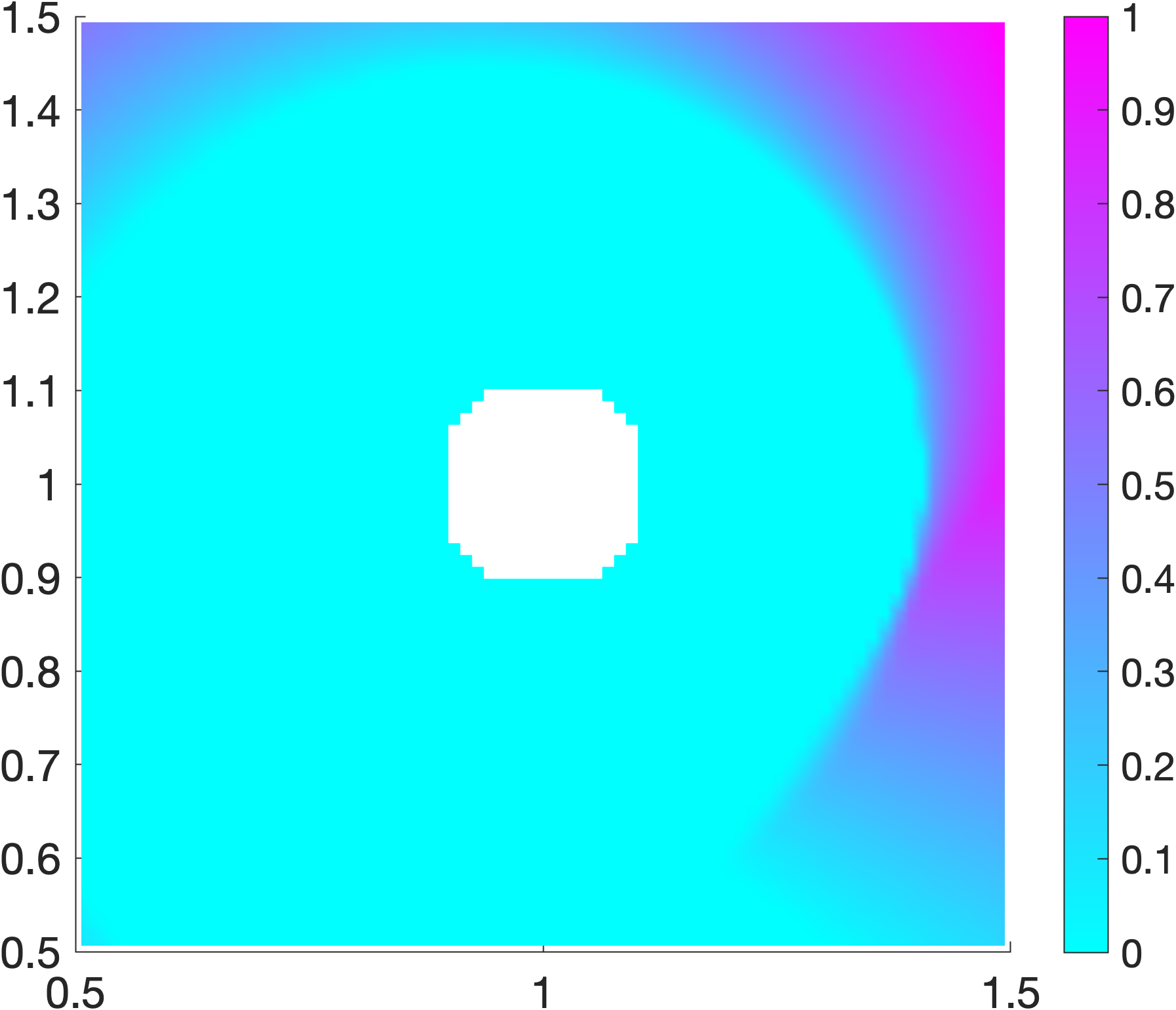}
        \caption{}
    \end{subfigure}
    \centering
    \caption{Three rotating-body initial pair-wise and overall collision measure fields.}
    \label{fig_threeSquareCollFields}
\end{figure}

\begin{figure} [h!]
    \centering
    \begin{subfigure}[t]{.3\linewidth}
        \includegraphics[width=0.9\linewidth]{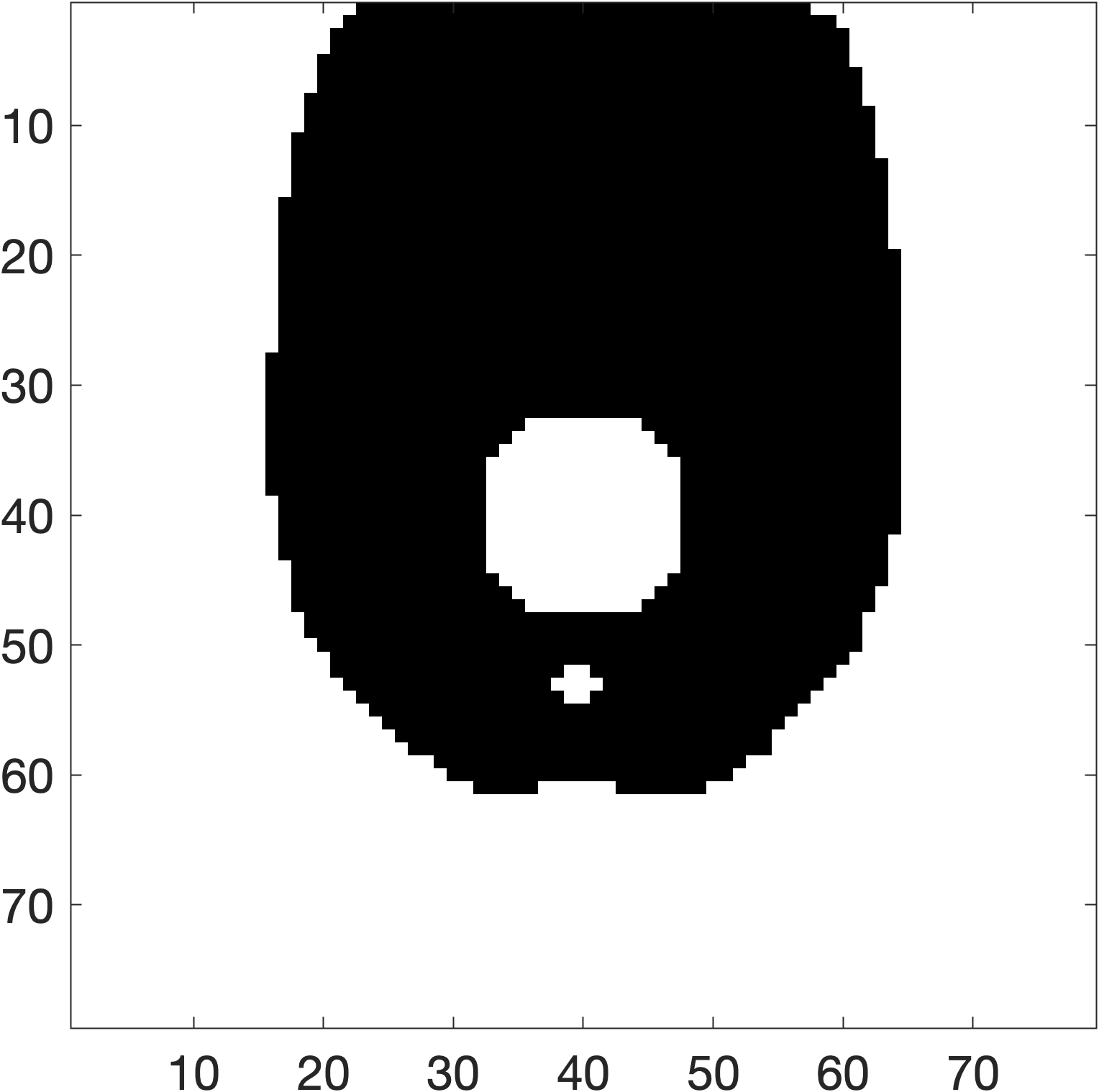}
        \caption{$\gamma_1 = 1$}

    \end{subfigure}
    \begin{subfigure}[t]{.3\linewidth}
        \includegraphics[width=0.9\linewidth]{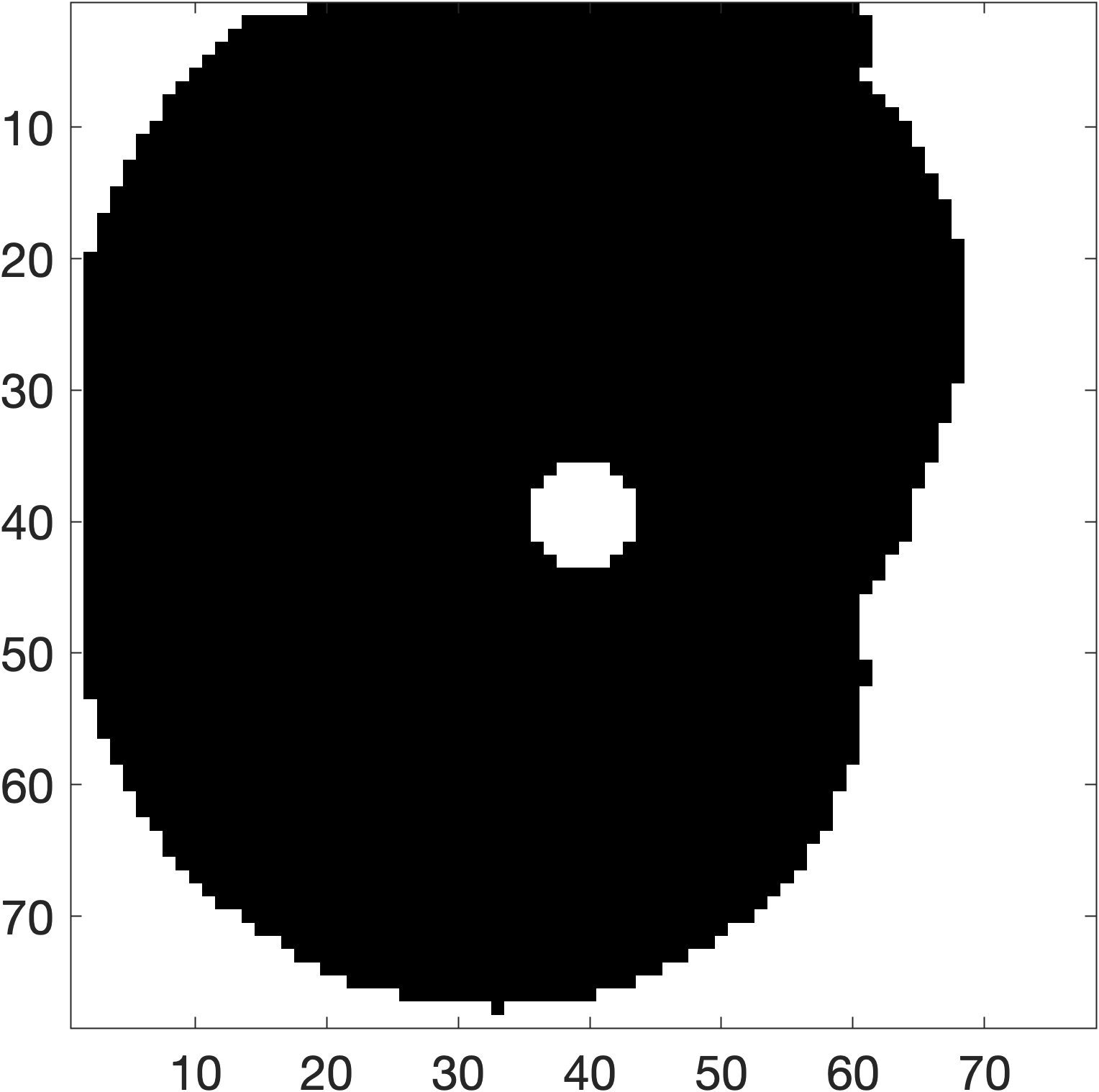}
        \caption{$\gamma_2 = 0.5$}

    \end{subfigure}
        \begin{subfigure}[t]{.3\linewidth}
        \includegraphics[width=0.9\linewidth]{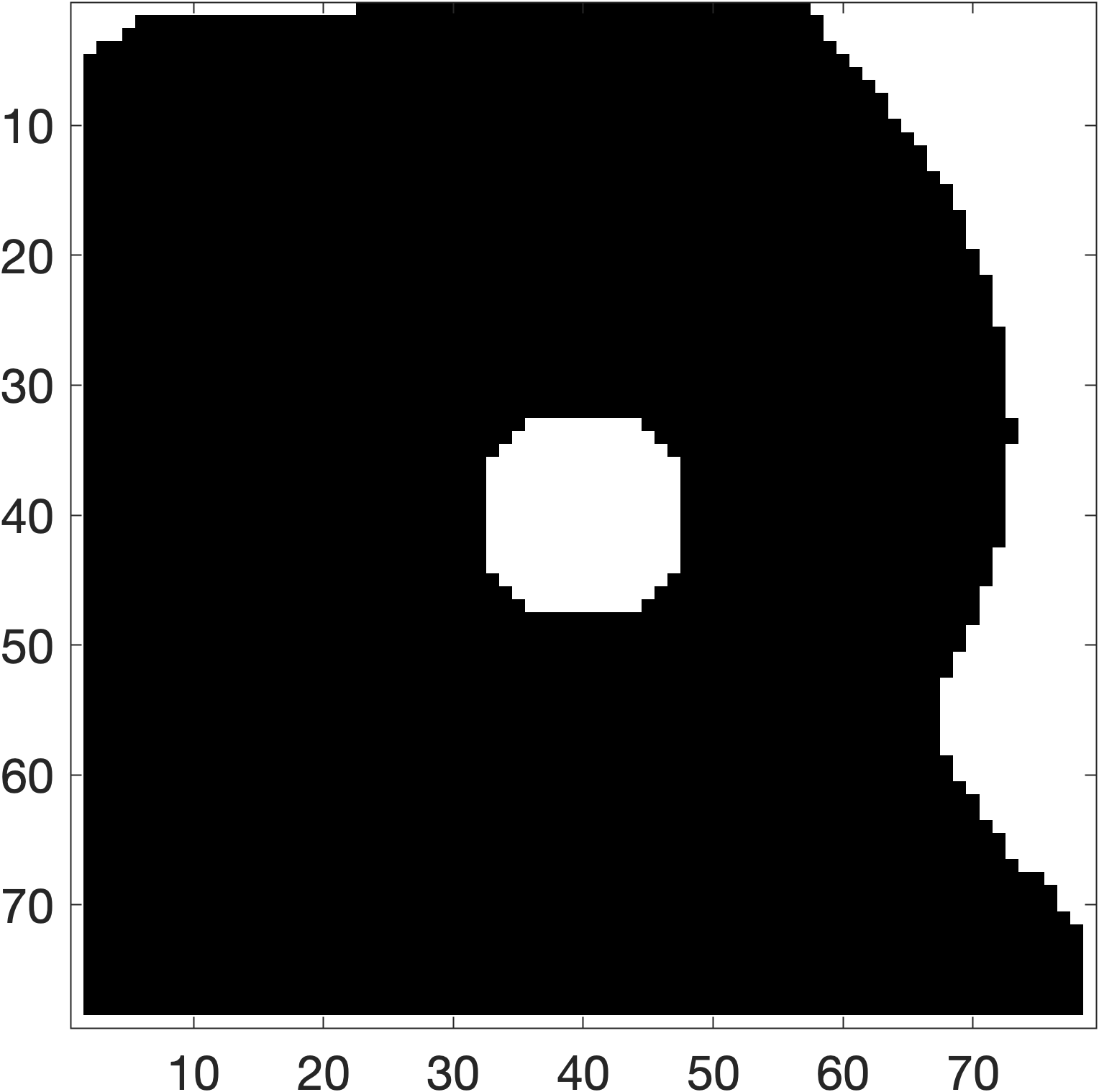}
        \caption{$\gamma_3 = 0.25$}
    \end{subfigure}

    \begin{subfigure}[t]{.3\linewidth}
        \includegraphics[width=0.9\linewidth]{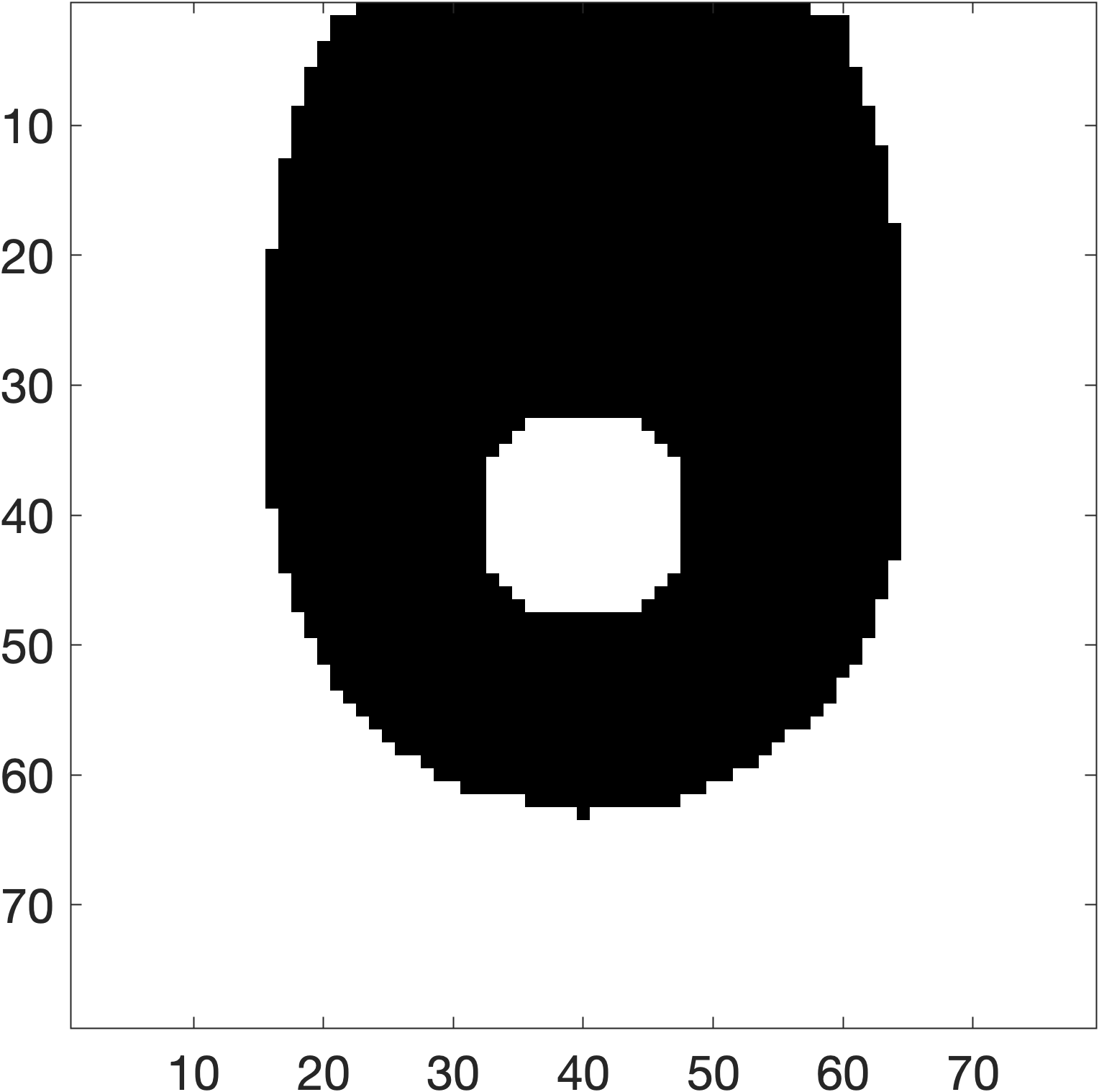}
        \caption{$\gamma_1 = 0.5$}

    \end{subfigure}
    \begin{subfigure}[t]{.3\linewidth}
        \includegraphics[width=0.9\linewidth]{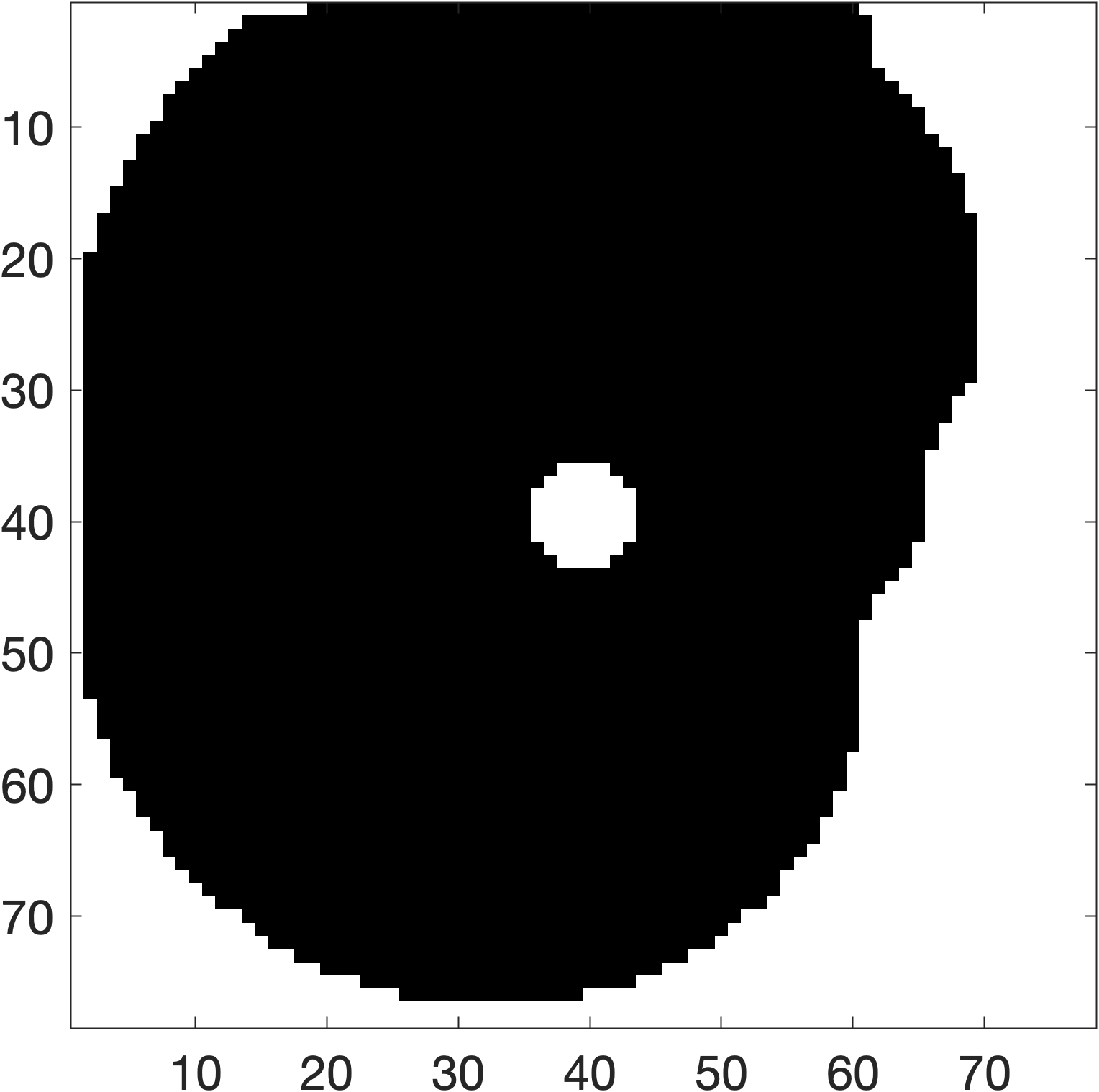}
        \caption{$\gamma_2 = 0.25$}

    \end{subfigure}
        \begin{subfigure}[t]{.3\linewidth}
        \includegraphics[width=0.9\linewidth]{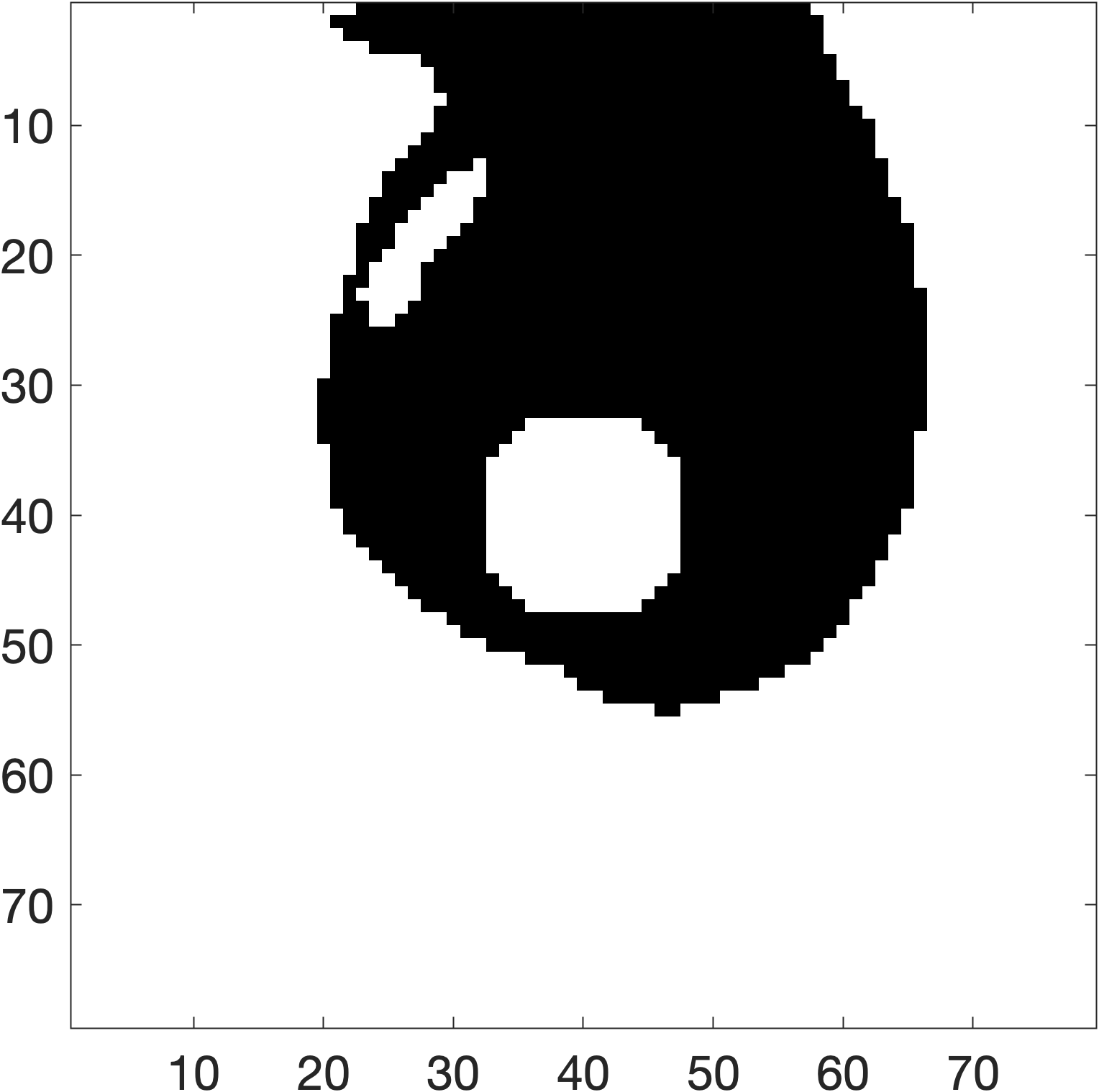}
        \caption{$\gamma_3 = 1$}
    \end{subfigure}

        \begin{subfigure}[t]{.3\linewidth}
        \includegraphics[width=0.9\linewidth]{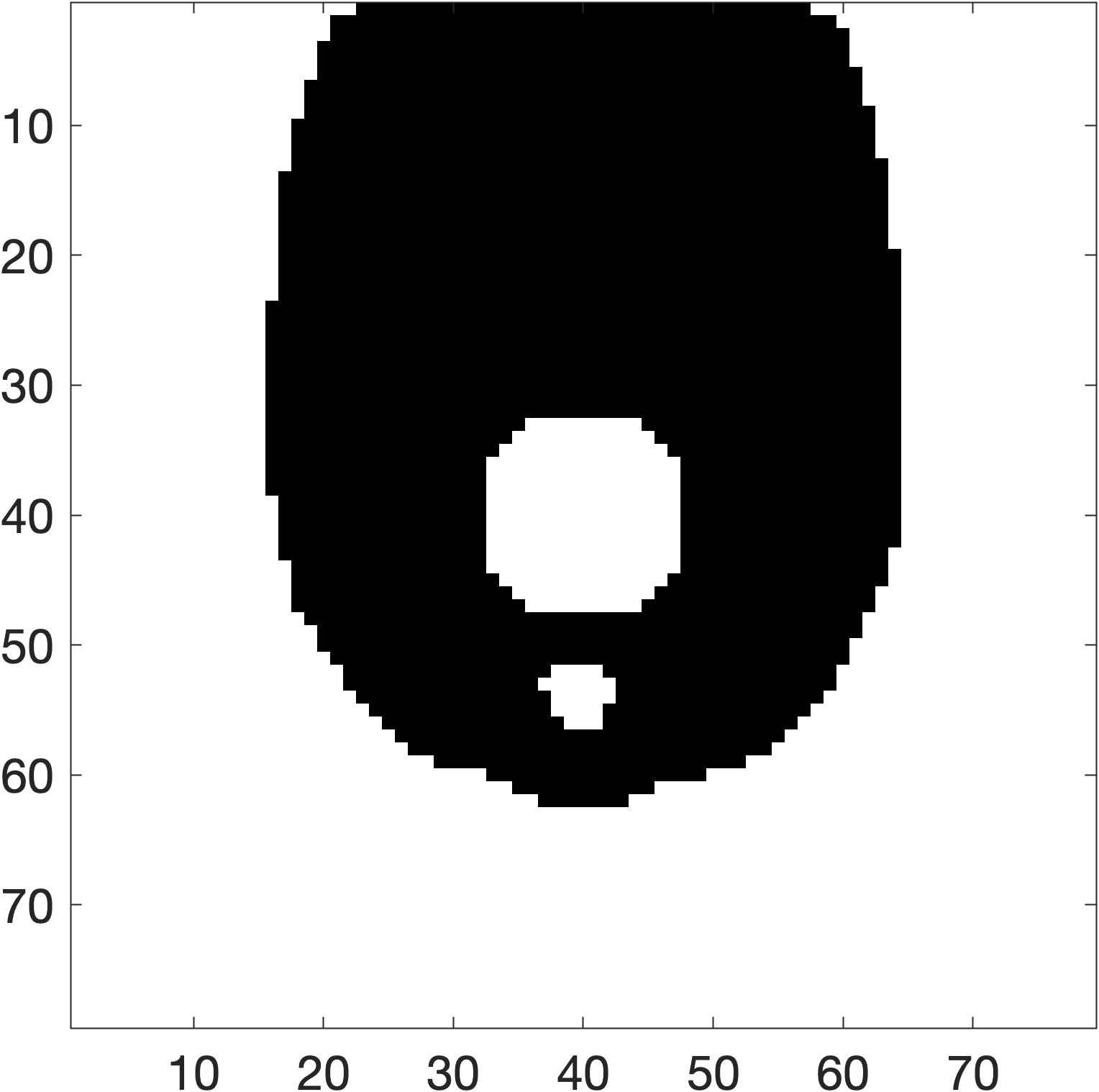}
        \caption{$\gamma_1 = 1$}

    \end{subfigure}
    \begin{subfigure}[t]{.3\linewidth}
        \includegraphics[width=0.9\linewidth]{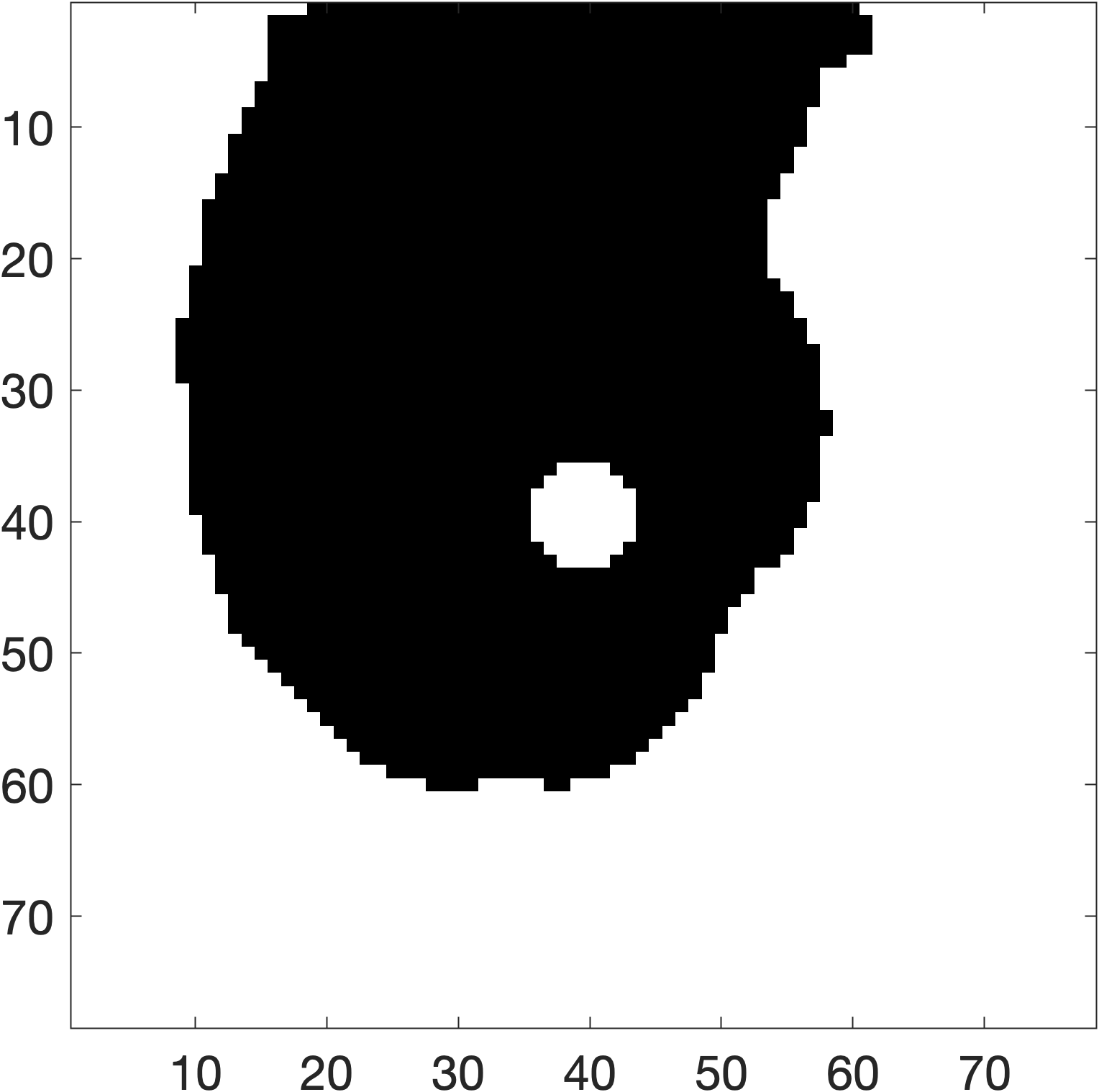}
        \caption{$\gamma_2 = 1$}

    \end{subfigure}
        \begin{subfigure}[t]{.3\linewidth}
        \includegraphics[width=0.9\linewidth]{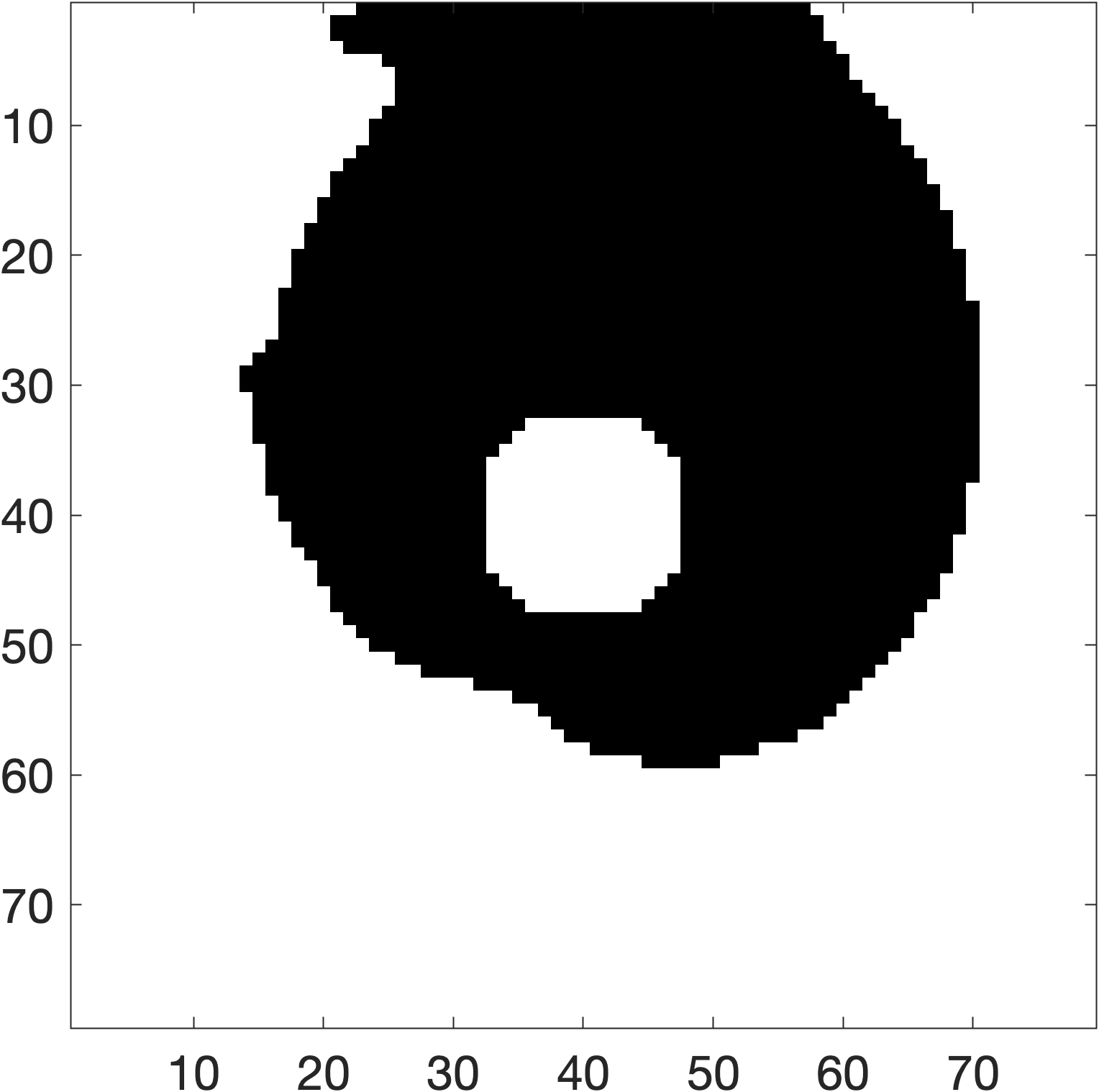}
        \caption{$\gamma_3 = 1$}
    \end{subfigure}
    
    \centering
    \caption{Co-optimized three rotating bodies with different volume decrement ratio $\gamma_i$.}
    \label{fig_threeSquareOptDesigns}
\end{figure}

\begin{figure} [h!]
    \centering
    \includegraphics[width=\linewidth]{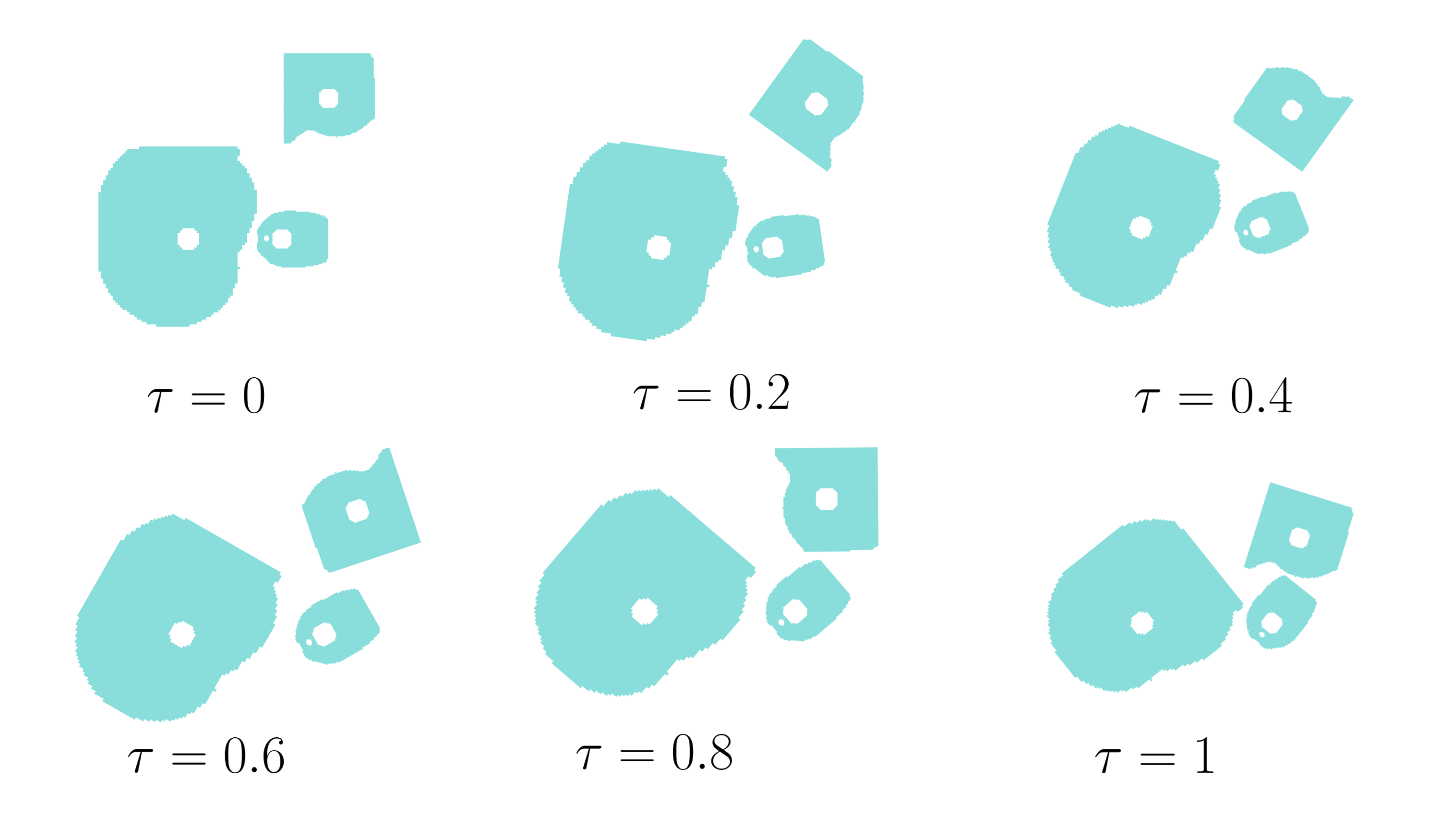}
    \caption{Co-optimized three rotating bodies system with ($\lambda_{g_i} = 0.5$ and $\gamma_1 = 1$, $\gamma_2 = 0.5$, and $\gamma_3 = 0.25$) to simultaneously optimize for stiffness and collision avoidance. }
    \label{fig_threeBodyMotion}
\end{figure}

\begin{figure} [h!]
    \centering
    \begin{subfigure}[t]{\linewidth}
        \includegraphics[width=\linewidth]{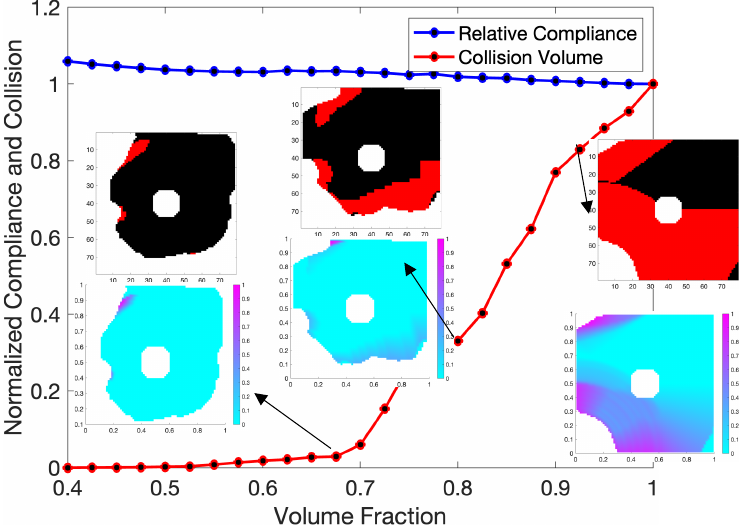}
        \caption{}

    \end{subfigure}
    \begin{subfigure}[t]{\linewidth}
        \includegraphics[width=\linewidth]{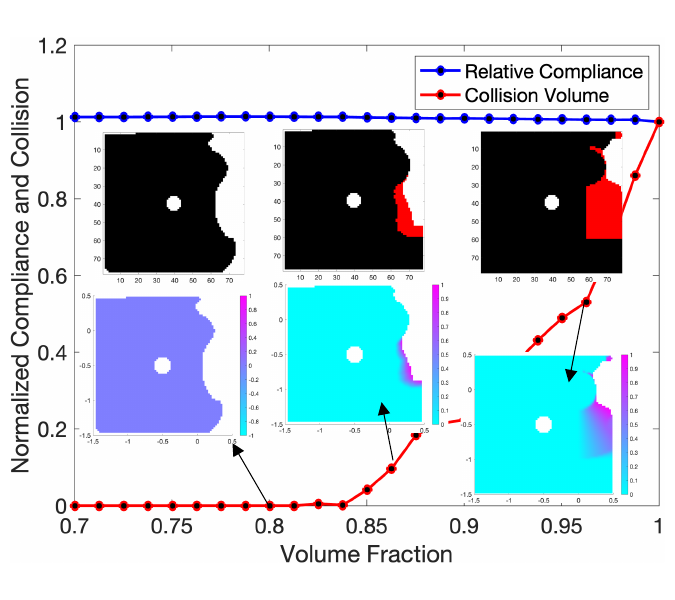}
        \caption{}

    \end{subfigure}
        \begin{subfigure}[t]{\linewidth}
        \includegraphics[width=\linewidth]{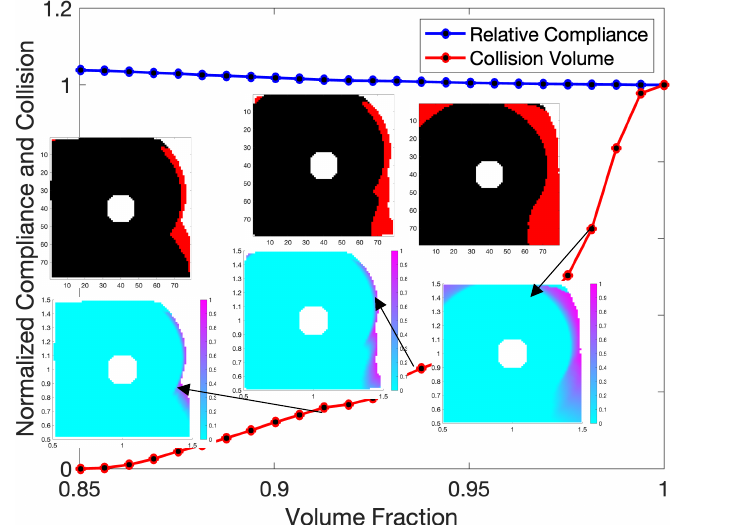}
        \caption{}
    \end{subfigure}
    \centering
    \caption{Convergence for co-optimization of the three-body system to a set of collision-free shapes.}
    \label{fig_threeSquareConvergence}
\end{figure}

As the second example, let us consider the three-body system of \Cref{fig_threeSquareConfig}, where $\theta_1= [\pi/2, \pi/5]$, $\theta_2 = [0,3\pi/10]$, $\theta_3 = [\pi/2,-3\pi/2]$ with 
500 steps temporal resolution for collision analysis. The design domains and their corresponding boundary conditions are shown in \cref{fig_threeSquareBC}, where all designs are assumed to be fixed at the center hole. The external forces are applied at top-middle points with $f^1_{ext} = [1,0]$, $f^2_{ext} = [1,1]$, and $f^3_{ext} = [-1,1]$. All designs are discretized into 6,000 elements. \\
\Cref{fig_threeSquareCollFields} illustrates the pair-wise collision measure fields and the overall collision measure fields for the initial domains. To demonstrate the impact of volume decrement ration on the final assembly, \Cref{fig_threeSquareOptDesigns} shows the co-optimized system considering three different scenarios, 1) $\gamma_1 = 1$, $\gamma_2 = 0.5$, $\gamma_3 = 0.25$, 2) $\gamma_1 = 0.5$, $\gamma_2 = 0.25$, $\gamma_3 = 1$, and 3) $\gamma_1 = 1$, $\gamma_2 = 1$, $\gamma_3 = 1$. The final volume fractions and compliance values are summarized in \Cref{table_threeSquareResults}.\\
\Cref{fig_threeBodyMotion} shows the collision-free motion of the co-optimized system for the first scenario at six different snapshots. The evolution of compliance and collision volume for all components are illustrated in \Cref{fig_threeSquareConvergence}.

% ================= Iter:   25, decrements: 1         0.5        0.25, collision weight: 0.5 =================
% Part1, Objective: 1.06, Volume: 0.40, Collision Volume: 0.00, Deformation: 2.6421e-09, von Mises: 0.13085
% Part2, Objective: 1.01, Volume: 0.70, Collision Volume: 0.00, Deformation: 2.7735e-08, von Mises: 0.65741
% Part3, Objective: 1.04, Volume: 0.85, Collision Volume: 0.00, Deformation: 2.06e-08, von Mises: 0.6387

\begin{table*}[!ht]
\centering
\caption{Impact of $\gamma_i$ on compliance and collision for the three-body system.}
\begin{tabular}{ccc|ccc|ccc|ccc}
\hline \hline 
{$\gamma_{1}$}& {$\gamma_{2}$}& {$\gamma_{3}$} &$v_1$&$v_2$&$v_3$& $f_1/f_1^0$ & $f_2/f_2^0$ &$f_3/f_3^0$ & $\Gield_1$ & $\Gield_2$ &$\Gield_3$ \\ \hline
1&1&1& 0.40&0.40&0.40&1.06&1.07&1.17 & 0.00 &0.00&0.00 \\ 
0.5&0.25&1& 0.40&0.70&0.30&1.05&1.01&1.36 & 0.00 &0.00 &0.00  \\
1&0.5&0.25& 0.40&0.70&0.85&1.06&1.01&1.04 & 0.00 &0.00 &0.00  \\ 
% 1&0.25&1& 0.55&0.90&0.55&1.07&1.01&1.07 & 0.00 &0.00 &0.00  \\ 
\end{tabular}

\label{table_threeSquareResults}
\end{table*}

%%%%%%%%%%%%
%%%%%%%%%%%%
\subsection{Gripper and Cams Assembly}
\begin{figure} [h!]
    \centering
    \begin{subfigure}[t]{0.6\linewidth}
    \centering
        \includegraphics[width=0.9\linewidth]{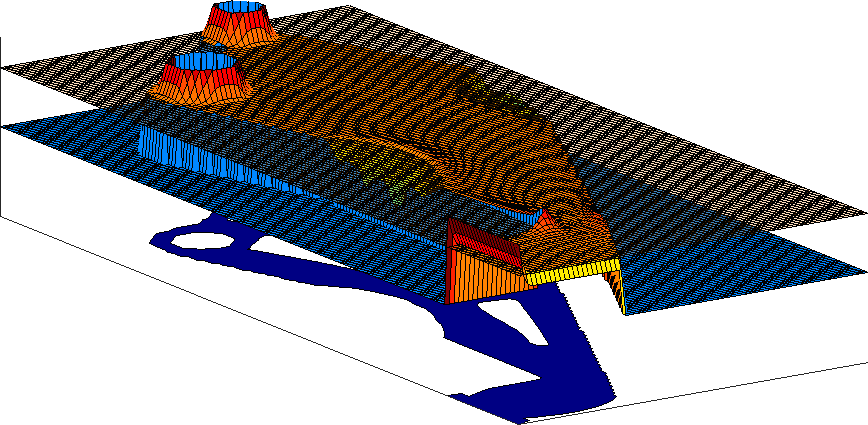}
        \caption{Gripper augmented TSF}
        \label{fig_GripperTSFWt5}
    \end{subfigure}
    \begin{subfigure}[t]{0.35\linewidth}
    \centering
        \includegraphics[width=0.5\linewidth]{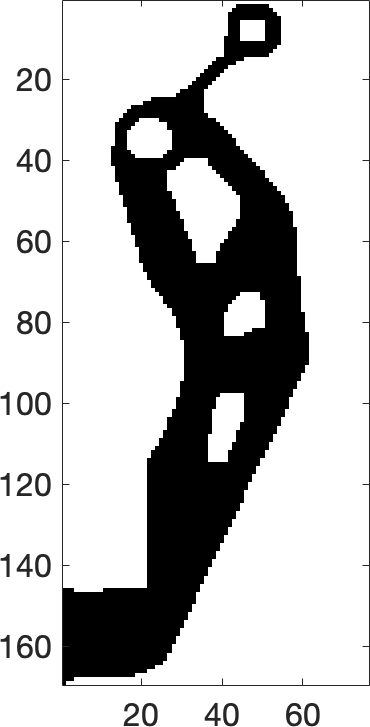}
        \caption{Optimized gripper at 0.6 volume fraction.}
        \label{fig_GripperOptWt5}
    \end{subfigure}

     \begin{subfigure}[t]{.6\linewidth}
        \includegraphics[width=0.7\linewidth]{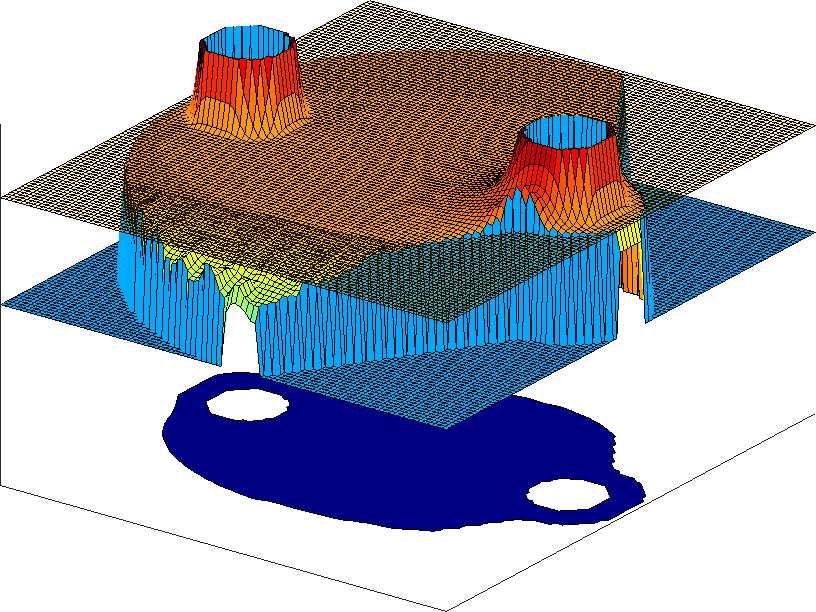}
        \caption{Cam \#1 augmented TSF.}
        \label{fig_cam1TSFWt5}
    \end{subfigure}
    \begin{subfigure}[t]{.35\linewidth}
        \includegraphics[width=0.7\linewidth]{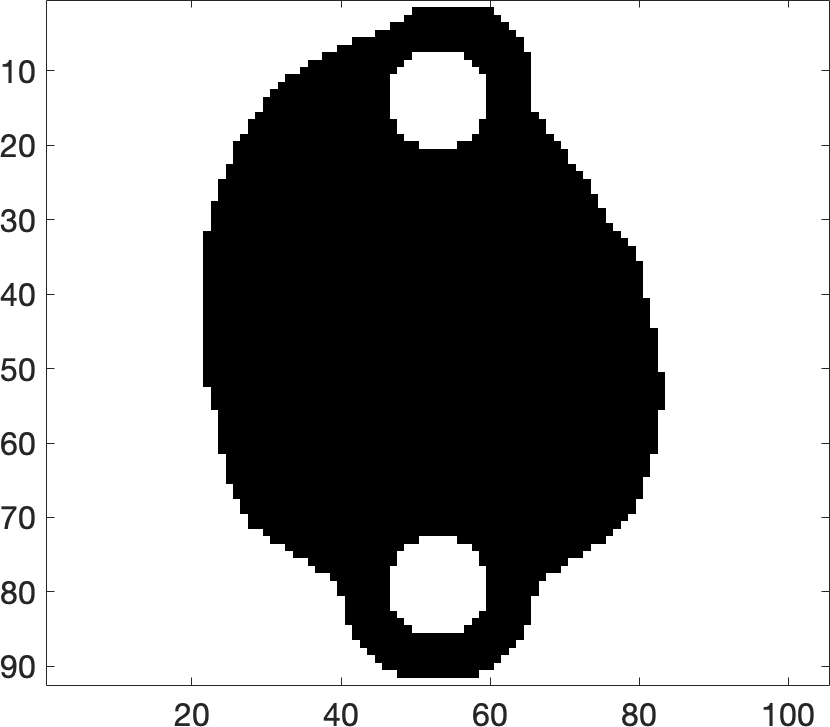}
        \caption{Optimized cam \#1 at 0.6 volume fraction.}
        \label{fig_cam1OptWt5}
    \end{subfigure}

    \begin{subfigure}[t]{.6\linewidth}
        \includegraphics[width=0.7\linewidth]{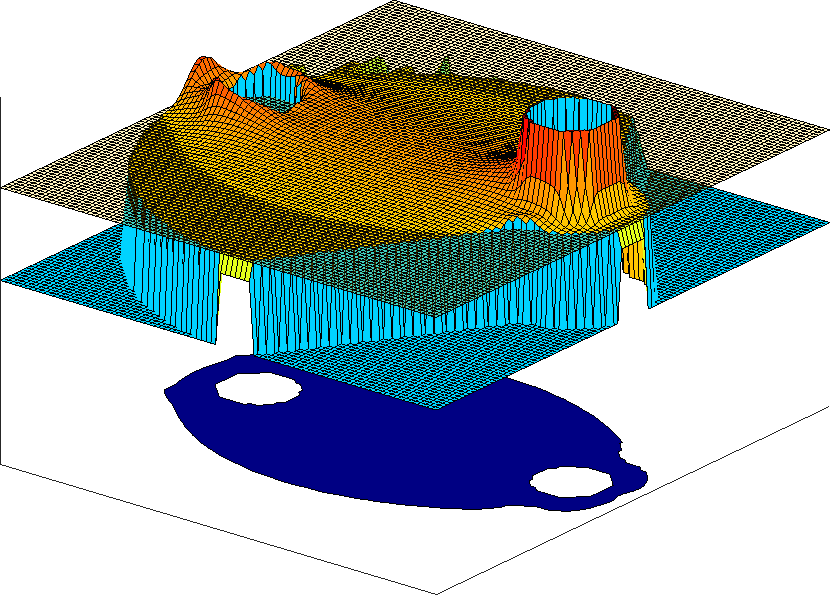}
        \caption{Cam \#2 augmented TSF.}
        \label{fig_cam2TSFWt5}
    \end{subfigure}
    \begin{subfigure}[t]{.35\linewidth}
        \includegraphics[width=0.85\linewidth]{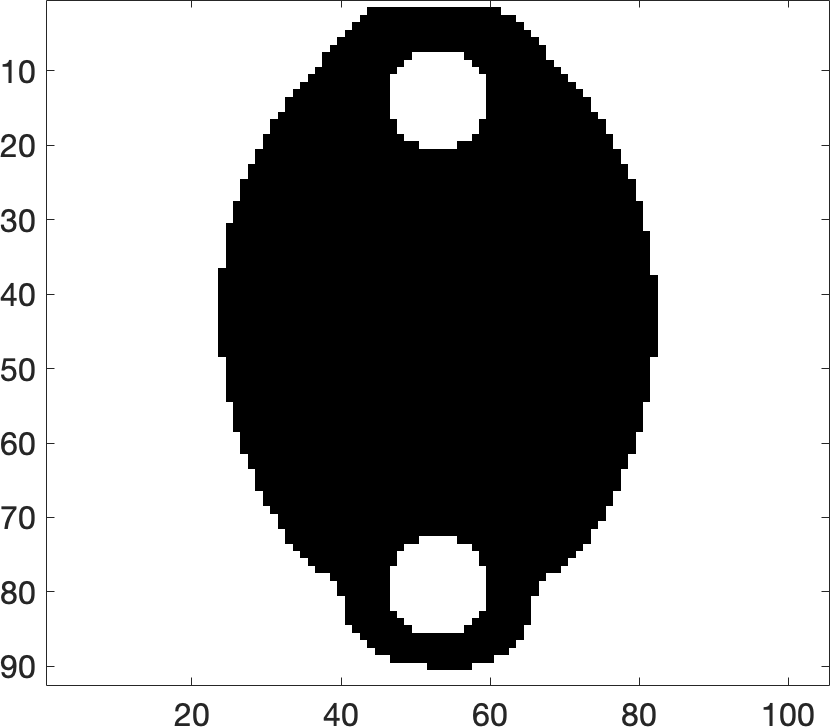}
        \caption{Optimized cam \#2 at 0.6 volume fraction.}
        \label{fig_cam2OptWt5}
    \end{subfigure}
    \centering
    \caption{Co-optimized designs at 0.6 volume fraction for all components \textit{with} considering collision avoidance constraint ($\lambda_{g_i} = 0.5$ and $\gamma_i = 1$, $\forall i$). }
    \label{fig_GripperCamsOptWt5}
\end{figure}

\begin{figure} [h!]
    \centering
    \includegraphics[width=\linewidth]{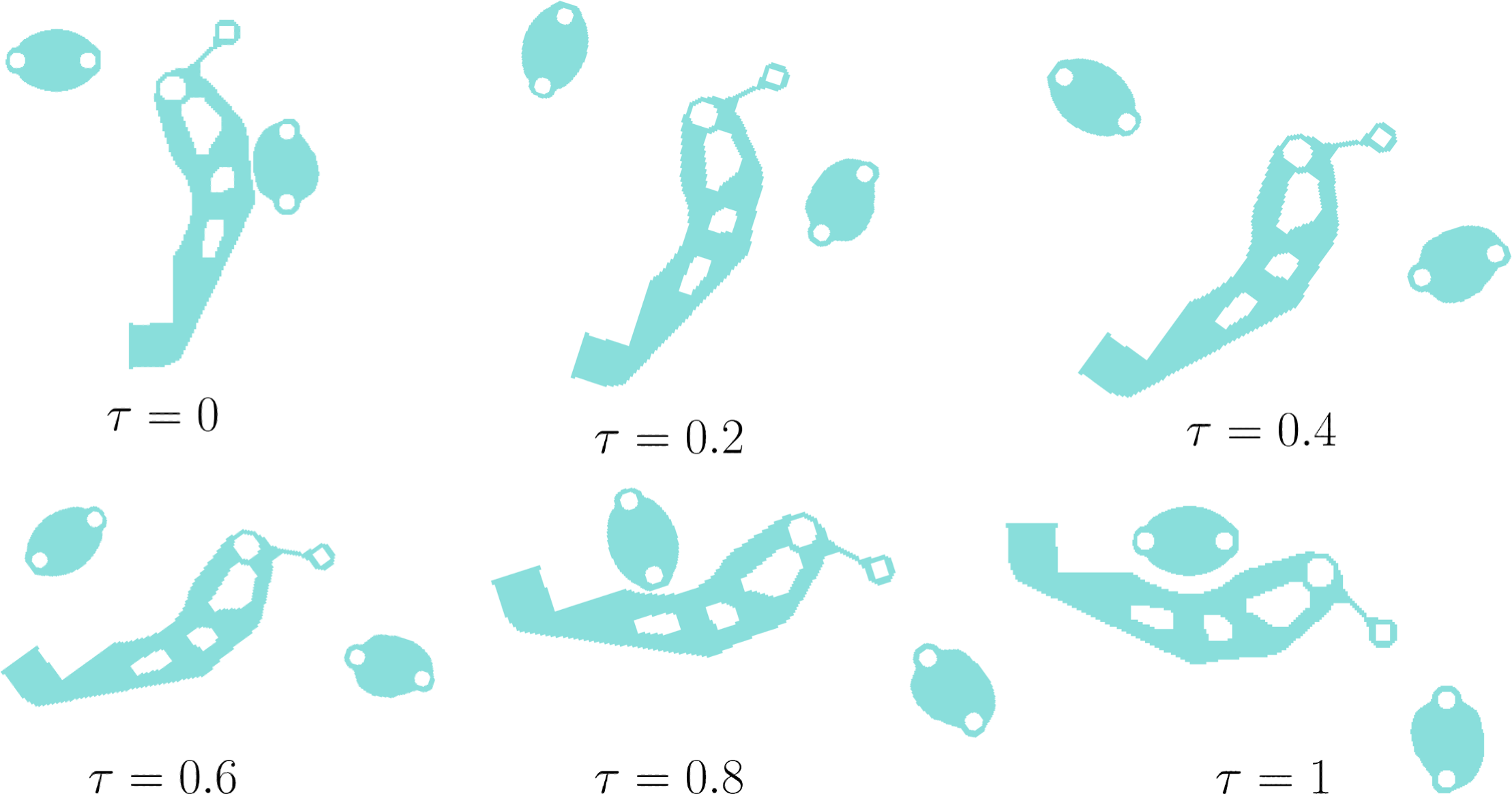}
    \caption{Co-optimized gripper-cams system with ($\lambda_{g_i} = 0.5$ and $\gamma_i = 1$, $\forall i$) to simultaneously optimize for stiffness and collision avoidance. }
    \label{fig_gripperOptMotionSnapshots}
\end{figure}

Next, let us revisit the system of \Cref{fig_GripperCamsMotion} comprising a gripper and two cams with $\theta_1=[0,\pi/2]$, $\theta_2=[0,\pi]$, $\theta_3=[\pi/2,-3\pi/2]$. The temporal resolution is 500 time steps and all components are discretized into 6,000 finite elements. The boundary conditions are shown in \Cref{fig_GripperCamsBC}.\\
Let us consider a case with $\lambda_{g_i} = 0.5$ and $\gamma_i = 1$ for all components. We reach a collision-free configuration at volume fraction of 0.6 for all designs. 
Figures \ref{fig_GripperTSFWt5}, \ref{fig_cam1TSFWt5}, and \ref{fig_cam2TSFWt5} illustrate the final augmented sensitivity field and the corresponding level-sets, which gives the iso-surfaces of the optimized designs. Figures \ref{fig_GripperOptWt5}, \ref{fig_cam1OptWt5}, and \ref{fig_cam2OptWt5} show the optimized grids. The optimized gripper has $f_1/f^0_1 = 1.30$ with maximum deformation of $1.7e-4$ (m) and maximum von Mises stress of $0.88$ (MPa). The two optimized cams have $f_2/f^0_2 = 1.02$ and $f_3/f^0_3 = 1.03$. The optimized cam 1 has maximum deformation of $1.1e-6$ (m) and maximum von Mises stress of $1.3e-5$ (MPa). The optimized cam 2 has maximum deformation of $6.4e-8$ (m) and maximum von Mises stress of $1.9e-6$ (MPa).
\Cref{fig_gripperOptMotionSnapshots} shows the collision-free motion of the optimized design at multiple snapshots.
%       ================= Iter:   40, decrements: 1  1  1, collision weight: 0.5 =================
% Part1, Objective: 1.30, Volume: 0.61, Collision Volume: 0.01, Deformation: 0.00017432, von Mises: 881.6419
% Part2, Objective: 1.02, Volume: 0.61, Collision Volume: 0.00, Deformation: 1.1444e-06, von Mises: 13.8104
% Part3, Objective: 1.03, Volume: 0.61, Collision Volume: 0.00, Deformation: 6.3775e-08, von Mises: 1.8958  

\begin{figure} [h!]
    \centering
    \begin{subfigure}[t]{0.65\linewidth}
        \includegraphics[width=\linewidth]{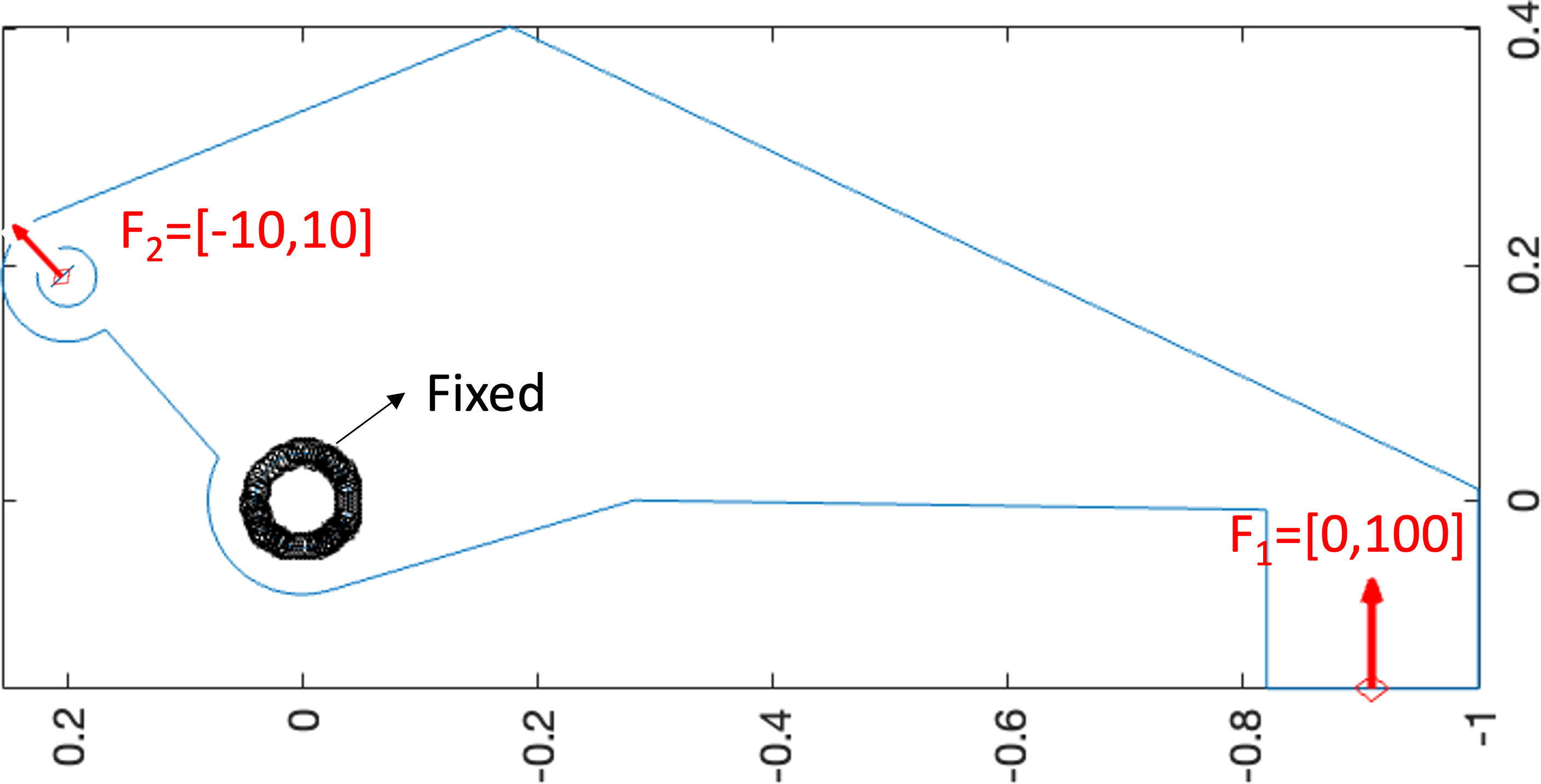}
        \caption{Gripper.}
        \label{fig_gripperBC}
    \end{subfigure}
    
    \begin{subfigure}[t]{.45\linewidth}
        \includegraphics[width=0.8\linewidth]{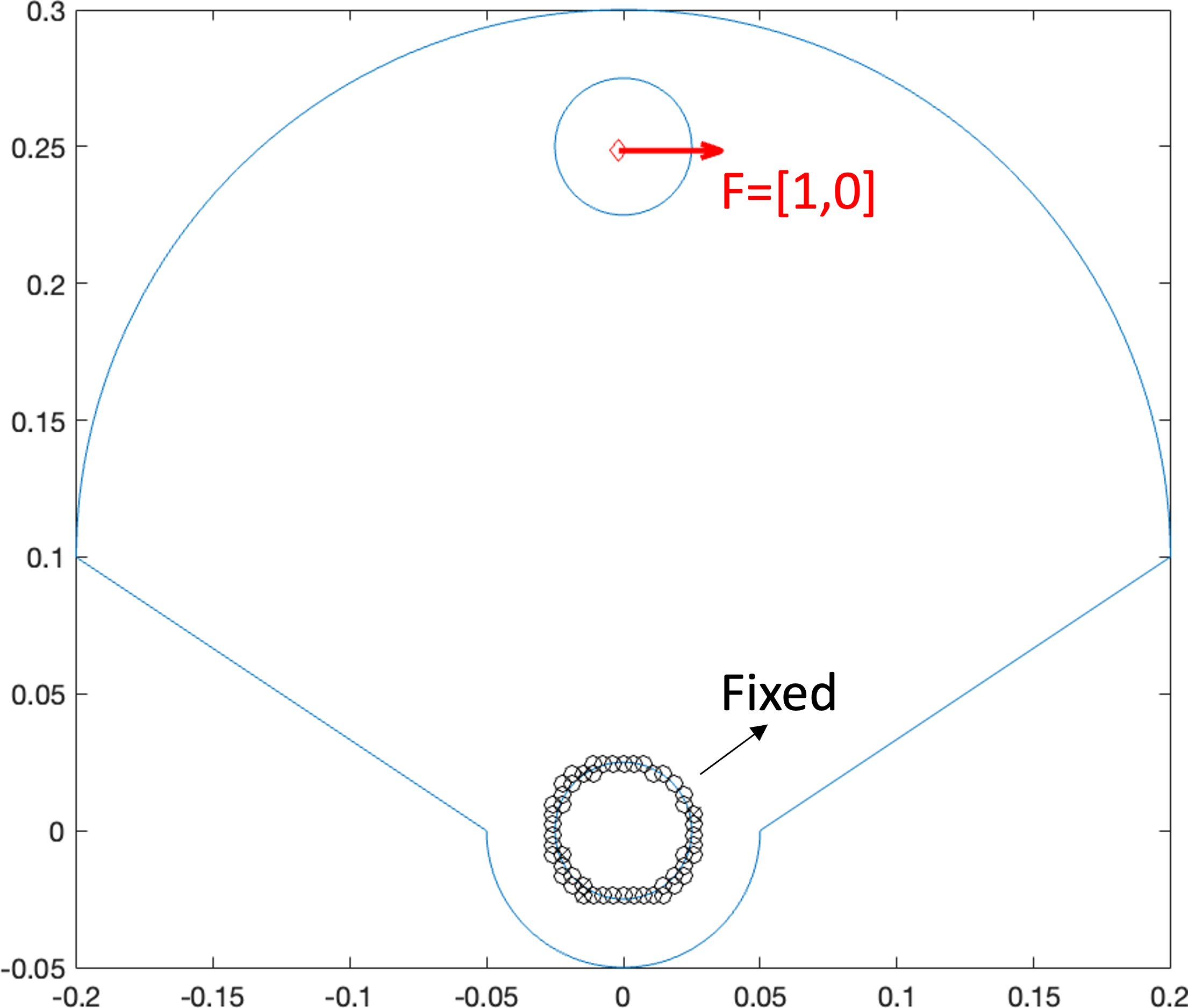}
        \caption{Cam \#1.}
        \label{fig_camBC_1}
    \end{subfigure}
        \begin{subfigure}[t]{.45\linewidth}
        \includegraphics[width=0.8\linewidth]{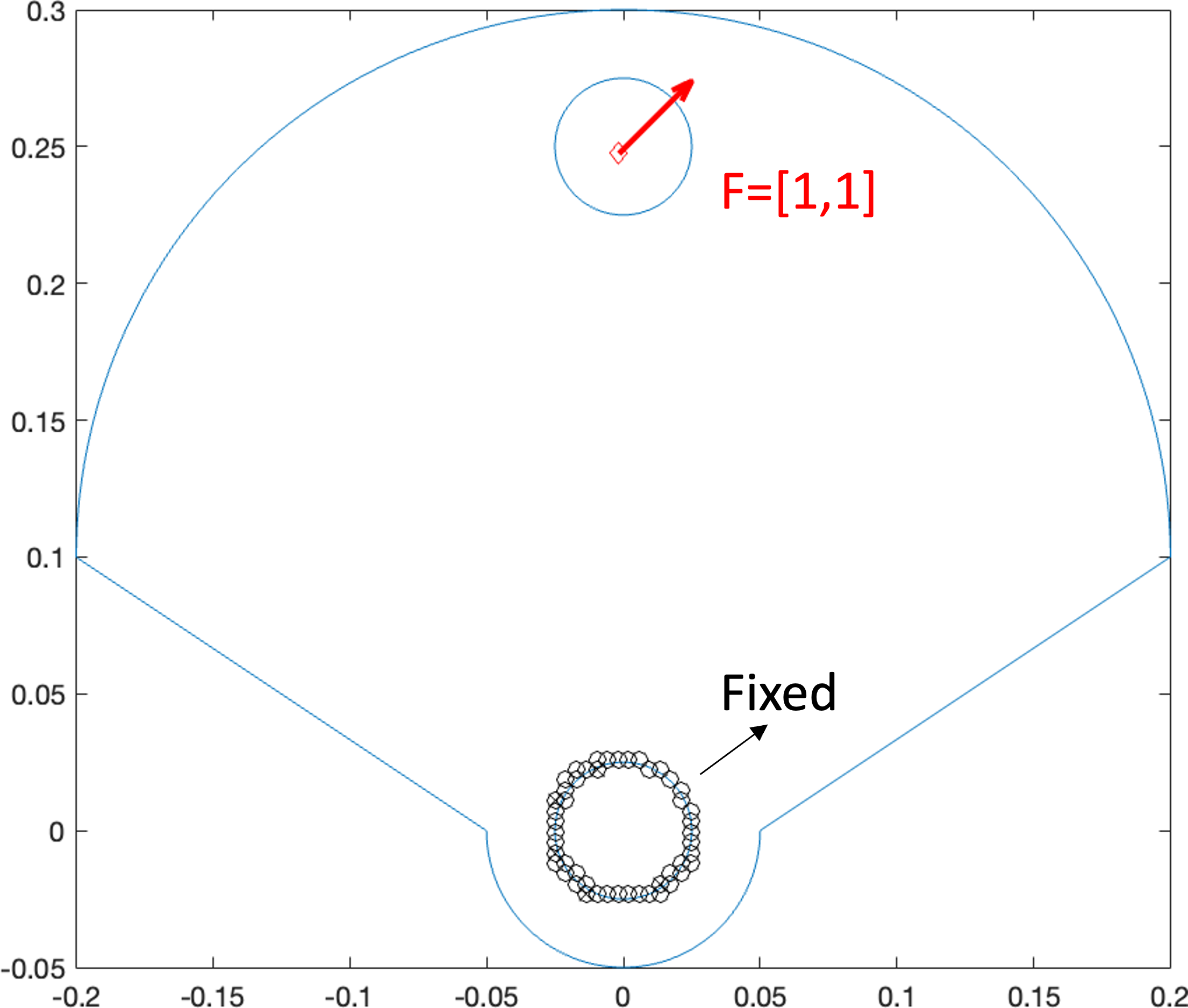}
        \caption{Cam \#2.}
        \label{fig_camBC_2}
    \end{subfigure}
    \centering
    \caption{Gripper and two cams boundary conditions.}
    \label{fig_GripperCamsBC}
\end{figure}

Table \ref{table_compTimes} summarizes the computation time for the gripper and cams example for the entire optimization process. Observe that the majority of the computation time is spent on evaluating the physical performance of the parts and FEA remains the bottleneck. The one-time computation of the CWMs takes about 18 seconds at the pre-processing stage. On the other hand, the evaluation of the collision measures during the optimization loop relies on fast matrix multiplications and only takes about 1 second in total, while compliance gradient computation takes about 8 seconds.    

\begin{table}[!ht]
\centering
\caption{\textcolor{red}{Computation times in seconds.}}
\begin{tabular}{lc}
\hline \hline 
Operation & Time (s) \\ \hline
FEA & 88.43 \\
Compliance Gradient & 8.01 \\
Collision Weights (pre-process) & 18.05 \\
Collision Measure & 1.09 \\ 
Overall & 121.75
\end{tabular}

\label{table_compTimes}
\end{table}

\subsection{Three-Body System in 3D}
\begin{figure} [h!]
    \centering
    \begin{subfigure}[t]{.3\linewidth}
        \includegraphics[width=\linewidth]{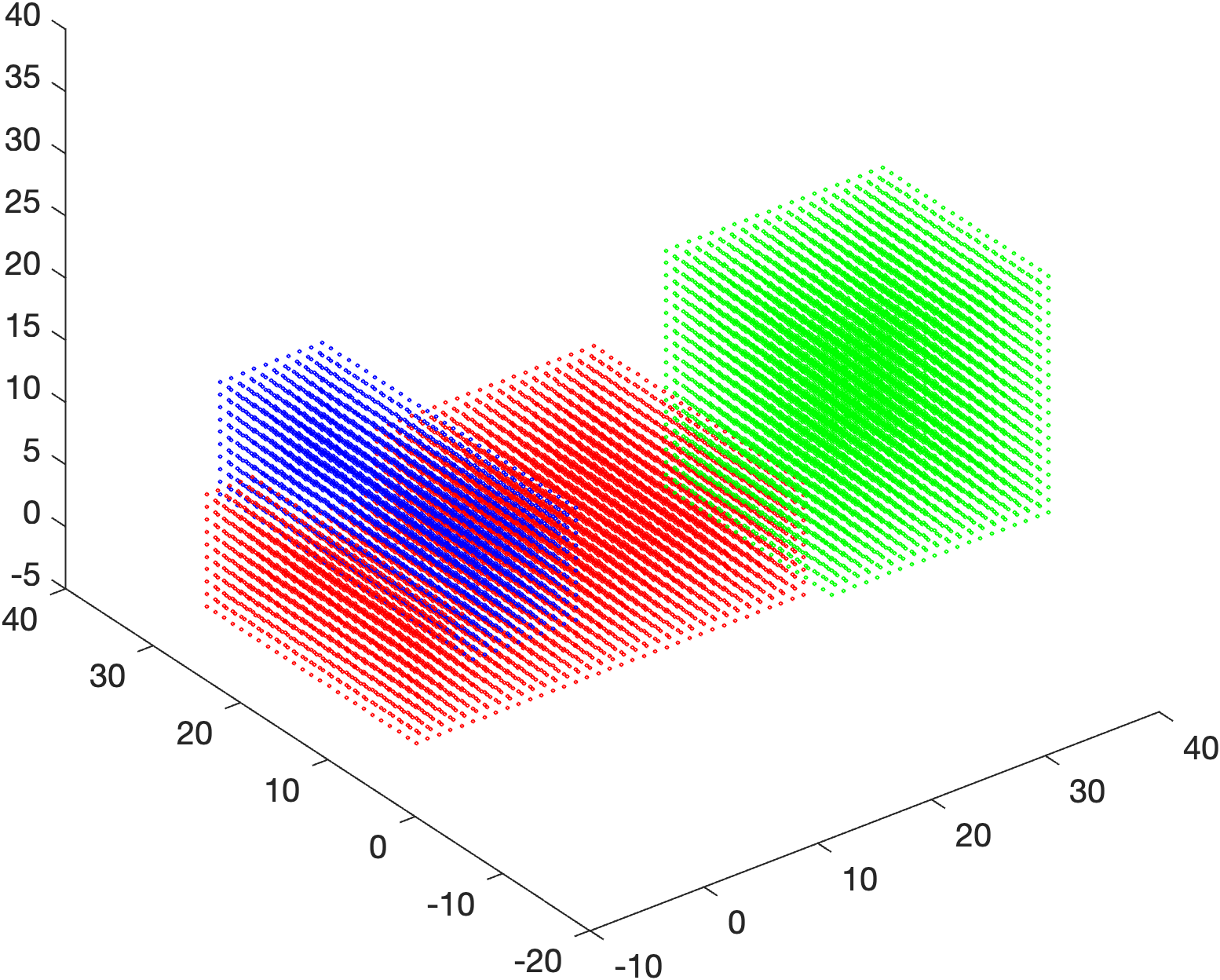}
        \caption{$\tau = 0$}
        \label{fig_3d_t1}
    \end{subfigure}
        \begin{subfigure}[t]{.3\linewidth}
        \includegraphics[width=\linewidth]{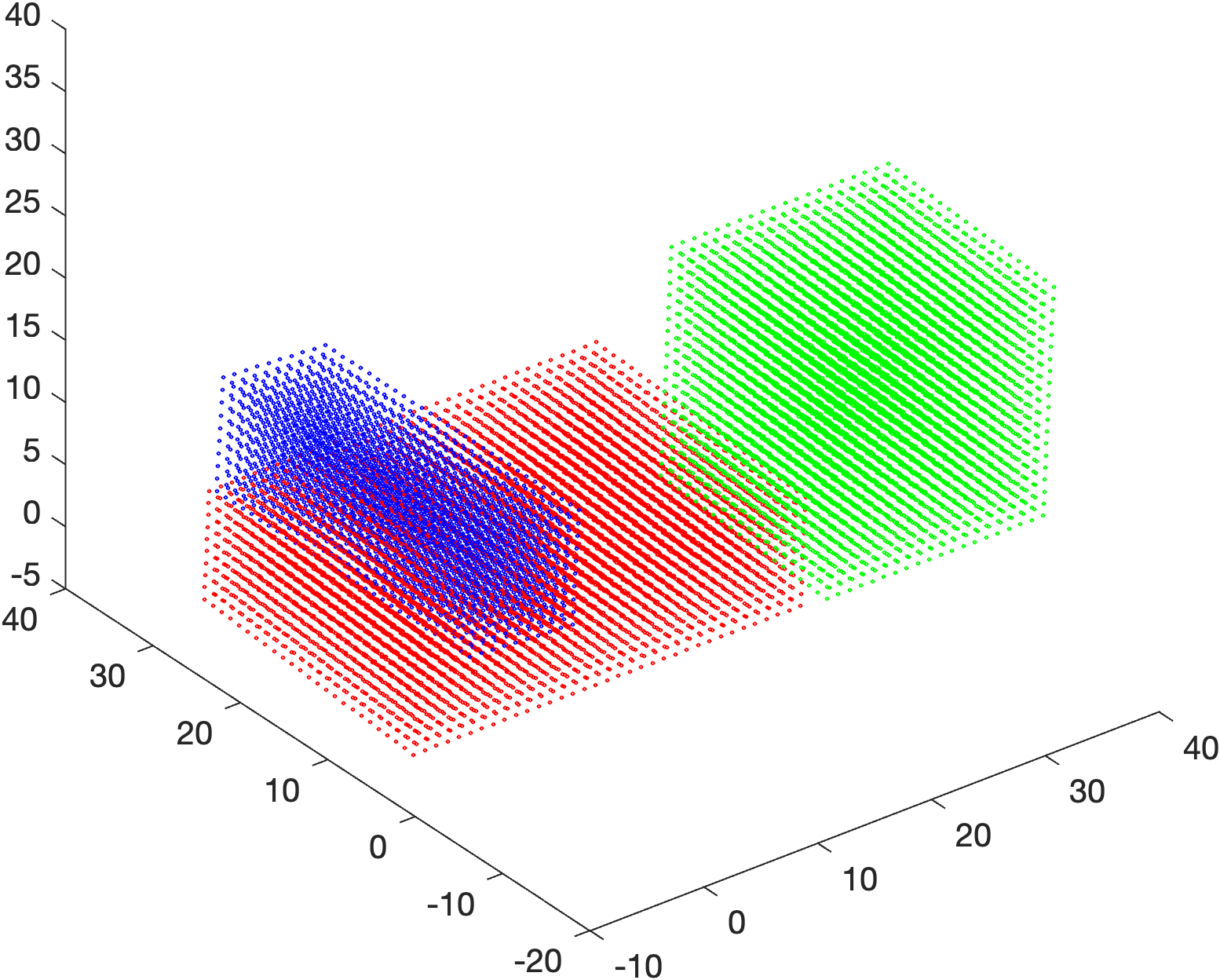}
        \caption{$\tau = 0.2$}
        \label{fig_3d_t1}
    \end{subfigure}
        \begin{subfigure}[t]{.3\linewidth}
        \includegraphics[width=\linewidth]{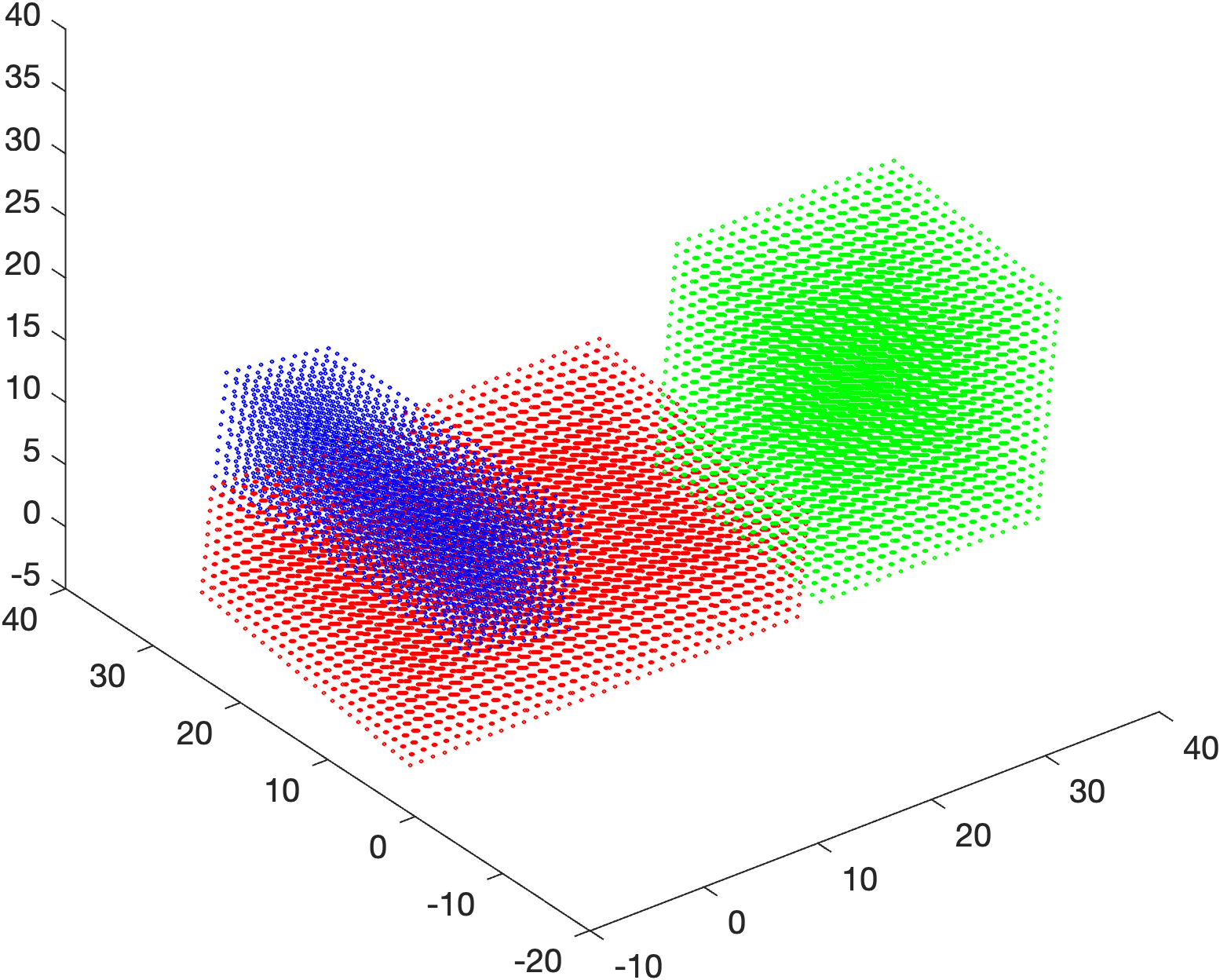}
        \caption{$\tau = 0.4$}
        \label{fig_3d_t1}
    \end{subfigure}
   \begin{subfigure}[t]{.3\linewidth}
        \includegraphics[width=\linewidth]{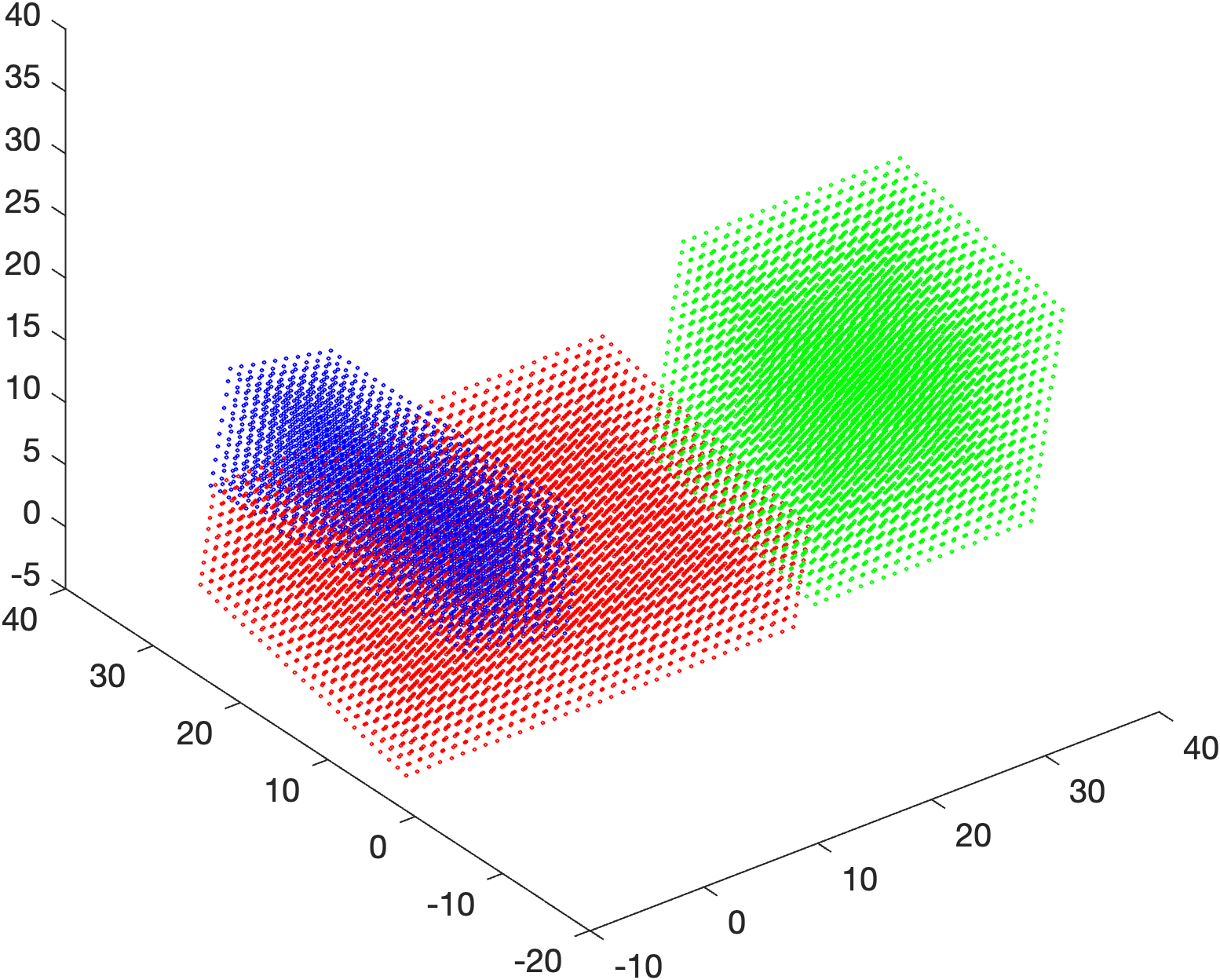}
        \caption{$\tau = 0.6$}
        \label{fig_3d_t1}
    \end{subfigure}
        \begin{subfigure}[t]{.3\linewidth}
        \includegraphics[width=\linewidth]{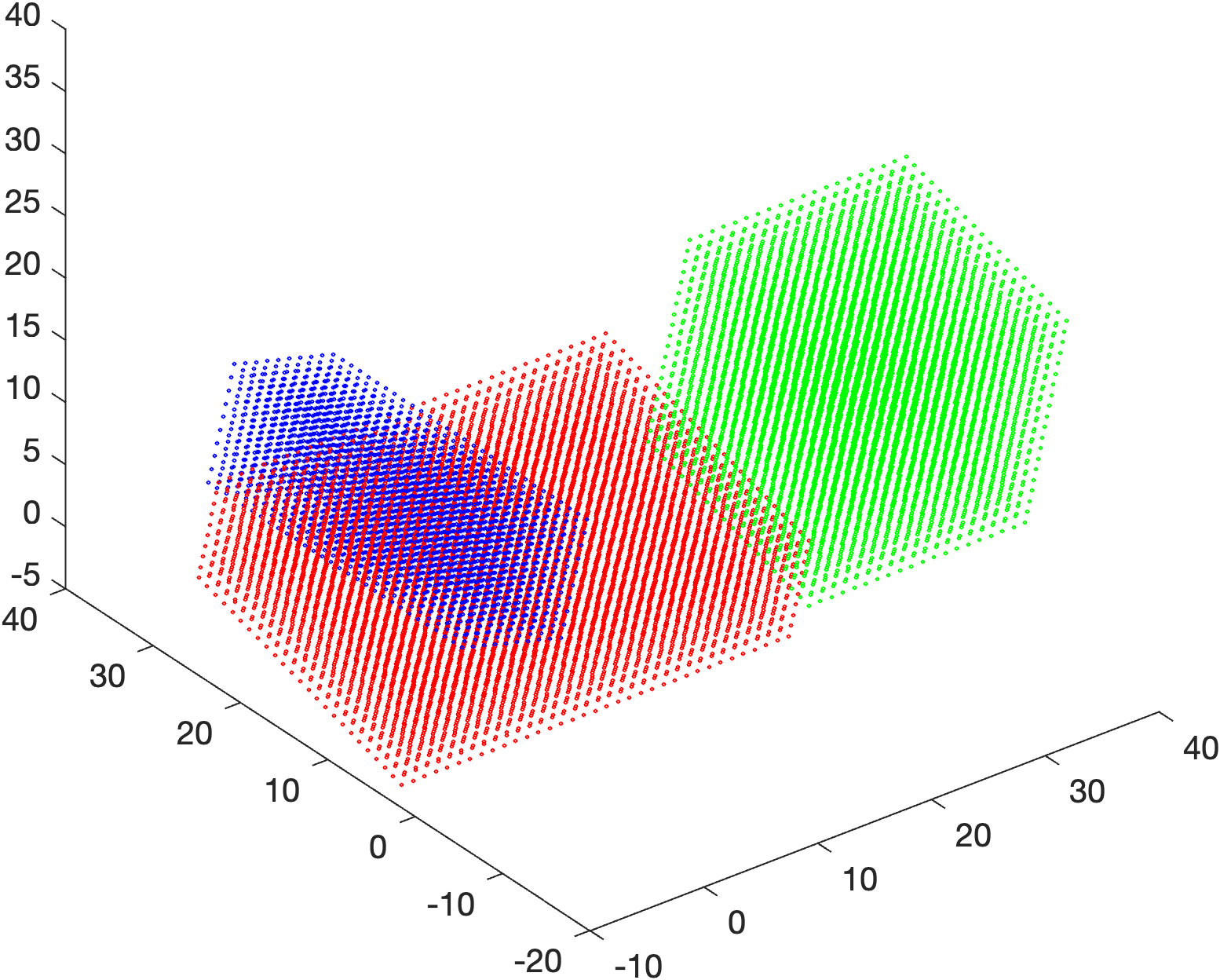}
        \caption{$\tau = 0.8$}
        \label{fig_3d_t1}
    \end{subfigure}
        \begin{subfigure}[t]{.3\linewidth}
        \includegraphics[width=\linewidth]{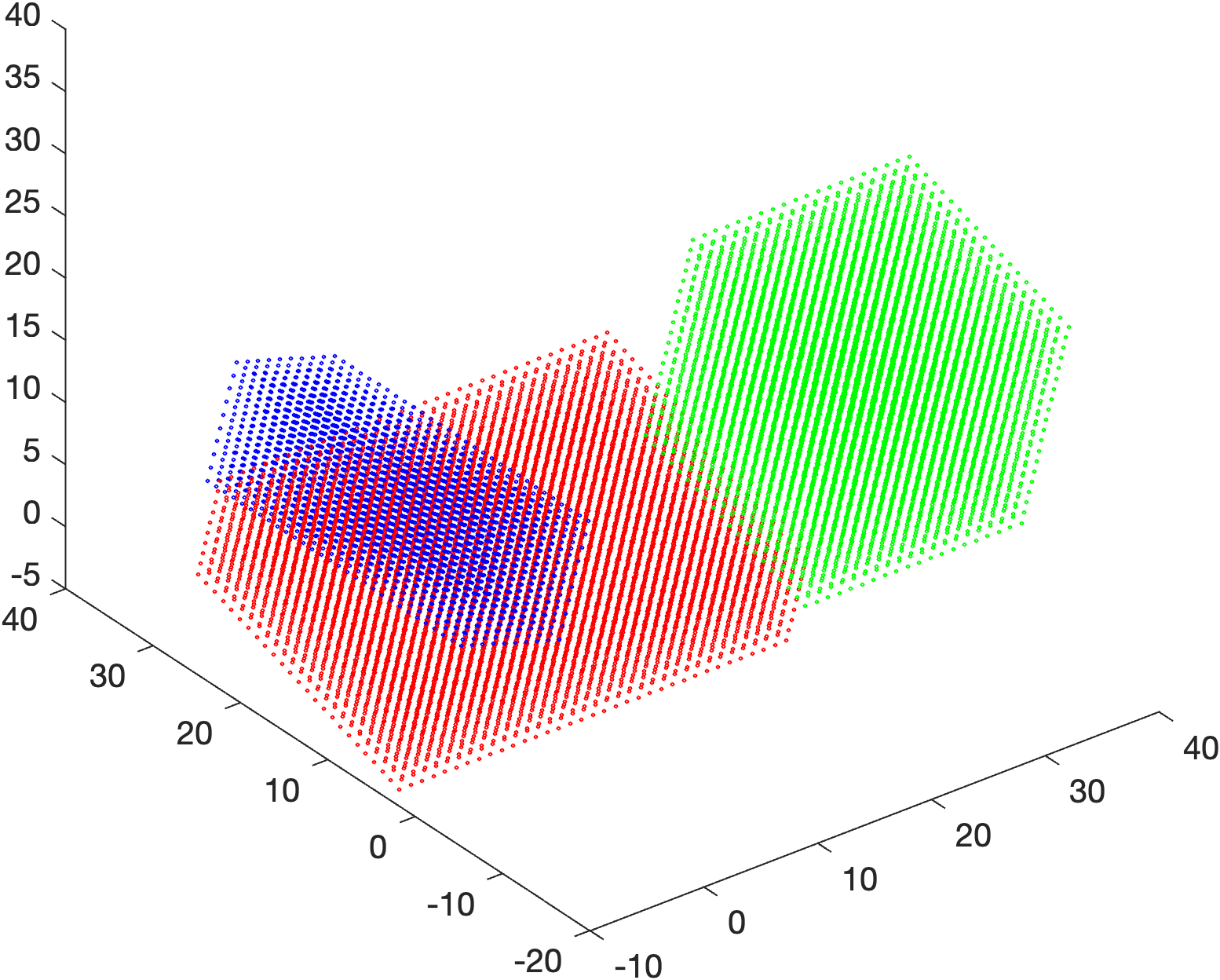}
        \caption{$\tau = 1$}
        \label{fig_3d_t1}
    \end{subfigure}
    \centering
    \caption{Three-body system motion snapshots.}
    \label{fig_3dMotions}
\end{figure}

\begin{figure} [h!]
    \centering
    \begin{subfigure}[t]{.3\linewidth}
        \includegraphics[width=\linewidth]{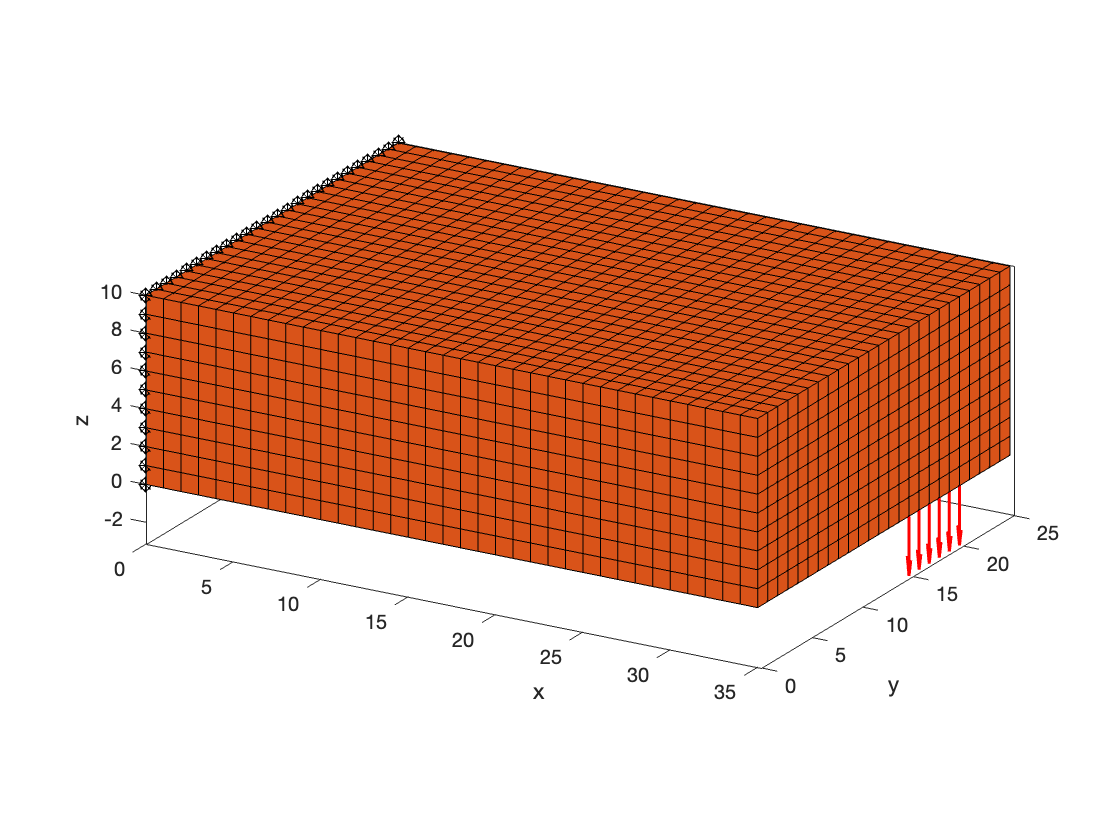}
        \caption{Part 1 loading.}
        \label{fig_3dBC1}
    \end{subfigure}
    \begin{subfigure}[t]{.3\linewidth}
        \includegraphics[width=\linewidth]{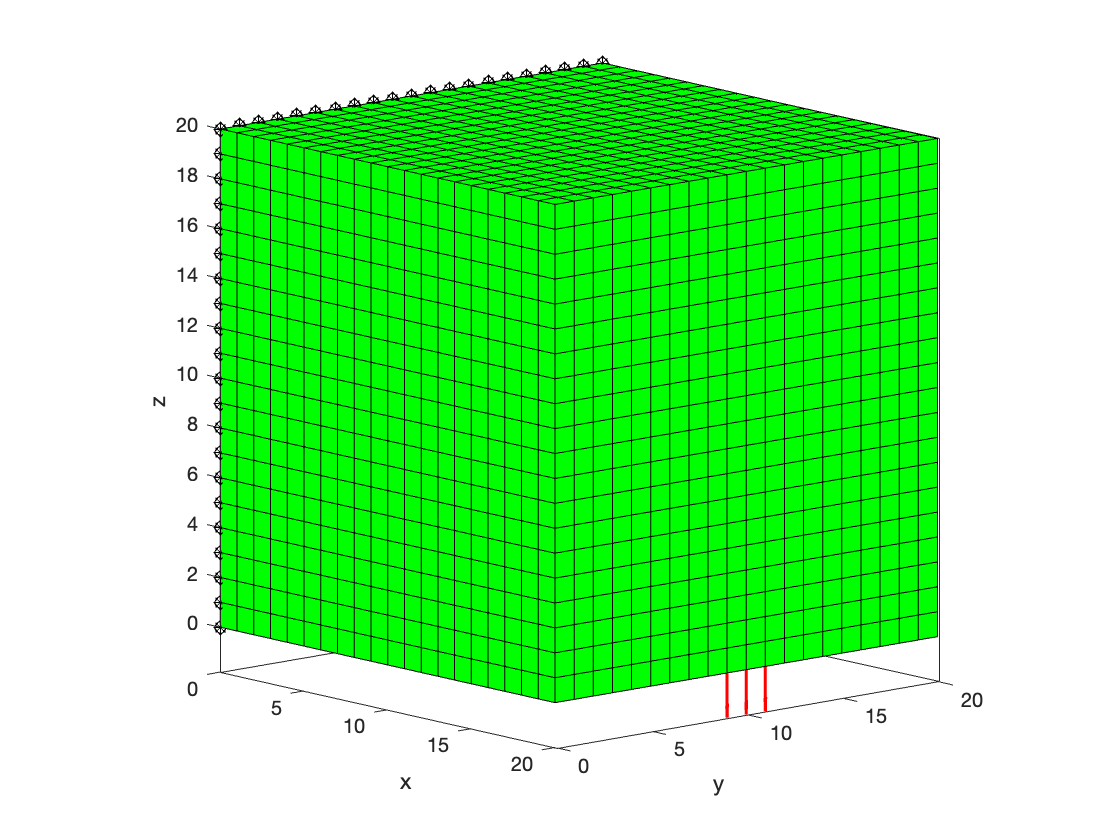}
        \caption{Part 2 loading.}
       \label{fig_3dBC2}
    \end{subfigure}
        \begin{subfigure}[t]{.3\linewidth}
        \includegraphics[width=\linewidth]{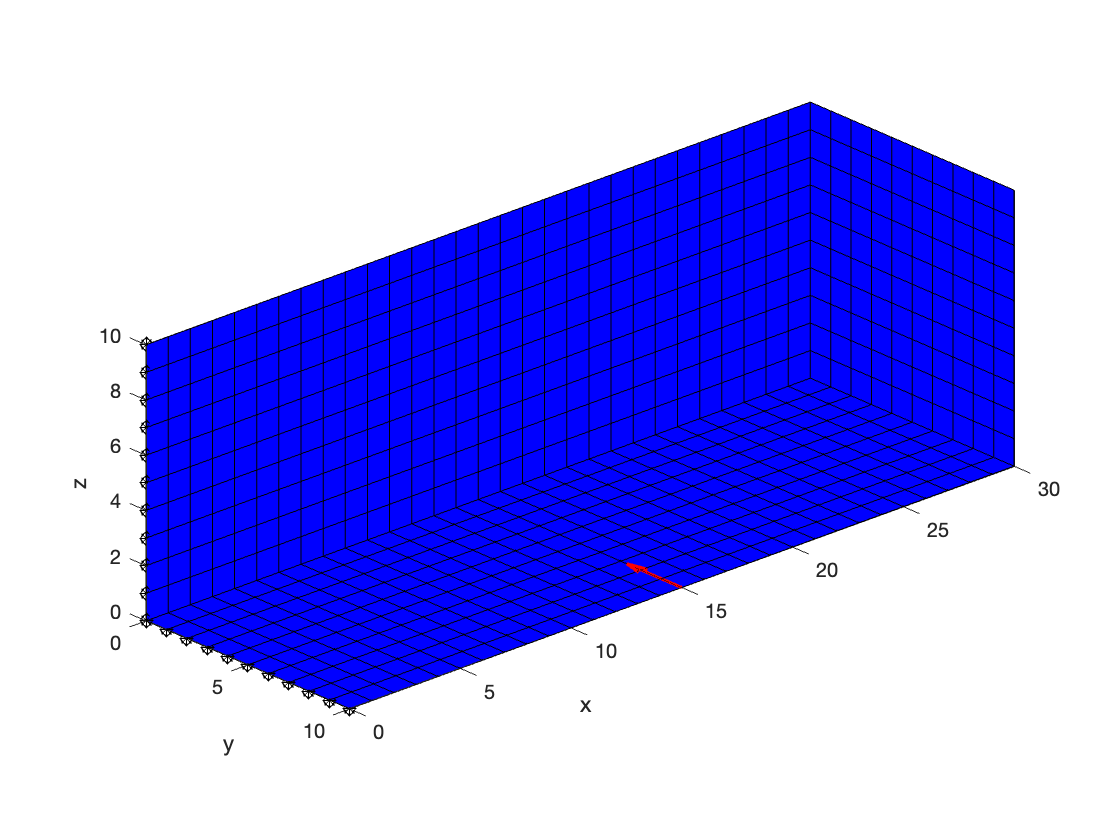}
        \caption{Part 3 loading.}
       \label{fig_3dBC3}
    \end{subfigure}
    \centering
    \caption{Three-body system boundary conditions.}
    \label{fig_3dBC}
\end{figure}

Finally, let us consider the three-body system of \Cref{fig_3dMotions} in three dimensions. The loading conditions and domain sizes are depicted in \Cref{fig_3dBC}. The parts are discretized as following, 1) $35\times25\times10$, 2) $20\times20\times20$, and 3) $30\times 10\times 30$. The first part rotates about the $x$ axis with angle $\theta_1=[\pi,1.1\pi]$. The second part is also rotating about the $x$ axis with angle $\theta_2=[0,0.1\pi]$ with its center at $[35,5,30]$. And the third part is first re-aligned by a $90^\circ$ rotation about $z$ axis and then rotated about $y$ axis with angle $\theta_3=[0,0.1\pi]$ with its center at $[15,20,15]$. The temporal resolution is 100 time steps.

\begin{figure} [h!]
    \centering
    \begin{subfigure}[t]{.3\linewidth}
        \includegraphics[width=\linewidth]{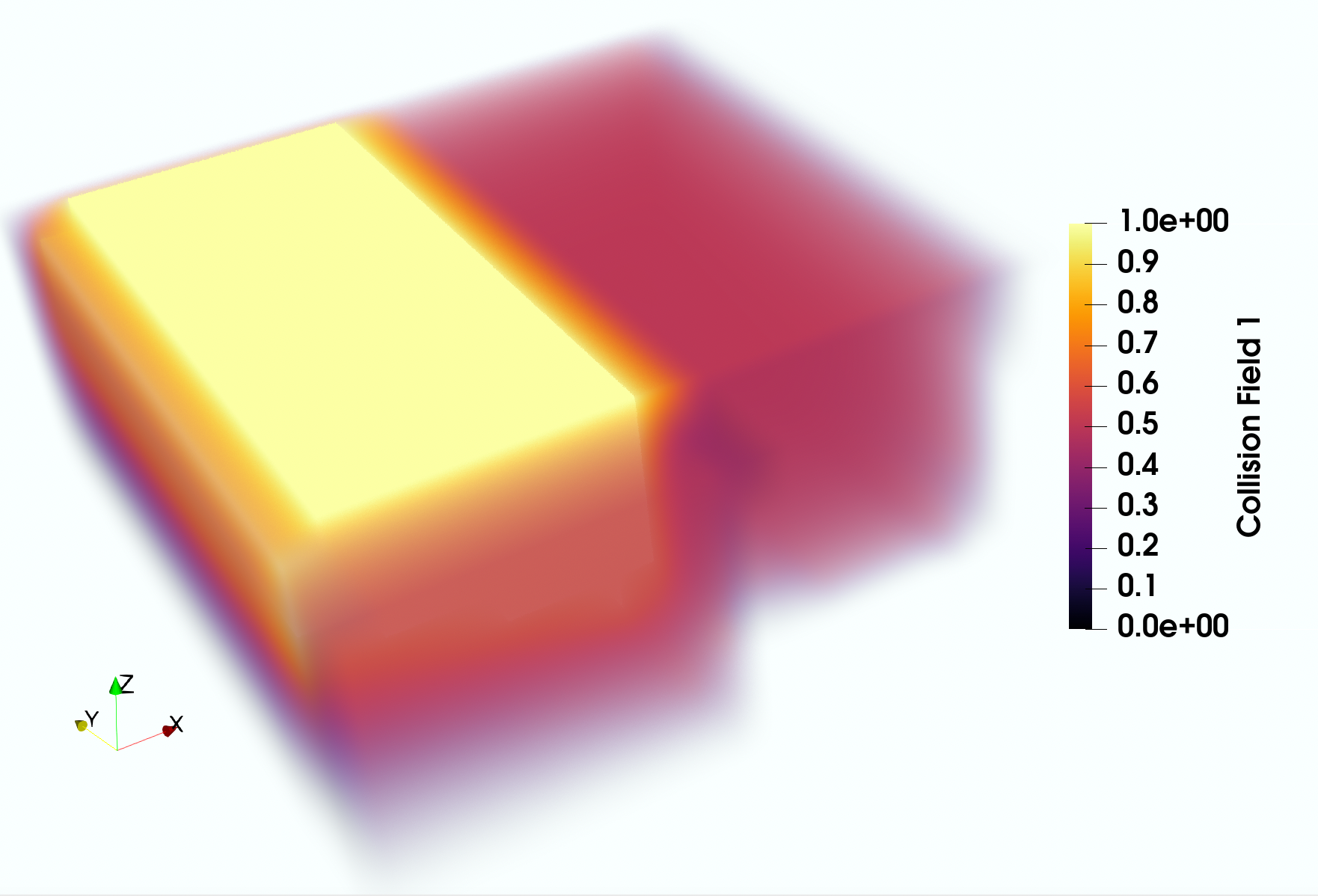}
        \caption{}
        \label{fig_3d_coll1}
    \end{subfigure}
    \begin{subfigure}[t]{.3\linewidth}
        \includegraphics[width=\linewidth]{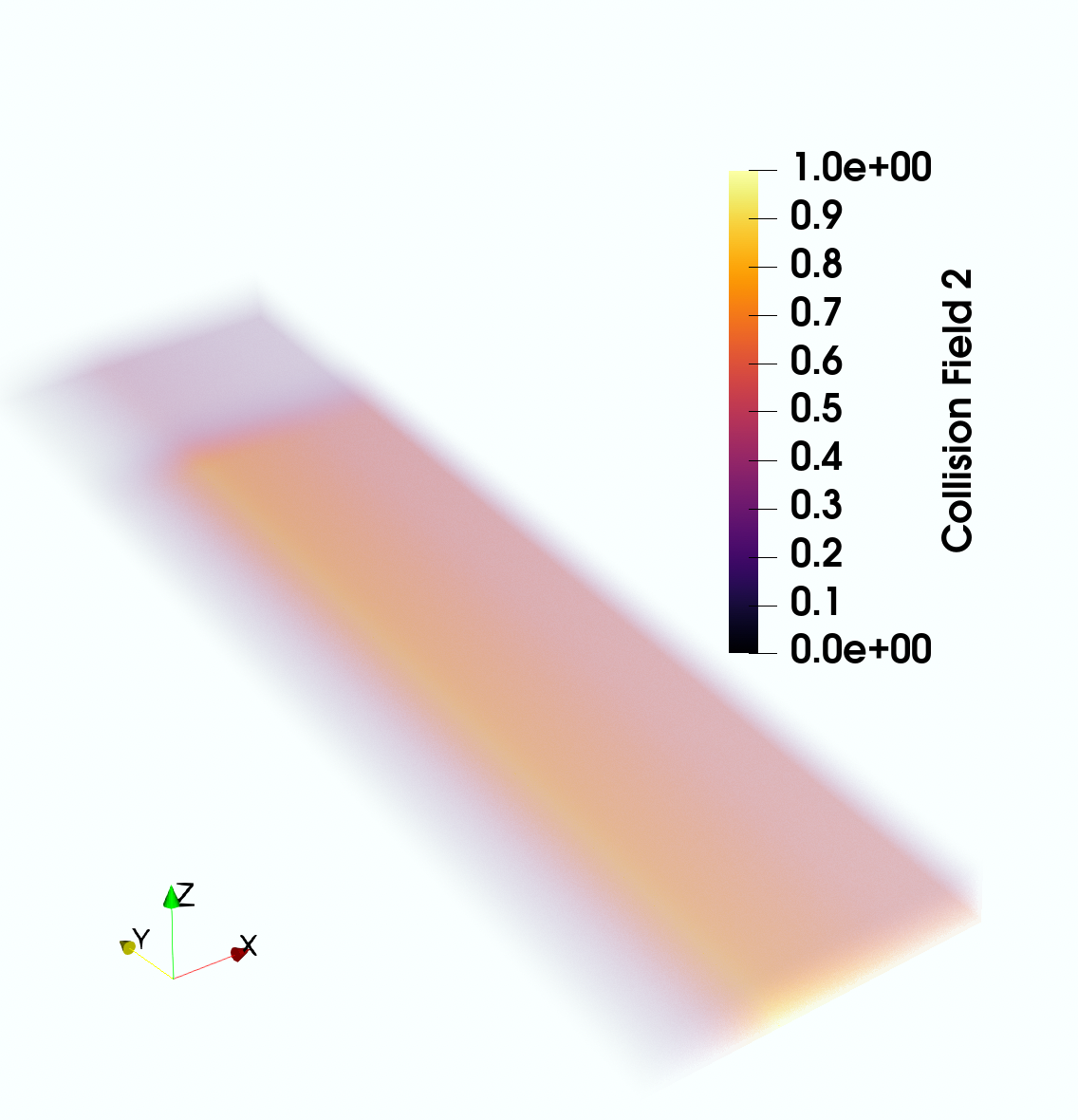}
        \caption{}
       \label{fig_3d_coll2}
    \end{subfigure}
        \begin{subfigure}[t]{.3\linewidth}
        \includegraphics[width=\linewidth]{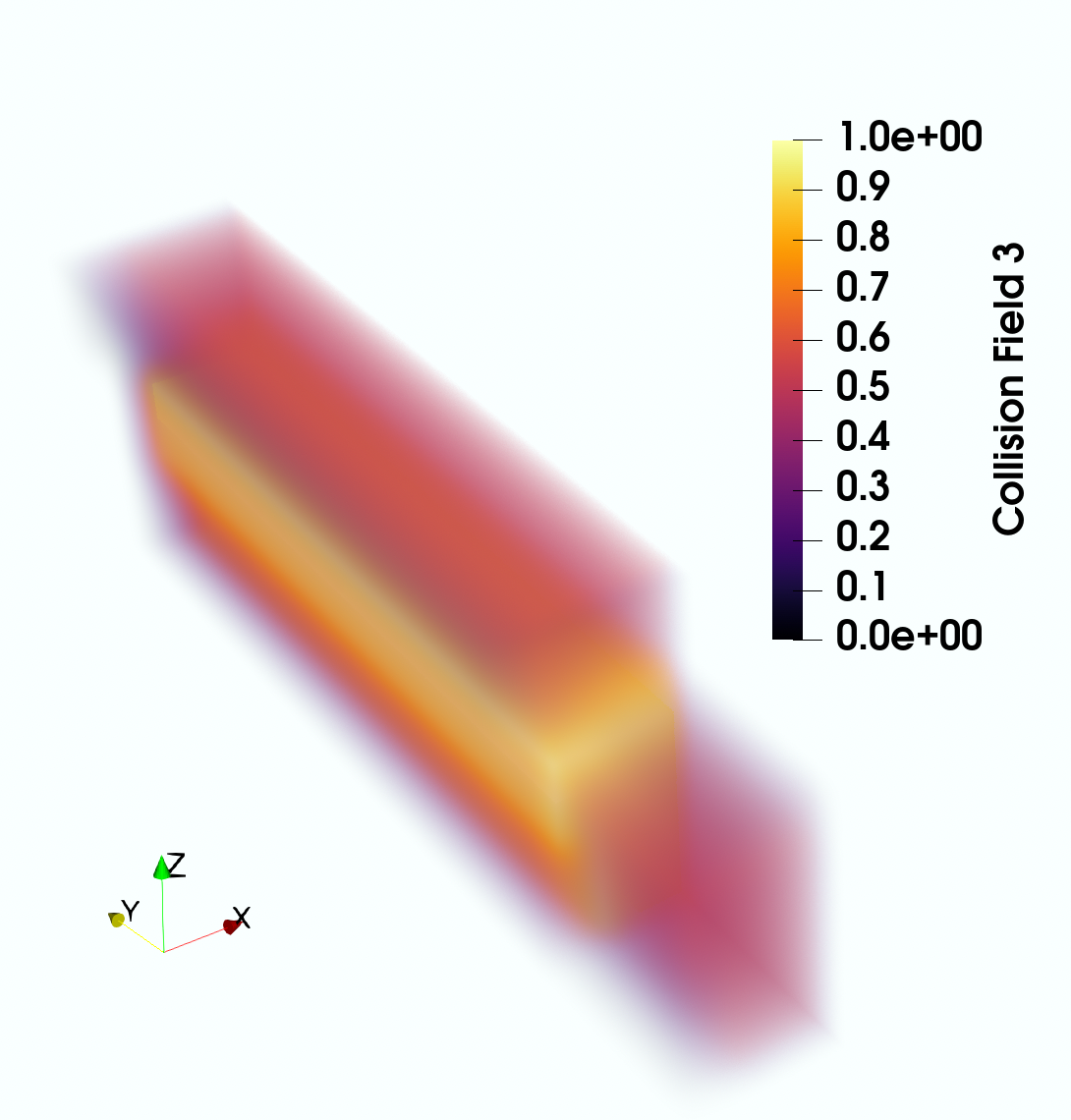}
        \caption{}
       \label{fig_3d_coll3}
    \end{subfigure}
    \centering
    \caption{Three-body system collision measure fields.}
    \label{fig_3dColFields}
\end{figure}

\begin{figure} [h!]
    \centering
    \begin{subfigure}[t]{.6\linewidth}
    \centering
        \includegraphics[width=\linewidth]{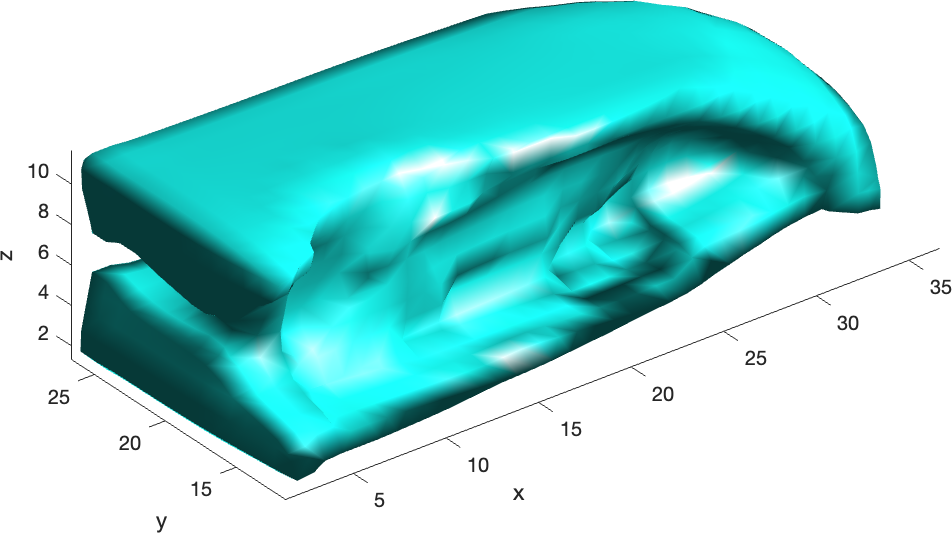}
        \centering
        \caption{Optimized part 1.}
        \label{fig_3d_iso1}
    \end{subfigure}\hfill
    \begin{subfigure}[t]{.4\linewidth}
    \centering
        \includegraphics[width=\linewidth]{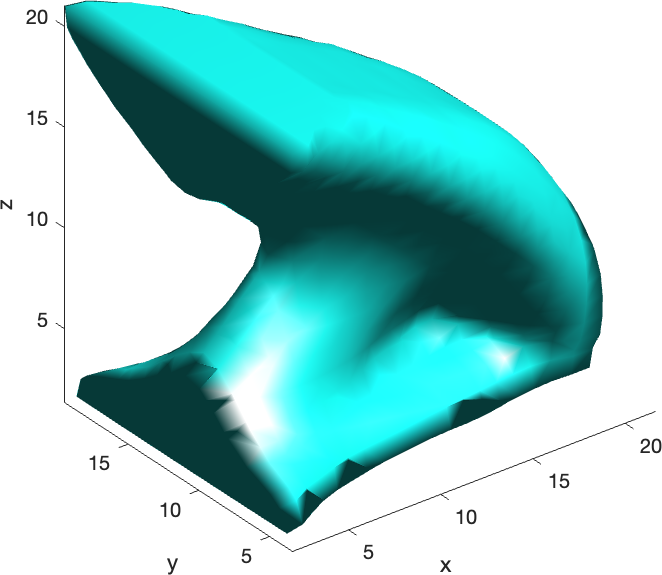}
        \centering
        \caption{Optimized part 2.}
       \label{fig_3d_iso2}
    \end{subfigure}
        \begin{subfigure}[t]{.5\linewidth}
        \centering
        \includegraphics[width=\linewidth]{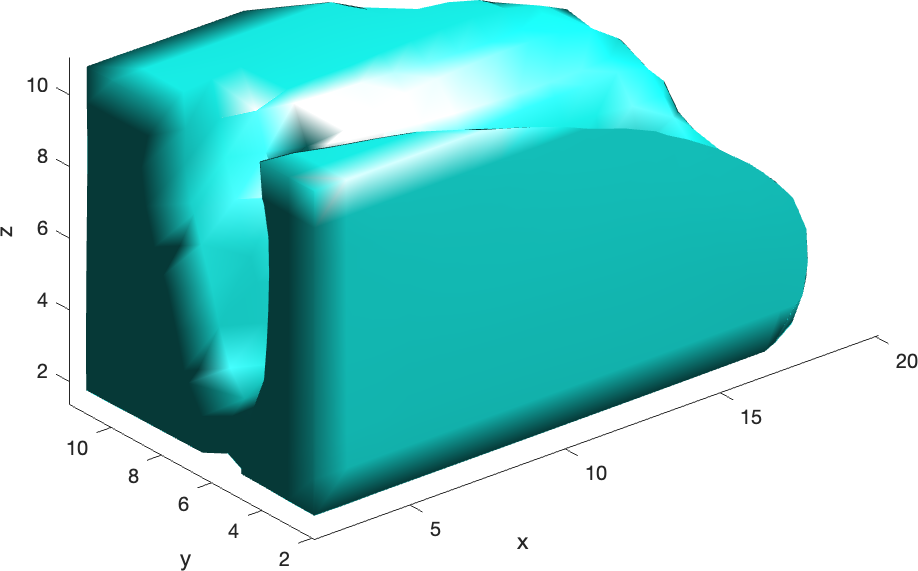}
        \centering
        \caption{Optimized part 3 (front view).}
       \label{fig_3d_iso3_front}
    \end{subfigure}\hfill
        \begin{subfigure}[t]{.5\linewidth}
        \centering
        \includegraphics[width=\linewidth]{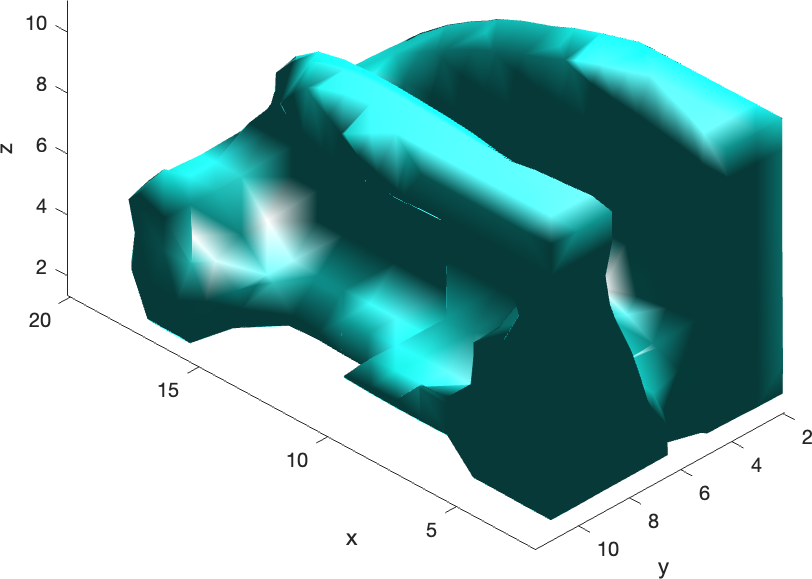}
        \centering
        \caption{Optimized part 3 (back view).}
       \label{fig_3d_iso_back}
    \end{subfigure}
    \centering
    \caption{Co-Optimized three-body system at 0.3 volume fraction with $\lambda_{g_i}=0.5$.}
    \label{fig_3dOpt}
\end{figure}

\begin{figure} [h!]
    \centering
    \begin{subfigure}[t]{0.45\linewidth}
        \includegraphics[width=\linewidth]{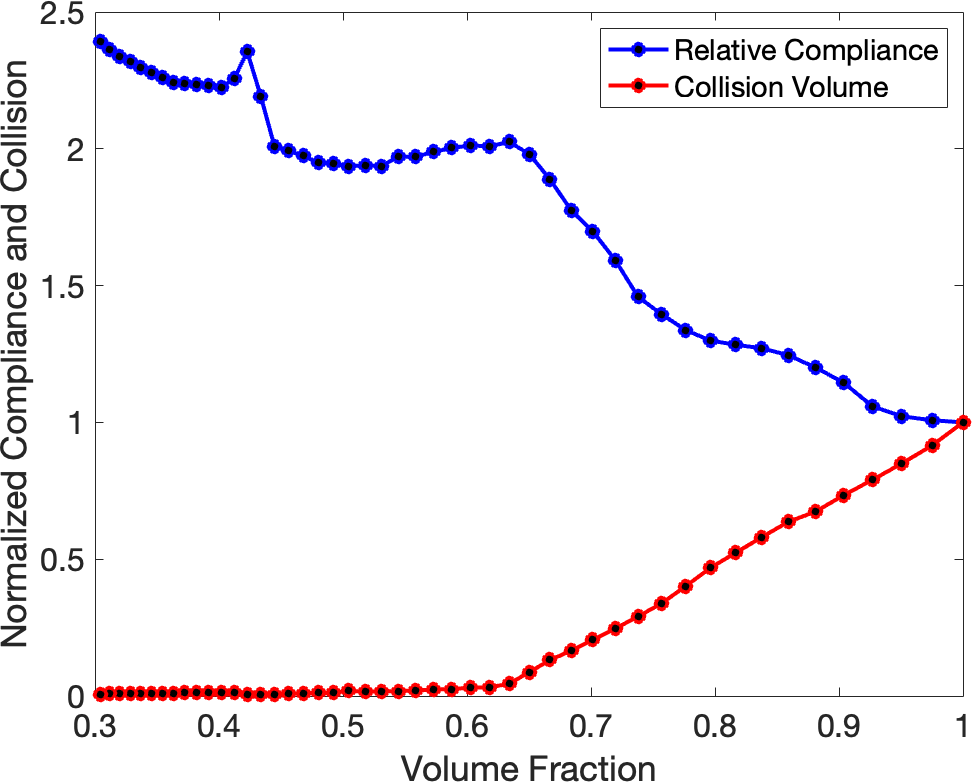}
        \caption{Convergence plot for part 1.}
        \label{fig_3d_conv1}
    \end{subfigure}
    \begin{subfigure}[t]{0.45\linewidth}
        \includegraphics[width=\linewidth]{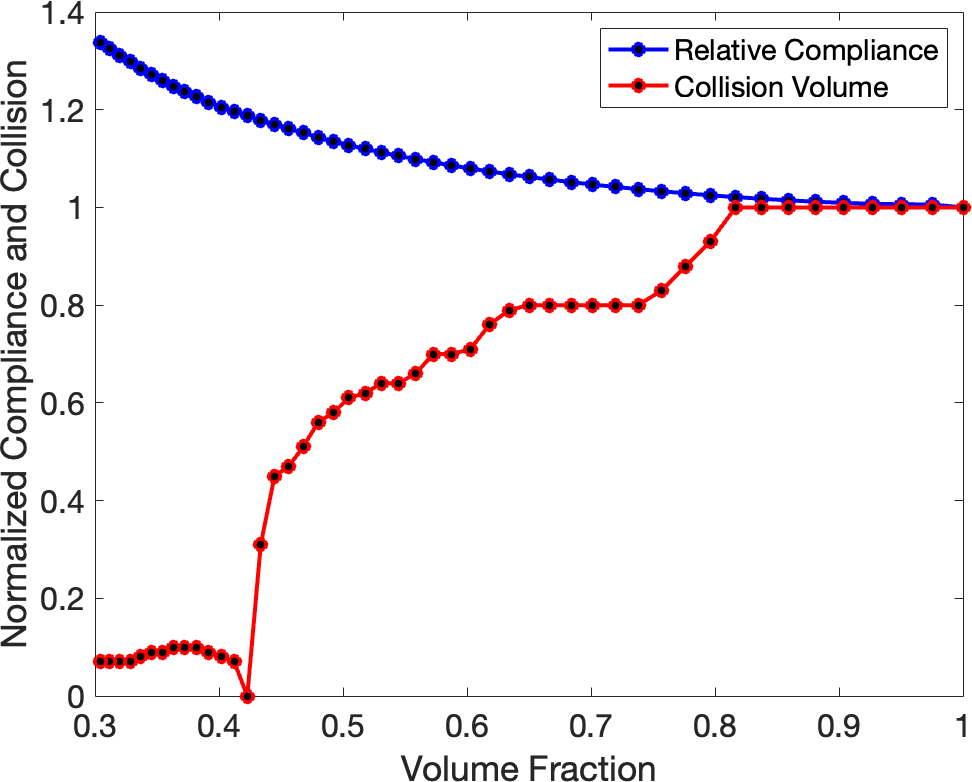}
        \caption{Convergence plot for part 2.}
       \label{fig_3d_conv2}
    \end{subfigure}
        \begin{subfigure}[t]{0.45\linewidth}
        \includegraphics[width=\linewidth]{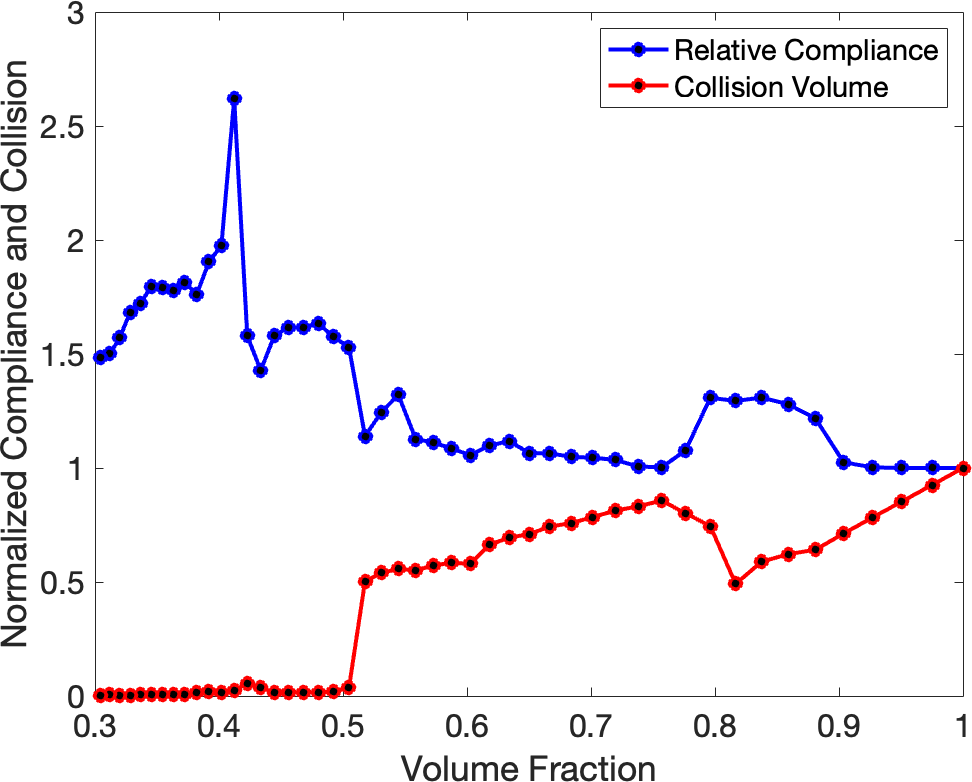}
        \caption{Convergence plot for part 3.}
       \label{fig_3d_conv3}
    \end{subfigure}
    \centering
    \caption{Three-body system convergence plots.}
    \label{fig_3dconvergence}
\end{figure}

\Cref{fig_3dColFields} illustrates the overall collision fields for all three components. Assuming $\lambda_{g_i} = 0.5, ~ \forall i$ and $\gamma_i = 1, ~ \forall i$, \Cref{fig_3dOpt} shows the co-optimized structures at 0.3 volume fraction. \Cref{fig_3dconvergence} illustrates the evolution of compliance and collision volume for each component.

\section{Conclusions} \label{sec_conclusions}

In this paper, we presented a TO formulation for simultaneously co-optimizing multiple components within an assembly. The method is based on a locally differentiable measure of aggregate pair-wise collision between moving parts. To efficiently explore the feasible design space, we have extended the Pareto-tracing TO, where we augment the compliance sensitivity field with the collision gradient and gradually remove material to co-generate high-performance, light-weight, and collision-free structures.

Regarding the design for assembly, our work extends the application of TO beyond the design of individual parts to encompass assembly-level design, where both parts and components are designed concurrently to achieve physics-based goals while also ensuring that parts do not collide.

In the present work, we have assumed that the pairwise collision between parts does not occur between functional surfaces, i.e., at least one of the parts can be modified to avoid collision. 
The proposed work focused on compliance as the performance objective. Future work will extend the current framework to consider local measures, such as stress \cite{suresh2013stress,mirzendehdel2018strength}.  
Future work will also incorporate inertia constraints, dynamic contact loads, and collision-avoidance under large deformations. Another possible extension to our co-design framework is to simultaneously optimize the shapes of parts and their relative motions using differentiable collision measures provided by collision detection surrogate models \cite{das2020learning,dai2020planning,pan2013efficient}.

\section*{Acknowledgments}

This research was developed with funding from the Xerox Corporation. The views, opinions and/or findings expressed are those of the authors and should not be interpreted as representing the official views or policies of the Xerox Corporation.

% The acknowledgements will be added in future revisions, to respect the anonymous review process.

%% The Appendices part is started with the command \appendix;
%% appendix sections are then done as normal sections
% \appendix

%% If you have bibdatabase file and want bibtex to generate the
%% bibitems, please use
%%
 \bibliographystyle{elsarticle-num} 
 \bibliography{CoDesignTopOpt}

%% else use the following coding to input the bibitems directly in the
%% TeX file.

% \begin{thebibliography}{00}

% %% \bibitem{label}
% %% Text of bibliographic item

% \bibitem{}

% \end{thebibliography}
\end{document}